\documentclass[preprint,5p,times,twocolumn]{elsarticle}
\usepackage{microtype}
\usepackage[utf8]{inputenc}
\usepackage{booktabs}
\usepackage[shortcuts,acronym]{glossaries}
\usepackage{xcolor}
\usepackage{siunitx}
\usepackage[hyphens]{url}
\usepackage[misc,geometry]{ifsym}

\usepackage[scaled]{helvet}
\usepackage{lscape}
\usepackage[hidelinks]{hyperref}
\usepackage{tikz}
\usepackage{bm}
\usepackage{diagbox}
\usepackage{siunitx}
\usepackage{amsfonts}

\usepackage{pgfplots}
\usetikzlibrary{calc}
\pgfplotsset{compat=1.8}
\usetikzlibrary{shapes,arrows}

\usepackage[caption=false]{subfig}
\definecolor{darkgreen}{RGB}{0,100,0}
\usepackage[final]{listofsymbols} 

\newacronym{lasso}{LASSO}{least absolute shrinkage and selection operator}
\newacronym{nwp}{NWP}{numerical weather prediction}
\newacronym{ml}{ML}{machine learning}
\newacronym{svr}{SVR}{support vector regression}
\newacronym{svm}{SVM}{support vector machine}
\newacronym{mlp}{MLP}{multi-layer perceptron}
\newacronym{gbrt}{GBRT}{gradient boosting regression tree}
\newacronym{mae}{MAE}{mean absolute error}
\newacronym{rmse}{RMSE}{root mean squared error}
\newacronym{nrmse}{nRMSE}{normalized root mean squared error}
\newacronym{mse}{MSE}{mean squared error}
\newacronym{kld}{KLD}{Kullback-Leibler Divergence}
\newacronym{pv}{PV}{photovoltaic}
\newacronym{wrt}{w.r.t.}{with respect to}
\newacronym{pdf}{PDF}{probability density function}
\newacronym{mtl}{MTL}{multi-task learning}
\newacronym{stl}{STL}{single-task learning}
\newacronym{sps}{SPS}{soft parameter sharing}
\newacronym{hps}{HPS}{hard parameter sharing}
\newacronym{csn}{CSN}{cross-stitch network}
\newacronym{sn}{SN}{sluice network}
\newacronym{ern}{ERN}{emerging relation network}
\newacronym{nlp}{NLP}{natural language processing}
\newacronym{lstm}{LSTM}{long-tem short memory}
\newacronym{tcn}{TCN}{temporal convolution network}
\newacronym{cnn}{CNN}{convolutional neural network}
\newacronym{tl}{TL}{transfer learning}
\newacronym{dtw}{DTW}{dynamic time warping}
\newacronym{elbo}{ELBO}{expected lower bound}
\newacronym{csge}{CSGE}{coopetitive soft gating ensemble}
\newacronym{ae}{AE}{autoencoder}
\newacronym{cnnae}{CAE}{convolutional autoencoder}
\newacronym{vae}{VAE}{variational autoencoder}
\newacronym{dae}{DAE}{denoising autoencoder}
\newacronym{pca}{PCA}{principal component analysis}
\newacronym{rl}{RL}{representation learning}
\newacronym{sgd}{SGD}{stochastic gradient decent}
\newacronym{ttl}{TTL}{transductive transfer learning}
\newacronym{itl}{ITL}{inductive transfer learning}
\newacronym{sqtl}{SQTL}{sequential transfer learning}
\newacronym{aic}{AIC}{Akaike information criterion}
\newacronym{bic}{BIC}{Bayesian information criterion}
\newacronym{bma}{BMA}{Bayesian model averaging}
\newacronym{elm}{ELM}{extreme learning machine}
\newacronym{belm}{BELM}{Bayesian extreme learning machine}
\newacronym{relu}{ReLU}{Rectified Linear Unit}
\newacronym{DBN}{DBN}{deep belief network}
\newacronym{RBM}{RBM}{restricted Boltzmann machines}
\newacronym{RBF}{RBF}{radial-basis function}
\newacronym{blr}{BLR}{Bayesian linear regression}
\newacronym{sp}{SP}{source parameters}
\newacronym{crps}{CRPS}{continuous ranked probability score}

\newcommand\SEC[1]{Section~\ref{#1}} 

\newcommand\FIG[1]{Fig.~\ref{#1}} 
\newcommand\TBL[1]{Table~\ref{#1}} 
\newcommand\TBLs[1]{Tables~\ref{#1}}
\newcommand\EQ[1]{Eq.~\ref{#1}} 
\newcommand{\baseline}{\texttt{baseline}~}

\newcommand{\baselineNS}{\texttt{baseline}}
\newcommand{\WO}{\texttt{WINDOPEN~}}
\newcommand{\WS}{\texttt{WINDSYN~}}
\newcommand{\WR}{\texttt{WINDREAL~}}
\newcommand{\PO}{\texttt{PVOPEN~}}
\newcommand{\PS}{\texttt{PVSYN~}}
\newcommand{\PR}{\texttt{PVREAL~}}

\newcommand{\WONS}{\texttt{WINDOPEN}}
\newcommand{\WSNS}{\texttt{WINDSYN}}

\newcommand{\PONS}{\texttt{PVOPEN}}

\newcommand{\PRNS}{\texttt{PVREAL}}

\opensymdef
\newsym[Size of input features.]{dimFeature}{D}
\newsym[The parameters of a linear model.]{parametersLinModel}{\boldsymbol{\theta}}
\newsym[All input features.]{inputFeatures}{\mathbf{X}}
\newsym[All response features.]{outputFeatures}{\mathbf{y}}
\newsym[A single input vector.]{inputFeatureVector}{\mathbf{x}}
\newsym[A single response respective target.]{outputFeatureVector}{y}
\newsym[A single response prediction.]{outputPredictionVector}{\hat{y}}
\newsym[The set of input features.]{setInputFeatures}{X}
\newsym[The set of response features.]{setOutputFeatures}{Y}
\newsym[Input feature space.]{inputFeatureSpace}{\mathcal{X}}
\newsym[Output feature space.]{outputFeatureSpace}{\mathcal{Y}}
\newsym[The number of samples.]{nSamples}{N}
\newsym[Index of samples.]{indexSamples}{n}
\newsym[Set of all real numbers.]{realNumber}{\mathbb{R}} 
\newsym[Set of all non-negative real numbers.]{realNumberNonNeg}{\mathbb{R}^{+}}
\newsym[Set of all positive real numbers.]{realNumberPos}{\mathbb{R}_{\geq 1}} 
\newsym[Set of all natural numbers.]{naturalNumber}{\mathbb{N}} 
\newsym[Set of all non-negative natural numbers.]{naturalNumberNonNeg}{\mathbb{N}^{+}} 
\newsym[Set of all positive natural numbers.]{naturalNumberPos}{\mathbb{N}_{\geq 1}}
\newsym[Normal distribution.]{normalDist}{\mathcal{N}} 
\newsym[A task.]{task}{\mathcal{T}} 
\newsym[The set of all tasks.]{allTasks}{\mathbb{T}} 
\newsym[Similarity measure between two task.]{similarityMeasure}{\mathcal{S}} 
\newsym[Precision matrix of a linear model.]{precisionMatrix}{\mathbf{S}} 
\newsym[Index of a source model.]{indexSourceModel}{m} 
\newsym[The number of source models.]{numberSourceModels}{M} 
\newsym[The dimension of the target.]{dimensionTarget}{I} 
\newsym[The index of the target task.]{indexTargetTask}{T} 
\newsym[CSGE weight not normalized.]{csgeWeightUnnormalized}{\bar{w}} 
\newsym[CSGE weight normalized.]{csgeWeightNormalized}{w} 
\closesymdef

\usepackage{wrapfig}
\usetikzlibrary{shapes.geometric, arrows}

\usepackage{enumitem}

\usepackage{amsthm}
\theoremstyle{definition}
\newtheorem{question}{Research Question}

\begin{document}
\title{Model Selection, Adaptation, and Combination for Transfer Learning in Wind and Photovoltaic Power Forecasts}



\begin{frontmatter}


   \author{Jens Schreiber\corref{mycorrespondingauthor}}
    \ead{j.schreiber@uni-kassel.de}
    \author{Bernhard Sick}
    \ead{bsick@uni-kassel.de}
    \address{University of Kassel, Wilhelmsh\"oher Allee 71, 34121 Kassel, Germany}



     \cortext[mycorrespondingauthor]{Corresponding author}


    \begin{abstract}
        There is recent interest in using model hubs, a collection of pre-trained models, in computer vision tasks.
        To utilize the model hub, we first select a source model and then adapt the model for the target to compensate for differences.
        While there is yet limited research on model selection and adaption for computer vision tasks, this holds even more for the field of renewable power.
        At the same time, it is a crucial challenge to provide forecasts for the increasing demand for power forecasts based on weather features from a numerical weather prediction.
        We close these gaps by conducting the first thorough experiment for model selection and adaptation for transfer learning in renewable power forecast, adopting recent results from the field of computer vision on 667 wind and photovoltaic parks.
        To the best of our knowledge, this makes it the most extensive study for transfer learning in renewable power forecasts reducing the computational effort and improving the forecast error.
        Therefore, we adopt source models based on target data from different seasons and limit the amount of training data.
        As an extension of the current state of the art, we utilize a Bayesian linear regression for forecasting the response based on features extracted from a neural network.
        This approach outperforms the baseline with only seven days of training data.
        We further show how combining multiple models through ensembles can significantly improve the model selection and adaptation approach.
    \end{abstract}




\end{frontmatter}

\section{Introduction}
With the extension of volatile energy resources, such as wind and~\ac{pv} parks, one fundamental problem is adding new parks to an operator’s portfolio.
The historical data for such a new (target) park is often limited.
At the same time, reliable forecasts are fundamental to assure grid stability due to the weather dependency.
However, typically there are numerous pre-trained models from existing parks that we can utilize for such a forecasting task~\cite{Schreiber2019a}.
Utilizing those pre-trained source models often increases the forecast accuracy, reduces the computational effort, and reduces the carbon emission for training a new model~\cite{Schwartz2019,Schreiber2021}.
Now, the question arises \textit{what is the best way to use} of this \textit{model hub} of pre-trained models.
The field of research~\ac{itl} provides methods for this problem~\cite{You2021LogMe}.

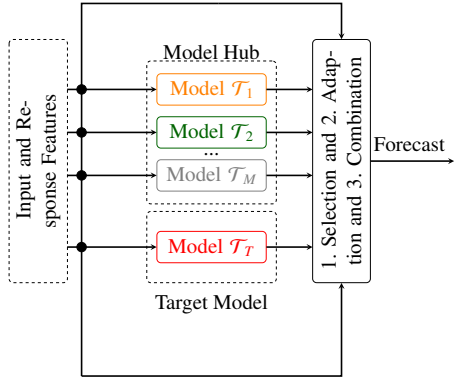
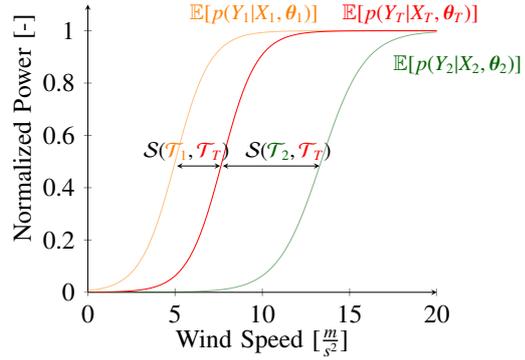
\begin{figure*}[tb]\centering
\subfloat[Process of model selection, adaptation, and combination.\label{fig_model_selection_and_adaption}]{
\scalebox{0.38}{
{
\huge

\tikzstyle{block} = [rectangle, rounded corners, minimum width=3.5cm, minimum height=2cm,text centered, text width=5cm, draw=black]
\tikzstyle{block2} = [rectangle, rounded corners, minimum width=1cm, minimum height=1cm, text centered, text width=3cm, draw=black]
\tikzstyle{cbox} = [rectangle, rounded corners, minimum width=1cm, minimum height=1cm, text centered, draw=black]

\begin{tikzpicture}[node distance=1.5cm]
    \tikzstyle{arrow} = [ultra thick,->,>=stealth]

    \node (sd) [xshift=0cm] {};
    \node (comb) [block, below of=sd, right of=sd, , xshift=3cm, yshift=-4cm, rotate=90, text width=8cm] {1.~Selection and 2.~Adaption and 3.~Combination};

    \node (m1) [block2, left of=comb, below of=sd, xshift=1.5cm, yshift=-1.5cm, color=orange, text width=3.3cm] {Model $\mathcal{T}_1$};
    \node (m2) [block2, below of=m1, color=darkgreen, text width=3.3cm] {Model $\mathcal{T}_2$};
    \node (m3) [block2, below of=m2, color=gray, text width=3.3cm] {Model $\mathcal{T}_M$};
    \node (m4) [block2, below of=m3, yshift=-1cm, color=red, text width=3.3cm] {Model $\mathcal{T}_T$};

    \draw [arrow] (sd) -| node[] {} (comb);

    \draw [ultra thick] (-4.5, -13) |- node[] {} (sd.east);
    \draw [arrow] (-4.5, -13) -| node[] {} (comb.west);

    \draw [arrow] (m1.east) -- node[] {} (3.5, -3);
    \draw [arrow] (m2.east) -- node[] {} (3.5, -4.5);
    \draw [arrow] (m3.east) -- node[] {} (3.5, -6);
    \draw [arrow] (m4.east) -- node[] {} (3.5, -8.5);

    \draw [arrow] (comb.south) -- node[anchor=east, yshift=0.6cm, xshift=1.4cm] {Forecast} (8.4, -5.5);

    \draw [arrow, ] (-5, -3) -- node[] {} (m1.west);
    \draw [arrow] (-5, -4.5) -- node[] {} (m2.west);
    \draw [arrow] (-5, -6) -- node[] {} (m3.west);
    \draw [arrow] (-5, -8.5) -- node[] {} (m4.west);

    \node (pv) [cbox, right of=m2, xshift=-7.5cm, yshift=-1cm, rotate=90, dashed, text width=8cm] {Input and Response Features};

    \node (bb) [cbox, xshift=0cm, yshift=-4.5cm, text width= 4cm, text height=4.5cm, draw=black, dashed] {};
    \node (ob) [cbox, xshift=0cm, yshift=-8.5cm, text width= 4cm, text height=2cm, draw=black, dashed] {};

    \node (sm) [above of=bb, yshift=1.3cm, color=black] {Model Hub};
    \node (tm) [below of=ob, yshift=-0.5cm, color=black] {Target Model};

    \node (dots1) [below of=m2, yshift=0.75cm] {...};

    \draw[black,fill=black] (-4.5, -3) circle (.5ex);
    \draw[black,fill=black] (-4.5, -4.5) circle (.5ex);
    \draw[black,fill=black] (-4.5, -6) circle (.5ex);
    \draw[black,fill=black] (-4.5, -8.5) circle (.5ex);

\end{tikzpicture}
}}}
\qquad
\subfloat[Similarity between sources $\task_1$ and $\task_2$ with a target task $\task_T$ for mean wind power predictions.\label{fig_task_sim_wind_example}]{
\scalebox{0.67}{
{
\Large
\begin{tikzpicture}
	\begin{axis}%
		[
			grid=minor,
			xmin=0,
			xmax=20,
			axis x line=bottom,
			ytick={0,0.2,...,1.0},
			ymax=1.1,
			axis y line=left,
		]
		\addplot%
		[
			orange,%
			mark=none,
			mark options={scale=4},
			opacity=0.5,
			samples=100,
			domain=-0:20,
		]
		(x,{1/(1+exp(-x+5))});
		\addplot%
		[
			red,%
			mark=none,
			samples=100,
			mark options={scale=4},
			xticklabels={\empty},
			yticklabels={\empty},
			domain=-0:20,
		]
		(x,{10/(10+exp(-x+10))});
		\addplot%
		[
			darkgreen,%
			mark=none,
			mark options={scale=4},
			opacity=0.5,
			samples=100,
			xticklabels={\empty},
			yticklabels={\empty},
			domain=-0:20,
		]
		(x,{0.5/(0.50+exp(-x*0.8+10))});
	\end{axis}
   
	\node[text width=3.5cm,xshift=-0.5cm] at (4,-1)  {Wind Speed [$\frac{m}{s^2}$]};
	\node[text width=3cm] at (3.5,5.5)  {{\large \color{orange} $\mathbb{E}[p(\setOutputFeatures_{1}|\setInputFeatures_{1}, \parametersLinModel_{1})]$}};
	\node[text width=3cm, text=red] at (6.5,5.5)  {\large $\mathbb{E}[p(\setOutputFeatures_{T}|\setInputFeatures_{T}, \parametersLinModel_{T})]$};
	\node[text width=3cm, text=gray] at (7.5,4.5)  {{\large \color{darkgreen} $\mathbb{E}[p(\setOutputFeatures_{2}|\setInputFeatures_{2}, \parametersLinModel_{2})]$}};

	\node[text width=5cm, rotate=90, yshift=0.25cm] at (-1,3.5)  {Normalized Power~[-]};

	\draw [stealth-stealth](1.75,2.5) -- (2.6,2.5) ;
	\node[text width=3cm] at (2.6,2.8)  {\large $\similarityMeasure({\color{orange} \mathcal{T}_1}, {\color{red} \mathcal{T}_T})$};

	\draw [stealth-stealth](2.65,2.5) -- (4.55,2.5) ;
	\node[text width=3cm] at (4.6,2.8)  {\large $\similarityMeasure({\color{darkgreen} \mathcal{T}_2}, {\color{red} \mathcal{T}_T})$};
	
\end{tikzpicture}

}}}
\caption{Diagrams illustrating the knowledge transfer based on a model hub.}
\end{figure*}
\FIG{fig_model_selection_and_adaption} summarizes our proposed strategy.
The first step is to select an appropriate source model.
Recently,~\cite{You2021LogMe,You2021Hub} showed that selecting an appropriate source model for knowledge transfer for a target substantially influences the test error for computer vision tasks.
To \textbf{select} a source model, consider, e.g., we have two source tasks $\task_1$ and $\task_2$ of a wind park with model parameters $\parametersLinModel_1$ and $\parametersLinModel_2$. 
These models have a set of input observations $\setInputFeatures_1$ and $\setInputFeatures_2$ as well as the sets of response values $\setOutputFeatures_1$ and $\setOutputFeatures_2$.
Based on this information, we want to select one of the models for knowledge transfer for a target task $\task_T$ with parameters $\parametersLinModel_T$ and its respective sets $\setInputFeatures_T$ and $\setOutputFeatures_T$.

The diagram in~\FIG{fig_task_sim_wind_example} visualizes this problem for wind power forecasts.
In renewable power forecasts, we utilize weather forecasts, such as the wind speed or radiation, from a so called~\ac{nwp} model.
These predicted weather features are the input $\setInputFeatures$ to \ac{ml} models, with parameters $\parametersLinModel$, predicting the expected power generation $\mathbb{E}[p(\setOutputFeatures|\setInputFeatures, \parametersLinModel)]$ in a \textit{day-ahead} forecasting task between $24$ and $48$ hours into the future.
In the diagram, we can observe that the similarity $\similarityMeasure$ depends on the relation between the input feature wind and the power generated by a wind park, i.e., for different wind speeds and models, we expect a different power generation.

Once a model is selected, the second step adapts the source knowledge with the limited target data with an~\textit{adaptation strategy}.
Often this adaptation strategy is fine-tuning the final layer of a neural network.
Only with such an adaptation can we make reliable and task-specific power forecasts for a new target task of a new park with limited data.
To the best of our knowledge, different adaptation and selection strategies have been so far not considered for~\ac{itl} in the field of renewable power forecasts.
We close this gap with this article.
Selecting and adapting a single source model from a model hub has the disadvantage of neglecting knowledge from other source tasks that are potentially beneficial.
However, we can additionally \textbf{combine} models through ensemble techniques.
Combining knowledge can be seen as a mixture of selection, where more relevant models are weighted higher, and adaptation, where source models are weighted to fit the target.
Since~\ac{itl} has been insufficiently studied so far for renewable power forecasts~\cite{Alkhayat2021}, especially for day-ahead forecast horizons between $24$ and $48$ hours into the future, we answer the following research questions:
\begin{question} 
What is an appropriate similarity measure for model selection for a new target park from a model hub with pre-trained models?
\end{question}
\begin{question} 
 What is the best adaptation strategy once a model is selected?
\end{question}
\begin{question} 
 Compared to selecting and adapting a single model, are ensemble strategies beneficial for combining knowledge?
\end{question}

To answer those research questions, we train a~\ac{belm}, a~\ac{mlp}, and a~\ac{tcn} as source models on six datasets including $667$ distinct parks.
We consider the~\ac{gbrt} as a \baseline that outperforms physical models even with limited target data~\cite{VogtSynData2022}.

Based on these datasets and source models, our main contributions to answering the research questions can be summarized as follows:
First, we propose the~\ac{nrmse}, due to its strong correlation with the forecasting error, as well as the marginal likelihood or evidence as a~\textbf{selection} strategy for question one.
Therefore, we replace the last layer of a neural network with a~\ac{blr}, as proposed in~\cite{You2021LogMe} for computer vision tasks. 
Second, we propose to utilize the~\ac{blr} as a model to~\textit{adapt} a source model to the target and answer question two. 
To ensure that we find the best adaption strategy, we \textit{directly} apply a source model on the target and consider the state-of-the-art such as weight decay as well as Bayesian tuning~\cite{You2021LogMe}. 
Considering the two selection strategies, we evaluate $18$ combinations making it the most extensive evaluation for~\ac{tl} in renewable power forecasts.
Third, due to the~\ac{blr} as the output layer, we propose~\textbf{combining} multiple source models through~\ac{bma} to answer question three. 
Finally, we propose the~\ac{csge} as ensemble technique for~\ac{itl}.
The source code is open accessible\footnote{\url{https://github.com/scribbler00/DEELEA}, accessed 2022-07-15}.

The remainder of this article is structured as follows. 
~\SEC{sec_itl_selection_related_work} describes related work. 
The following~~\SEC{sec_itl_selection_method} introduces relevant definitions and details the proposed approach. 
We describe the datasets and discuss the experiment's most essential findings in~\SEC{sec_itl_selection_experiment}. 
In the final~\SEC{sec_itl_selection_conclusion}, we summarize our work and provide insights for future work.
\listofsymbols
\section{Related work}
\label{sec_itl_selection_related_work}
In the following section, we overview recent developments for~\ac{tl} and, more specifically for~\ac{itl} in computer vision that has been not considered for renewable power forecasts.
This review determines relevant techniques that we consider for renewable power forecasts.
Afterward, we summarize related work for~\ac{itl} on deterministic renewable power forecasts.

There are two crucial dimensions in~\ac{itl}.
The first is the model selection and the second is the adaptation strategy.
The authors of~\cite{You2021LogMe,You2021Hub} provide a study on selection strategies for the field of computer vision.
They utilize a~\ac{blr} replacing the final layer of a source model and train it through empirical Bayes, also referred to as evidence approximation, on the target data.
The authors repeat this approach for each available model from the model hub.
Finally, they determine the similarity through the evidence of a source model on the target.
As an adaptation strategy, they proposed Bayesian tuning, which regularizes the fine-tuning process by predictions from multiple sources.
These proposed selections and adaptations need to be considered and extended for renewable energies.
For instance, we can directly forecast through the~\ac{blr} and compare it with fine-tuning of the final layer.
Often, the adaptation through fine-tuning is regularized by a weight decay regarding zero~\cite{Li2020}.
However, this regularizer does not consider parameters originating from the source model.
Therefore, in~\cite{Li2018} a deviation from a source model is penalized by weight decay considering the source model parameters.

Compared to research areas like computer vision, there is limited research on~\ac{itl} for renewable power forecasts, compare~\cite{Alkhayat2021}.
There has been some work to learn a transferable representation of the input utilizing autoencoders~\cite{Qureshi2019,Liu2021,Ju2020,Henze2020}.
While the principle approach of transferring an autoencoder for a target is combinable with our approach, we argue that considering the conditional distribution of the power forecast is more relevant for model selection and combination.
The data driven~\ac{tl} approaches presented in~\cite{Cao2018} and \cite{Cai2019} are outside the scope of this article.

Also, most of the current research on~\ac{tl} in renewable power forecasts focuses on meteorological measurements as input features, see \cite{Liu2021,Ju2020,Cao2018,Liu2022,Chen2020,Sheng2022,Yin2021,Khan2022,Almonacid2022,Yan2019}. 
These articles consider forecast horizons between ten seconds and two hours. 
At the same time, larger forecast horizons, such as day-ahead forecasts, are inherently more difficult as they utilize~\ac{nwp} as input features and forecast errors increase with an increasing forecast horizon~\cite{Jens2019}.

Most of the previously mentioned related work for~\ac{tl} in renewable energies is treating power forecast not as a time series approach.
Instead, the authors treat it as a regression.
At the same time, periodic influences from, e.g., the diurnal cycle, are well known.
Therefore, the authors of \cite{Zhou2020} consider recurrent networks and fine-tuning to achieve good results for an ultra-short-term forecast horizon of~\ac{pv}.
Additionally, the article \cite{Schreiber2021} considers time series methods in a~\ac{mtl} architecture.

The authors of~\cite{Ceci2017} achieve improvements in day-ahead~\ac{pv} forecasts through multi-target models.
The idea of model combination is similar to our ensemble approach.
However, the authors of this article do not evaluate it in the context of~\ac{itl}.
The study of~\cite{Shireen2018} proposes an \ac{mtl} strategy for Gaussian processes to forecast \ac{pv} targets.
By clustering wind parks, a weighting scheme provides predictions for a new park in~\cite{Tasnim2018}.
In this article, no actual historical power measurements are used for evaluation; instead, the authors used synthetic data.

A number of articles apply~\ac{mtl} architectures for~\ac{tl}~\cite{Schreiber2021,Vogt2019,Schreiber2020} in day-ahead forecasts.
The proposed task embedding in~\cite{Schreiber2021} and~\cite{Vogt2019} for~\acp{mlp} and~\acp{cnn}, encodes task-specific information through an embedding to learn latent similarities between tasks.
The article~\cite{Schreiber2021} is especially interesting as we have a similar experimental set-up.

However, the authors look at errors per season and results can be misleading as an~\ac{itl} approach should avoid catastrophic forgetting for all seasons.
Also,~\ac{mtl} architectures are due to their additional training complexity not common in the industry, e.g., due to data pre-processing and training complexity.
Therefore, for the extension of renewable energies making the best use of existing single-task models for~\ac{itl} is essential.

Furthermore, none of the related work studies different model selection and model adaptation strategies for neural network architectures.
While there has been some work on combining knowledge from multiple sites for~\ac{pv} through ensemble-like strategies, the studies are insufficient as they do not consider the amount of available data.
Also, the expected power is solely based on a characteristic curve, or authors consider solely~\ac{pv} or wind data.
We close these research gaps by providing an extensive study that overcomes those limitations for day-ahead forecasts.
\section{Proposed Methods}
\label{sec_itl_selection_method}
The following sections define the proposed model selection, model adaptation, and model combination strategies. 
Beforehand we introduce the~\ac{blr} as it is the basic model for one selection and one adaption strategy.

\subsection{Bayesian Linear Regression}
The following definitions of a~\ac{blr} makes use of the introductions in~\cite{Bishop2006} and~\cite{deisenroth2020mathematics}.
In contrast to a deterministic perspective to learn the model weights of a linear regression model, a Bayesian approach gives additional insights through the posterior, especially when there is insufficient data~\cite{Bishop2006} as for~\ac{tl}.
It helps in measuring task similarity for model selection and allows assessing the quality of the model in terms of its uncertainty, as proposed in~\cite{You2021LogMe}.
We propose to utilize it also for an adaptation for renewable forecasts.
We achieve this by replacing the final layer of a neural network with a~\ac{blr}, training it with the target data, and making predictions for the target afterward.
This approach additionally allows combining models through~\ac{bma}, see~\SEC{sec_method_model_combination}.
Finally, we can utilize it to learn a~\ac{belm}.

Therefore, let us assume that the following equation details the posterior distribution of such a linear model:
\begin{equation}
    \underbrace{p(\parametersLinModel|\setInputFeatures, \setOutputFeatures)}_{\text{posterior}} = \frac{\overbrace{p(Y|X, \parametersLinModel)}^{\text{likelihood}}\overbrace{p(\parametersLinModel}^{\text{prior}})}{\underbrace{p(\setOutputFeatures|\setInputFeatures)}_{\text{marginal likelihood}}},
\end{equation}
where $\setInputFeatures = \{\inputFeatureVector_n\}_{n=1}^{n=\nSamples}$ and $\setOutputFeatures=\{\outputFeatureVector_n\}_{n=1}^{n=\nSamples}$ are the sets of observed input and response values with $\nSamples \in \naturalNumberPos$ samples from a training dataset.
In this setting, a single feature vector $\inputFeatureVector_n \in \realNumber^\dimFeature$ has $\dimFeature \in \naturalNumberPos$ features and $\outputFeatureVector_n$ is of size $\realNumber$.
Then the likelihood $p(\setOutputFeatures|\setInputFeatures, \parametersLinModel)$ describes how well $\setInputFeatures$ 
and the weights $\parametersLinModel\in \realNumber^D$ describe the response values.
Through the prior, we encode our initial beliefs about the model weights.
The marginal likelihood normalizes the posterior.
Finally, after observing training data, the posterior encodes what we know about the target.

To calculate the distributions of the posterior of a linear regression model consider that we have a prior over the weights $\parametersLinModel$ with $p(\parametersLinModel|\alpha) = \normalDist(\parametersLinModel| 0, \alpha^{-1}\mathbf{I})$,
where $\alpha \in \realNumberNonNeg$ is the precision of the zero mean isotropic Gaussian distribution.
Note, choosing an isotropic Gaussian distribution for the prior allows deriving a closed form solution that reduces computational effort for calculating the mean and covariance matrix, as detailed in the following.

Now consider that we have a target $\outputFeatures$ of size $1 \times N$ and $\inputFeatures$ is the $N \times \dimFeature$ design matrix, where each row corresponds to the $n$-th observation.
For multivariate problems, we can train one model per response.
In our case, $\inputFeatures$ are either weather predictions from an~\ac{nwp} such as wind speed or radiation, random features from an~\ac{belm}, or features extracted from a neural network at the second last layer.
In the latter two cases the weather predictions are transformed through the neural network or the~\ac{belm}.

Finally, the posterior distribution is given by $\outputFeatures$, $\inputFeatures$, and the noise precision parameter $\beta \in \realNumberNonNeg$ through $p(\parametersLinModel| \inputFeatures, \outputFeatures) = \normalDist(\parametersLinModel | \mathbf{m}_N ,\precisionMatrix_N)$, where
\begin{equation}\label{eq_blr_online_update_mean}
    \mathbf{m}_N = \beta \precisionMatrix_N \inputFeatures^T \outputFeatures\text{~and~}\precisionMatrix_N^{-1} = \alpha\mathbf{I} + \beta \inputFeatures^T\inputFeatures.
\end{equation}
In this setting, $\nSamples$ indicates the number of training samples used to update our prior beliefs of the model weights.
In most cases, we are interested in predicting an unknown response $\outputFeatures_{*}$ based on input $\inputFeatures_{*}$ from a (test) dataset not seen during the training of the model.
Therefore, the predictive posterior is defined by:
\begin{align}\label{eq_lin_predictive_posterior}
    p(\inputFeatures_{*}|\outputFeatures, \alpha, \beta) & =  
    \int p(\inputFeatures_{*}|\outputFeatures, \parametersLinModel) p(\parametersLinModel|\inputFeatures, \outputFeatures) \text{d}\parametersLinModel \nonumber \\                                                 &=\normalDist(y_{*}|\inputFeatures_{*} \mathbf{m}_N, \boldsymbol{\sigma}_N^2(\inputFeatures_{*})),
\end{align}
where $\outputFeatures$ and $\inputFeatures$ are from the training set and the posterior variance is given by
\begin{equation}
    \boldsymbol{\sigma}^2_N(\inputFeatures_{*}) = \beta^{-1} + \inputFeatures^T_{*} \precisionMatrix_N \inputFeatures_{*}.
\end{equation}
\subsection{Model Selection for Inductive Transfer Learning}\label{sec_model_selection_definition}

Measuring task similarity between a target task and multiple source tasks is a critical challenge in~\ac{itl}~\cite{You2021LogMe} as it allows selecting an appropriate source task for knowledge transfer.
Ideally, a valid model selection avoids negative transfer, so utilizing knowledge from the source model has a minor error than training a target model from scratch.

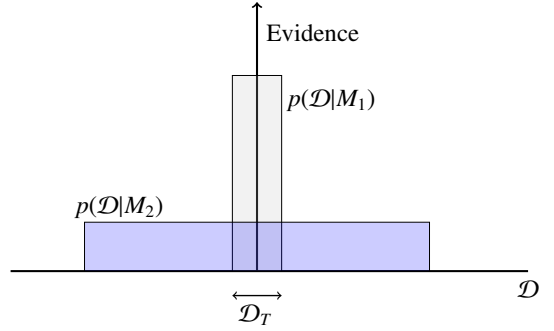
\begin{figure}
    \centering
    {
\Large
\resizebox{0.39\textwidth}{!}{%
    \begin{tikzpicture}

        \draw[very thick,-] (0,0) -- (10.5,0) node[anchor=north] {$\mathcal{D}$};
        \draw[very thick,->] (5,0) -- (5,5.5) node[anchor=south west, yshift=-1cm] {Evidence};

        \filldraw[draw=black,fill=lightgray, fill opacity=0.2] (4.5,0) rectangle (5.5,4) ;
        \draw[thick,<->] (4.5,-0.5) -- (5.5,-0.5) node[anchor=north east] {$\mathcal{D}_T$};
        \node at (6.5,3.5) {$p(\mathcal{D}|M_1)$};

        \filldraw[draw=black,fill=blue, fill opacity=0.2] (1.5,0) rectangle (8.5,1) ;
        \node at (2.2,1.3) {$p(\mathcal{D}|M_2)$};

    \end{tikzpicture}
}
}
    \caption{Bayesian model selection adapted from~\cite{deisenroth2020mathematics}.}
    \label{fig_bayesian_model_selection}
\end{figure}
Before we formalize the concept of model selection in the context of~\ac{tl} we will provide an intuition behind (Bayesian) model selection in a broader sense.
In~\cite{Bishop2006}, it is argued that a model selection (outside the context of~\ac{tl}) approach should find a trade-off between model complexity and the fit for the data.
This trade-off is visualized from a Bayesian perspective in~\FIG{fig_bayesian_model_selection}.
On the horizontal axis, the space of all possible datasets is given.
The evidence of a model for a given dataset $\mathcal{D}_T$ is given on the vertical axis.
In this case, consider that model $M_2$ is a larger model with more parameters than model $M_1$ and, therefore, can express a larger number of datasets.
We can see that with the model evidence $p(\mathcal{D}|M_\indexSourceModel)$ we would favor the simpler model for dataset $\mathcal{D}_T$ through the Bayesian perspective.
The concept that a Bayesian perspective on model selection favors the simpler model is also known as \textit{Occam's razor}.

The general concept of model selection is also valid in the context of~\ac{tl}.
We aim to find a source model, from $m \in \{1,\ldots,M\}$ source models with $M \in \mathbb{N}_{>1}$, that explains the limited target data $T$ best.
Consider two tasks $\task_1 = \{ \outputFeatureSpace,~P_1(\setOutputFeatures_1 \mid \setInputFeatures_1)\}$ and $\task_2 = \{ \outputFeatureSpace,~P_2(\setOutputFeatures_2 \mid \setInputFeatures_2)\}$, where the tasks $\task_1,\task_2 \in \allTasks$ and $\allTasks$ is the set of all possible tasks.
The sets $\setOutputFeatures_m$ and $\setInputFeatures_m$ are from the response space $\outputFeatureSpace$ and feature space $\inputFeatureSpace$.
By defining a similarity measure $\similarityMeasure$ with $\similarityMeasure \colon \allTasks \times \allTasks \to \realNumberPos$, the mapping into a scalar allows making quantitative statements.
For instance, given two source tasks $\task_1, \task_2$ and a target task $\task_T$ if $\similarityMeasure\big(\task_1,\task_T\big) > \similarityMeasure\big(\task_2,\task_T\big)$ then $\task_1$ is more similar to the target $\task_T$ compared to $\task_2$, which means, that a high value implicates a high similarity.
Respectively, we define dissimilarity by the inverse of a similarity measure.

Now the question arises what (similarity) measure and what kind of data should be considered to select a source model from a model hub for a specific target.
One choice would be to measure similarity solely based on the input feature space.
However, the input feature space contains limited information on the expected power generation, the response variable, in renewable power generation.
For example, different amounts of energy will be produced with the same radiation for different solar modules.
Consequently, we need to take the response variable into account.

\subsubsection{Evidence or Marginal Likelihood:} The authors of~\cite{You2021LogMe} utilize the marginal likelihood or evidence as a similarity measure $\similarityMeasure$.
For that purpose, the final layer of a (source) neural network $\task_m$ is replaced by a~\ac{blr}, where the priors $\alpha$ and $\beta$ of this model are optimized through empirical Bayes~\cite{You2021Hub,Bishop2006} on limited target data.
In this way, the source model acts as a feature extractor.
The marginal likelihood is then given by
\begin{align}\label{eq_marginal_likelihood}
   \similarityMeasure(\task_T,\task_m) &=  \ln p(\outputFeatures|\alpha,\beta)  \nonumber \\
        &= \frac{D}{2}\ln \alpha + \frac{N}{2} \ln \beta - E(\mathbf{m}_N) - \frac{1}{2}\ln{|\mathbf{S}^{-1}_N| - \frac{N}{2}\ln{2\pi}},
\end{align}
where $D$ is the number of features, e. g., defined by the dimension of the second last layer of the neural network $m$ with $N$ samples and $E(\mathbf{m}_n) = \frac{\beta}{2} \cdot ||\outputFeatures - \inputFeatures \mathbf{m}_n|| + \frac{\alpha}{2} \mathbf{m}_n^T \mathbf{m}_n$. 
This way, we consider features extracted from the source neural network of task $\task_m$ and the response feature from the target $\task_T$.
If we do this for each source model $m$, we can calculate the marginal likelihood of each source model on the target to calculate $\similarityMeasure(\task_m, \task_T)$.
We then select the model with the most extensive evidence as the appropriate source model.
We repeat this for each dimension for multivariate problems and average the results~\cite{You2021LogMe}.

While this approach is theoretically appealing and generalizes for a broad number of problems, it has one drawback in the context of \ac{itl}: It is not considering already learned weights from the final layer of a model.
This consideration is essential, as we often do not need to remove the final layer to assure compatibility between a source and a target task in renewable energies.
At the same time, a pre-trained layer is usually beneficial in contrast to a randomly initialized layer.

\subsubsection{Normalized Root Mean-squared Error:}
Respectively, we propose to directly measure the similarity through the~\ac{nrmse} based on the pre-trained layer of a source model by
\begin{equation}\label{eq_nrmse}
    \similarityMeasure(\task_T,\task_m)^{-1} = \text{nRMSE} = \sqrt{\frac{1}{N}\sum_{i=1}^{i=N}{(y_i^{(T)}-\hat{y}_i^{(m)})^2}},
\end{equation}
where $y_i^{(T)}$ is the $i$-th normalized response from the target and $\hat{y}_i^{(m)}$ is the prediction from source model $m$ on the target.
Note that in the context of renewable power forecasts, we normalize the response $y_i$ by the nominal power to assure comparability of the error for different parks.
We can directly measure how well a source model performs on the available target data to measure the similarity $\similarityMeasure(\task_m, \task_T)$.
Consequently, we can select the source model with a lower~\ac{nrmse} on, e.g., a validation error from the target data.
\subsection{Adaptation Strategies for Inductive Transfer Learning}
\label{sec_itl_selection_method_adaption}
\TBL{tbl_model_selection_overview} overviews all $18$ combinations of models and adaptation strategies.
As a simple~\ac{tl} model, we consider an \textit{online} update of the posterior of the~\ac{belm}.
Therefore, the posterior from a source model acts as a prior for the target.
Additionally, we evaluate \textit{directly} applying a selected source model on the target without any adaptation of a source model's parameter.

We also consider two standard fine-tuning methods from the field of computer vision.
The first one is \textit{weight decay} which penalizes the deviation of weights from zero and \textit{weight decay source} which penalizes a deviation from the source model's weights.
Additionally, we examine \textit{Bayesian tunning} as introduced in~\cite{You2021Hub}.

\begin{table*}[tb]
    \centering
    \caption{Overview of different combinations for models, selections, and adaptations. RM abbreviates the RMSE selection strategy, EV the selection through evidence, DI stands for directly applying the model, WD for fine-tuning through weight decay regarding the origin, WDS for a fine-tuning through weight decay regarding the source parameters, BT for fine-tuning with Bayesian tuning.}\label{tbl_model_selection_overview}
    \begin{tabular}{l|l|l|l}
        \textbf{Model}   & \textbf{Selection} & \textbf{Adaptation}            & \textbf{Abbreviation}  \\
        \textbf{Type}    & \textbf{Strategy}  & \textbf{Strategy}            &               \\
        \hline
        MLP/TCN & RMSE~[ours]      & direct~[ours]             & MLP-/TCN-RM-DI   \\
        MLP/TCN & RMSE~[ours]      & weight decay~\cite{Li2020}        & MLP-/TCN-RM-WD   \\
        MLP/TCN & RMSE~[ours]      & weight decay source~\cite{Li2018} & MLP-/TCN-RM-WDS  \\
        MLP/TCN & EVIDENCE~\cite{You2021Hub}  & direct~[ours]             & MLP-/TCN-EV-DI   \\
        MLP/TCN & EVIDENCE~\cite{You2021Hub}  & direct linear [ours]      & MLP-/TCN-EV-DILI \\
        MLP/TCN & EVIDENCE~\cite{You2021Hub}  & weight decay~\cite{Li2020}         & MLP-/TCN-EV-WD   \\
        MLP/TCN & EVIDENCE~\cite{You2021Hub}  & weight decay source~\cite{Li2018} & MLP-/TCN-EV-WDS  \\
        MLP/TCN & EVIDENCE~\cite{You2021Hub}  & Bayesian tuning~\cite{You2021Hub}     & MLP-/TCN-EV-BT   \\
        BELM     & RMSE~[ours]      & Online~[ours]            & BELM-RM       \\
        BELM     & EVIDENCE~\cite{You2021Hub}  & Online~[ours]            & BELM-EV       \\
    \end{tabular}
\end{table*}

The three latter adaption strategies can be considered as a type of regularization.
In general, this means that we add an additional penalty term $L_{pen}$ to the loss function $L_{task}$ of a task through
\begin{equation}
    L = L_{task} + \lambda \cdot L_{pen}, 
\end{equation}
where $\lambda \in \mathcal{R}$ is a hyper-parameter for the amount of regularization that is typically selected by hyper-parameter optimization.
$L_{task}$ is given by
\begin{equation}
    L_{task} = \frac{1}{\nSamples}\sum^\nSamples_{\indexSamples=1}{l(f(\inputFeatureVector_\indexSamples, \boldsymbol{\theta}),\outputFeatureVector_\indexSamples)},
\end{equation}
where $\parametersLinModel \in \realNumber^p$ and $p\in\naturalNumberPos$ is a vector of the parameters we update, $\inputFeatureVector_\indexSamples$ is the $\indexSamples$-th input vector with $\indexSamples \in \nSamples$ and $\nSamples\in\naturalNumberPos$, and $\outputFeatureVector_\indexSamples$ is the respective response.
For simplicity, we consider a uni-variate response here.
For weight decay (WD) with respect to the origin~\cite{Li2020}, $L_{pen}$ is then given by
\begin{equation}
    L_{WD} = \frac{1}{2}||\boldsymbol{\theta}||^2_2
\end{equation}
To penalizes a deviation from the source model, \cite{Li2018} proposes a weight decay w.r.t. to source weights (WDS) given by
\begin{equation}
    L_{WDS}  = \frac{1}{2}||\boldsymbol{\theta}-\boldsymbol{\theta}^0||^2_2,
\end{equation}
where $\parametersLinModel^0 \in \realNumber^p$ is the vector of parameters from the source model before fine-tunning.
Finally, in Bayesian tunning $L_{pen}$ is given by~\cite{You2021Hub}:
\begin{equation}
    L_{Bayesian} = \frac{1}{\nSamples}\sum^{\nSamples}_{\indexSamples=1}{\frac{1}{K}\sum^{K}_{k=1}}{(\frac{1}{M}\sum^{M}_{\indexSourceModel=1}{\inputFeatureVector_{\indexSourceModel,\indexSamples}^T\parametersLinModel_{\indexSourceModel,k}-\inputFeatureVector_{t,\indexSamples}^T\parametersLinModel_{t,k}})^2},
\end{equation}
where $\indexSamples \in \nSamples$ is the $\indexSamples$-th data sample, $\indexSourceModel$ is the $\indexSourceModel$-th source model adapted with~\ac{blr}, $k$ is the $k$-th dimension of the response for example for different forecast horizons.
$\inputFeatureVector_{\indexSourceModel,k}$ are features extracted from the $\indexSourceModel$-th source model, $\inputFeatureVector_{t,\indexSamples}$ are the respective features extracted from the target model $t$.
$\parametersLinModel_{k,c}$ and $\parametersLinModel_{t,c}$ are the mean vectors calculated by the~\ac{blr}.

\subsection{Model Combination for Inductive Transfer Learning}\label{sec_method_model_combination}
We discussed the model selection and adaptation strategies for a single source model for a target.
However, a single model might be prone to overfitting with limited data.
Combining source models through an ensemble reduces this risk.

\subsubsection{Bayesian Model Averaging} 
We extend the concept of~\cite{You2021LogMe} so that instead of choosing a single model based on the evidence, we combine models adapted through~\ac{blr} by~\ac{bma}.
\Ac{bma} is theoretically appealing as it considers the predictive posterior~\cite{Hoeting1999} and therefore considers the uncertainty of a model through
\begin{equation}\label{eq_bayesian_model_averaging}
    p(\outputFeatures_{T*}|\outputFeatures_T) = \sum_{i=1}^{i=M}p(\outputFeatures_{T*}|\outputFeatures_T, \parametersLinModel_{M_i}) p(\parametersLinModel_{M_i}|\outputFeatures_T).
\end{equation}
The prior probability $p(\parametersLinModel_{M_i}|\outputFeatures_T)$ encodes our prior belief that a model $M_i$ is more likely to another one for the target data set.
For simplicity, we consider an equal prior for all source models.
Note that we have omitted the input here to simplify notations.
$p(\outputFeatures_{T*}|\outputFeatures_T, \parametersLinModel_{M_i})$ is the predictive posterior of a model $M_i$ given by~\EQ{eq_lin_predictive_posterior}, where, e.g, the model results from the proposed~\textit{direct linear} adaptation strategy.


\subsubsection{Coopetitive Soft Gating Ensemble:} We also propose to utilize the~\ac{csge} for model combination in the context of~\ac{itl}.
The~\ac{csge} was initially introduced for renewable power forecast in~\cite{GS18}.
The idea of the CSGE is to link the weights to the ensemble members' performance, i.e., good source models are weighted stronger than weaker ones.
The~\ac{csge} characterizes the overall weight of a source model using three aspects:
\begin{itemize}
    \item The \textit{global weight} is defined by how well a source model performs with the available training data on the target task.
    \item The \textit{local weight} is defined by how well a source model performs on the target tasks for different areas in the feature space. For example, in the case of wind, one model might perform well for low wind speeds, while another source model might perform well for larger wind speeds on the target.
    \item The \textit{forecast horizon-dependent weight} is defined by how well a source model performs for different lead times on the target task. Here, between $24$ and $48$ hours into the future.
\end{itemize}
\begin{figure*}[tb]
    \centering
    {
    \large
    \begin{center} 
    \resizebox{0.75\textwidth}{!}{%
        \tikzstyle{block} = [rectangle, rounded corners, text centered, draw=black]
        \tikzstyle{clear} = [rectangle, rounded corners, text centered]
        \tikzstyle{arrow} = [thick,->,>=stealth]
        \tikzstyle{connect} = [thick,-,>=stealth]
            \begin{tikzpicture}[node distance=1cm]
            
            \node [block, text width=2cm, text height=6cm,] (cw) {};
            \node [block, text width=2cm, text height=6cm, right of=cw, xshift=2.5cm] (nw) {};
            \node [block, text width=2cm, text height=6cm, right of=nw, xshift=4cm] (ap) {};
            \node[text width=2.6cm] at (0.3,0.0) {\textbf{Compute Weights}};
            \node[text width=2.6cm] at (3.75,0.0) {\textbf{Normalize Weights}};
            \node[text width=2.3cm] at (8.6,0.0) {\textbf{Aggregate Forecasts}};

            \node [block, text width=1.5cm, text height=0.25cm, above of=cw, opacity=0.75, yshift=1.75cm, dashed] (global) {Global};
            \node [block, text width=1.5cm, text height=0.25cm, below of=global, opacity=0.75, yshift=-0.5cm, dashed] (local) {Local};
            \node [block, text width=1.5cm, text height=0.25cm, below of=local, opacity=0.75, yshift=-1.8cm, dashed] (fh) {Forecast Horizon};
            
            \node [block, text width=3cm, text height=0.5cm, left of=cw, xshift=-3cm, yshift=2.7cm] (sm1) {Source Model (1)};
            \node [block, text width=3cm, text height=0.5cm, below of=sm1, yshift=-0.5cm] (sm2) {Source Model (2)};
            \node [clear, text width=3cm, text height=0.5cm, below of=sm2, yshift=-0.5cm] (sdots) {...};
            \node [block, text width=3cm, text height=0.5cm, below of=sdots, yshift=-0.5cm] (smj) {Source Model (M)};
            
            \node [clear, text width=0.5cm, text height=0.5cm, left of=cw, xshift=-6.5cm] (x) {};
            \node [clear, text width=0.0cm, text height=0.0cm, right of=x] (xa) {};
            \draw [connect] (x.east) -- node[anchor=north, yshift=0.6cm, xshift=-0.25cm] {$\inputFeatureVector_{t + k | t}$} (xa.east);
            \draw [arrow] (xa.east) |- node[] {} (sm1);
            \draw [arrow] (xa.east) |- node[] {} (sm2);
            \draw [arrow] (xa.east) |- node[] {} (smj);
            
            \node [clear, text width=0.5cm, text height=0.5cm, right of=ap, xshift=2cm] (y) {};
            \draw [arrow] (ap.east) -- node[anchor=north, yshift=0.75cm] {$\bar{\outputPredictionVector}_{t+k|t}$} (y.west);
            
            \node [clear, right of=sm1, xshift=3cm, text width=2cm] (cw1) {};
            \node [clear, right of=sm2, xshift=3cm, text width=2cm] (cw2) {};
            \node [clear, right of=smj, xshift=3cm, text width=2cm] (cwj) {};
            \draw [arrow] (sm1.east) -- node[anchor=north, yshift=0.8cm] {$\outputPredictionVector^{(1)}_{t+k|t}$} (cw1.west);
            \draw [arrow] (sm2.east) -- node[anchor=north, yshift=0.8cm] {$\outputPredictionVector^{(2)}_{t+k|t}$} (cw2.west);
            \draw [arrow] (smj.east) -- node[anchor=north, yshift=0.8cm] {$\outputPredictionVector^{(\numberSourceModels)}_{t+k|t}$} (cwj.west);
            
            \node [clear, right of=cw1, xshift=2.5cm, text width=2cm] (nw1) {};
            \node [clear, right of=cw2, xshift=2.5cm, text width=2cm] (nw2) {};
            \node [clear, right of=cwj, xshift=2.5cm, text width=2cm] (nwj) {};
            \draw [arrow] (cw1.east) -- node[anchor=north, yshift=0.8cm] {$\csgeWeightUnnormalized^{(1)}_{t+k|t}$} (nw1.west);
            \draw [arrow] (cw2.east) -- node[anchor=north, yshift=0.8cm] {$\csgeWeightUnnormalized^{(2)}_{t+k|t}$} (nw2.west);
            \draw [arrow] (cwj.east) -- node[anchor=north, yshift=0.8cm] {$\csgeWeightUnnormalized^{(\numberSourceModels)}_{t+k|t}$} (nwj.west);
            
            \node [clear, right of=nw1, xshift=4cm, text width=2cm] (ap1) {};
            \node [clear, right of=nw2, xshift=4cm, text width=2cm] (ap2) {};
            \node [clear, right of=nwj, xshift=4cm, text width=2cm] (apj) {};
            \draw [arrow] (nw1.east) -- node[anchor=north, yshift=0.8cm] {$\csgeWeightNormalized^{(1)}_{t+k|t} \cdot \outputPredictionVector^{(1)}_{t+k|t}$} (ap1.west);
            \draw [arrow] (nw2.east) -- node[anchor=north, yshift=0.8cm] {$\csgeWeightNormalized^{(2)}_{t+k|t} \cdot \outputPredictionVector^{(2)}_{t+k|t}$} (ap2.west);
            \draw [arrow] (nwj.east) -- node[anchor=north, yshift=0.8cm] {$\csgeWeightNormalized^{(\numberSourceModels)}_{t+k|t} \cdot \outputPredictionVector^{(\numberSourceModels)}_{t+k|t}$} (apj.west);
            
            \node [clear, right of=sm1, xshift=1.25cm] (sc1) {};
            \node [clear, right of=cw1, xshift=2.5cm, yshift=-0.5cm, text width=2cm] (nw1h) {};
            \draw [arrow, opacity=0.5] (sc1.center) |- node[] {} (nw1h.west);
            \node [clear, right of=sm2, xshift=1.25cm] (sc2) {};
            \node [clear, right of=cw2, xshift=2.5cm, yshift=-0.5cm, text width=2cm] (nw2h) {};
            \draw [arrow, opacity=0.5] (sc2.center) |- node[] {} (nw2h.west);
            \node [clear, right of=smj, xshift=1.25cm] (scj) {};
            \node [clear, right of=cwj, xshift=2.5cm, yshift=-0.5cm, text width=2cm] (nwjh) {};
            \draw [arrow, opacity=0.5] (scj.center) |- node[] {} (nwjh.west);
            
            \node [clear, text width=3cm, text height=0.5cm, right of=sdots, xshift=1.25cm] (ydots) {...};
            \node [clear, text width=3cm, text height=0.5cm, right of=sdots, xshift=4.75cm] (wdots) {...};
            \node [clear, text width=3cm, text height=0.5cm, right of=sdots, xshift=9cm] (wydots) {...};

            \end{tikzpicture}
    }
    \end{center}
}
    \caption{The architecture of the CSGE. The source models' predictions $\outputPredictionVector^{(\indexSourceModel)}_{t + k | t}$ for the input $\inputFeatureVector$ are passed to the CSGE. The ensemble member's weights are given by aggregating the respective global-and local- and forecast horizon-dependent weights. The weights are normalized. The source models' predictions are weighted and aggregated in the final step.}
  
    \label{fig:coop_int_fusion:csge_overview_graphic}
\end{figure*}
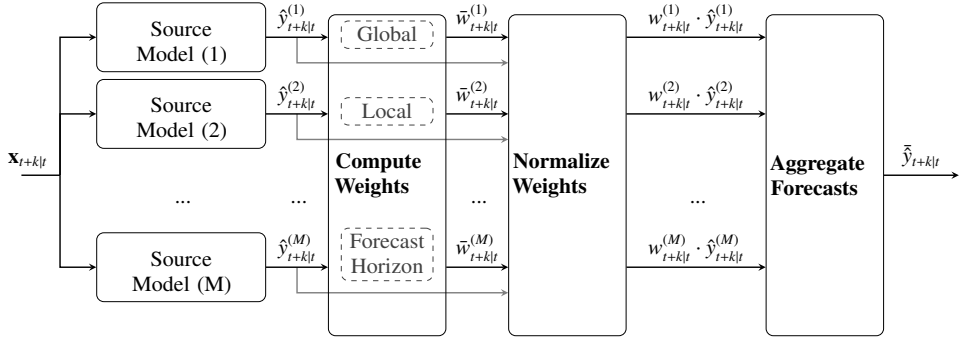

Fig.~\ref{fig:coop_int_fusion:csge_overview_graphic} provides an overview of the~\ac{csge}.
The CSGE includes $\numberSourceModels$ ensemble members, with $\indexSourceModel \in \{1, \ldots, \numberSourceModels\}$.
Each of these ensemble members is a source model with a predictive function $f_\indexSourceModel$. 
Each source model forecasts an univariate estimate $\hat{y}^{(\indexSourceModel)}_{t + k | t} \in \mathbb{R}$ for the input $\inputFeatureVector_{t + k | t} \in \mathbb{R}^{\dimFeature}$ of a target task $\indexTargetTask$, we omit the subscript $\indexTargetTask$ for reasons of clarity and comprehensibility.
Let $\dimFeature$ be the dimension of the input feature vector $\inputFeatureVector$.
Then, $k$ denotes the forecast horizon, denoted by the subscript, for the forecast origin $t$.
For each prediction of each source model, we compute an aggregated weight $\csgeWeightNormalized^{(\indexSourceModel)}_{t + k | t}$.

The weight incorporating the global $w_g$, local $w_l$, and forecast horizon-dependent weight $w_h$ for a single source model and lead time is given by
\begin{align}
    \label{eq:coop_int_fusion:csge:overallweights}
    \csgeWeightUnnormalized^{(\indexSourceModel)}_{t + k | t} = \csgeWeightNormalized^{(\indexSourceModel)}_g \cdot \csgeWeightNormalized^{(\indexSourceModel,t)}_l \cdot \csgeWeightNormalized^{(\indexSourceModel,k)}_{h},
\end{align}
where $\csgeWeightUnnormalized^{(\indexSourceModel)}_{t + k | t}$ is normalized to sum up to one to calculate $\csgeWeightNormalized^{(\indexSourceModel)}_{t + k | t}$.

To calculate the weights $\csgeWeightNormalized^{(\indexSourceModel)}_{t + k | t}$, we utilize the definition of the inverse similarity measurement $\similarityMeasure^{-1}$ from~\SEC{sec_model_selection_definition} and the coopetitive soft gating principle from~\EQ{eq_soft_gating}.
\begin{align}\label{eq_soft_gating}
    \varsigma^{'}_\eta(\mathbf{\Phi}, \phi) = \frac{\sum_{j=1}^{J} \mathbf{\Phi}_j}{\phi^{\eta} + \epsilon}
\end{align}
By calculating the weighting through the inverse $\similarityMeasure^{-1}$, here the~\ac{nrmse}, we estimate how well a source model performs on the target.
Let us assume that $\mathbf{\Phi} \in \realNumber^{J}$ contains all $J \in \naturalNumberPos$ estimates based on the~\ac{nrmse} and $\phi \in \mathbf{\Phi}$.
Then, $\eta \geq 0$ depicts the amount of exponential weighting and the small constant $\epsilon > 0$ avoids division by zero.
For greater $\eta$, the \ac{csge} tends to work as a gating ensemble, thereby considering only a few source models.
For smaller $\eta$ result in a weighting ensemble.
After calculating all weights from $\mathbf{\Phi}$ through~\EQ{eq_soft_gating}, we normalize the results to sum up to one estimating the final weights $\csgeWeightNormalized^{(\indexSourceModel)}_{t + k | t}$.
This approach is repeated for each of the three weighting aspects as detailed in~\cite{GS18} and~\ref{sec_appendix_model_selection_method}.

\section{Experimental Evaluation}
\label{sec_itl_selection_experiment}

In the following~\SEC{sec_model_selection_doe}, we summarize the experimental setup.
We conduct experiments on six datasets with a total of $667$ parks.
Due to the utilized cross-validation, each park is once a target park.
Thereby, we provide the most extensive study for~\ac{itl} for renewable power forecasts.

We evaluate models through the mean performance rank, calculated across parks within a dataset, to show significant improvements against the \baseline.
In the appendix, we provide additional results concerning the forecast error and comparison to results from~\cite{Schreiber2021}.
\SEC{sec_model_selection_model_selection_experiment} provides the details of our first experiment to answer research questions one and two.
The second experiment in~\SEC{sec_model_selection_model_combination_experiment} details our findings for research question three.

\subsection{Overall Experimental Setup}\label{sec_model_selection_doe}
The preprocessing of the data is aligned with~\cite{VogtSynData2022,Schreiber2021} to assure comparability with the current state of the art.
As source models, we considered a~\ac{belm},~\ac{mlp}, and a~\ac{tcn}.
To have a robust \baseline that generalizes well with a limited amount of data and is known to mitigate the effects of overfitting, we trained a~\ac{gbrt} for each target task identical to~\cite{VogtSynData2022}.

\subsubsection{Datasets:} We conducted all experiments for day-ahead forecasts, between $24$ and $48$ hours into the future.
All datasets, summarized in \TBL{tbl_dataset_summary}, have~\ac{nwp} features as inputs, e.g., wind speed, wind direction, air pressure, or radiation.
For all datasets, we align those weather forecasts with the historical power measurements as the response for day-ahead forecasts.
These input features are weather forecasts from the European center for medium-range weather forecasts (ECMWF) or the Icosahedral Nonhydrostatic-European Union (ICON-EU) weather model.
\begin{table*}[t]
    \centering
    \caption{Overview of the evaluated datasets.}\label{tbl_dataset_summary}
    \begin{tabular}{l|llllll}

        \textbf{Dataset}           & \#\textbf{parks} & \#\textbf{features} & \textbf{train}   & \textbf{mean}    & \textbf{resolution} & \textbf{NWP}   \\
                                   &                  &                     & \textbf{samples} & \textbf{samples} &                     & \textbf{model} \\
        \hline
        \PO~\cite{Schreiber2021}   & 21               & 47                  & 6336             & 8424             & hourly              & ECMWF          \\
        \PS~\cite{VogtSynData2022} & 114              & 20                  & 30385            & 14920            & 15-min              & ICON-EU        \\
        \PR                        & 42               & 25                  & 58052            & 19344            & 15-min              & ICON-EU        \\
        \WO~\cite{Schreiber2021}   & 45               & 13                  & 27724            & 26636            & 15-min              & ECMWF          \\
        \WS~\cite{VogtSynData2022} & 260              & 29                  & 33714            & 16678            & 15-min              & ICON-EU        \\
        \WR                        & 185              & 33                  & 36129            & 12092            & 15-min              & ICON-EU
    \end{tabular}
\end{table*}

In all datasets, we have varying amounts of input features, resolutions, and different numbers of samples for training and testing.
For instance, the \PO has $47$ features, where various manually engineered features take seasonal patterns of the sun into account.
In contrast, these manually engineered features are not included in other datasets.

Also note that four datasets, the \PONS, \WONS, \WSNS, and \PS have already been investigated, see e.g.~\cite{Schreiber2021,VogtSynData2022}.
This is not the case for \WR and \PRNS.
These two datasets are not publicly available.
However, they are the most realistic datasets due to their diversity.
The \WR dataset consists of 99 different nominal capacities, 13 turbine manufacturers, and six hub heights.
All parks are located in Germany.
PV power plants in the \PR dataset have 31 different nominal capacities, ten tilt orientations, and nine azimuth orientations and are also located in Germany.
It is also important to note that forecasting the expected power generation from wind parks is more challenging than for~\ac{pv} parks.
For additional insights on the challenges and the datasets refer to the~\ref{sec_appendix_model_selection_data}.

Each dataset was split through five-fold cross-validation so that \textit{each park is once a target task} and four times a source park.
We trained source models and their hyperparameters on the training and validation data.
We split the training into the four seasons for training target models and
limited the training data to $7,14,30,60$ or $90$ days of training data, respectively.
The presented results are mean values for all tasks and seasons.
This setup assures that results are not biased by seasonality~\cite{Jens2019}.
All input features of all datasets are normalized.
We normalized the historical power by the nominal power to make errors comparable.
We resampled all datasets to have a $15$-minute resolution except the \PO dataset, which we resampled for an hourly resolution due to the low initial resolution.
A predefined test set is given for the \WS and \PS datasets.
In the case of the \WO and \PONS, we used the first year's data as training data and the remaining data as test data, identical to~\cite{Schreiber2021}.
Due to this diversity in the number of historical power measurements for the \PR and \WR datasets, $25\%$ randomly sampled days are considered test data.
As each day is based on an independent day ahead~\ac{nwp} forecasts, no information is leaked from the future to the past~\cite{Gensler2018}.
We use $25\%$ of the remaining days for validation and the rest for training.

\subsubsection{Source Models}\label{sec_itl_experiment_source_models}
As pointed out earlier, due to the weather dependency for renewable power forecasts, the input features of the models are themselves forecasts from the ~\ac{nwp} model.
Respectively, we can directly utilize those to train, e.g., an~\ac{mlp} to forecast the expected power of the next day.
To optimize hyperparameters of those models, we utilize a tree-structured Parzen sampler for $200$ samples on the validation data.
Details of the chosen hyperparameters are provided in the~\ref{sec_appendix_model_selection_experiment}.

In total, we train four kinds of models.
The trained~\ac{belm} is particularly interesting as a source model because it can directly measure similarity by the evidence and has a linear increase in time for updating the model.
We train an~\ac{mlp} as it is common practice in the renewable power forecast industry~\cite{Schreiber2021}.
To account for cyclic behavior within the forecast, we also train a~\ac{tcn} architecture, similar to~\cite{Schreiber2021}.
To have a strong \baseline that generalizes well we trained a~\ac{gbrt}~\cite{VogtSynData2022}.

\subsubsection{Evaluation Method}
We calculated the error on the test dataset through the~\ac{nrmse} through~\EQ{eq_nrmse} for all combinations of seasons and available training data.
For a given dataset, season, and the number of days of training data, we calculated the mean performance rank based on the~\ac{nrmse}.
We test for a significant improvement compared to the \baseline by the Wilcoxon test ($\alpha = 0.01$) across all parks within a dataset.

\begin{table*}[!ht]
    \centering
    \caption{Rank summary for all models, selections, and adaptation strategies on the PV datasets, cf.~\TBL{tbl_model_selection_overview}. Only those within the top four ranks for a dataset are included. GBRT is the \baseline and all models are tested if the forecasts error is significantly ($\alpha=0,01$) better ($\vee$), worse ($\wedge$), or not significantly different ($\diamond$). We conduct this hypothesis test for all parks within a dataset for the given number of days training data. The colors denote the respective rank. Blue indicates a smaller (better) rank and red a higher (worse) rank.}\label{tbl_model_selection_rank_pv}
    \includegraphics[width=0.92\textwidth,trim=0.1cm 0.09cm 0.0cm 0cm, clip]{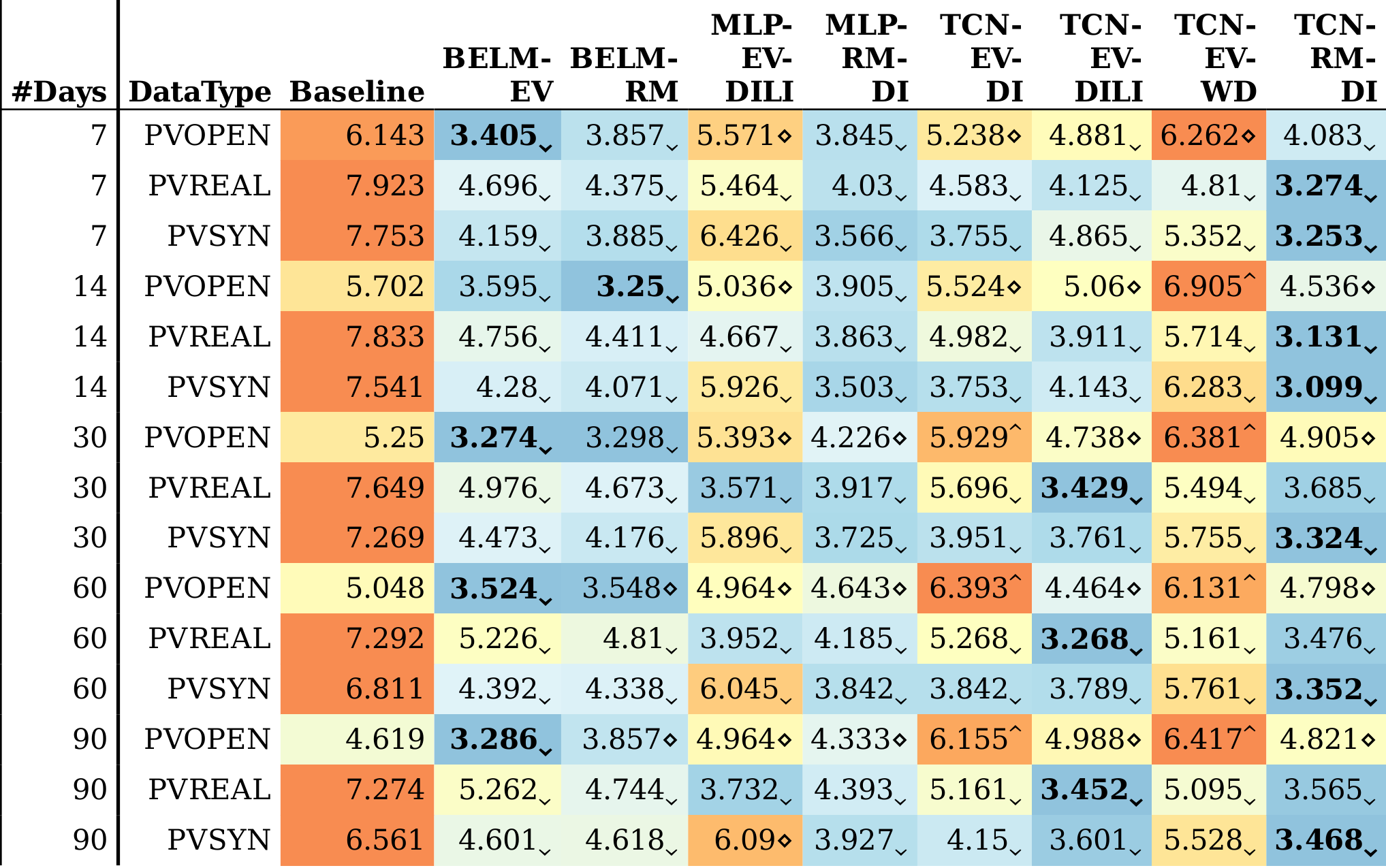}
\end{table*}
\begin{table*}[!ht]
    \centering
    \caption{Rank summary for all source models, selections, and adaptation strategies on the wind datasets. Cf.~\TBLs{tbl_model_selection_overview} and~\ref{tbl_model_selection_rank_pv}.}\label{tbl_model_selection_rank_wind}
    \includegraphics[width=0.92\textwidth,trim=0.1cm 0.09cm 0.0cm 0cm, clip]{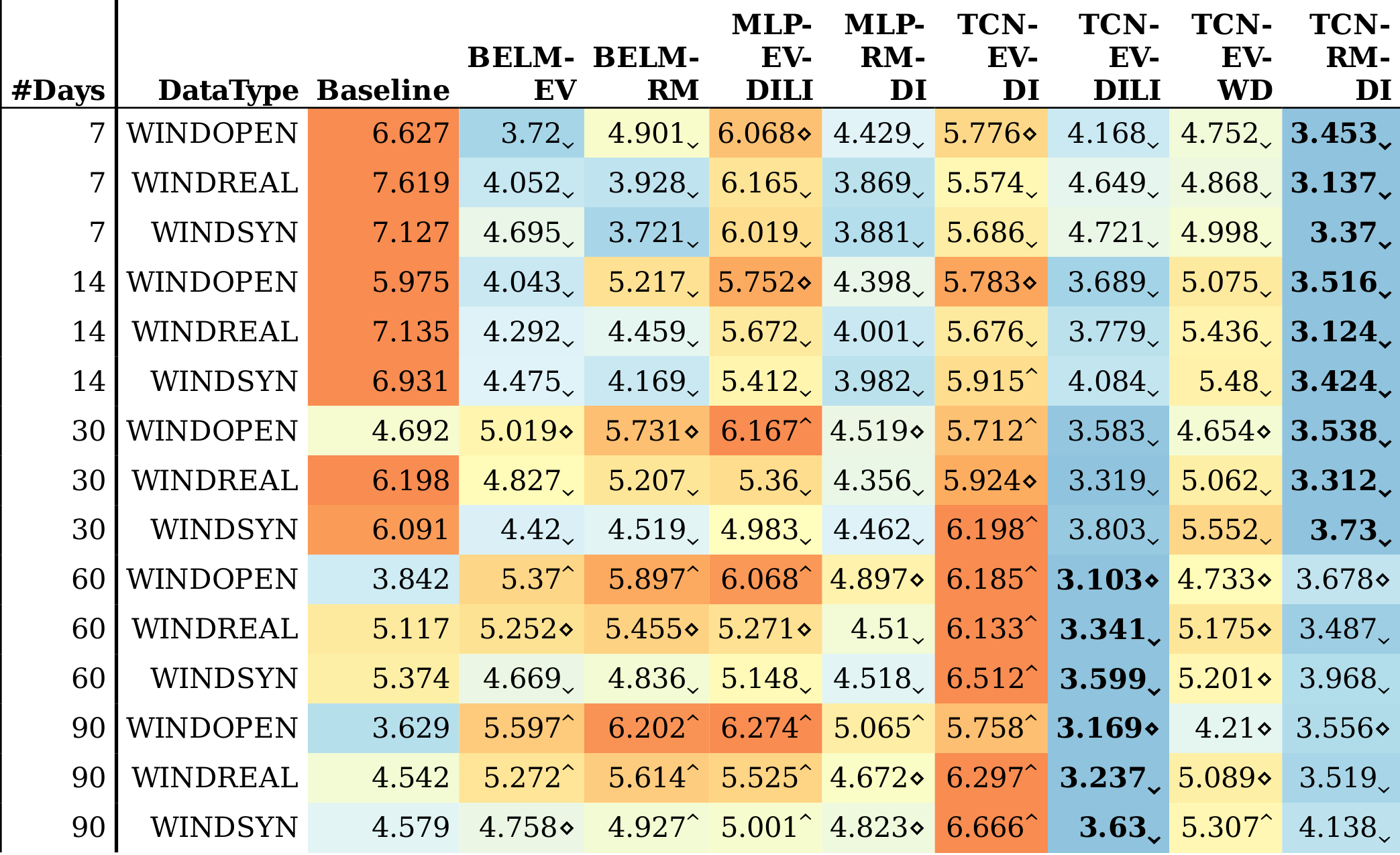}
\end{table*}
\begin{table*}[!ht]
    \centering
    \caption{Rank summary of ensembles on the PV datasets. The best model, the TCN-EV-DILI, from the experiment in ~\SEC{sec_model_selection_model_selection_experiment} is the baseline. Cf.~\TBLs{tbl_model_selection_overview} and~\ref{tbl_model_selection_rank_pv}.}\label{tbl_model_selection_rank_pv_ensemble}
    \includegraphics[width=0.92\textwidth,trim=0.1cm 0.09cm 0.0cm 0cm, clip]{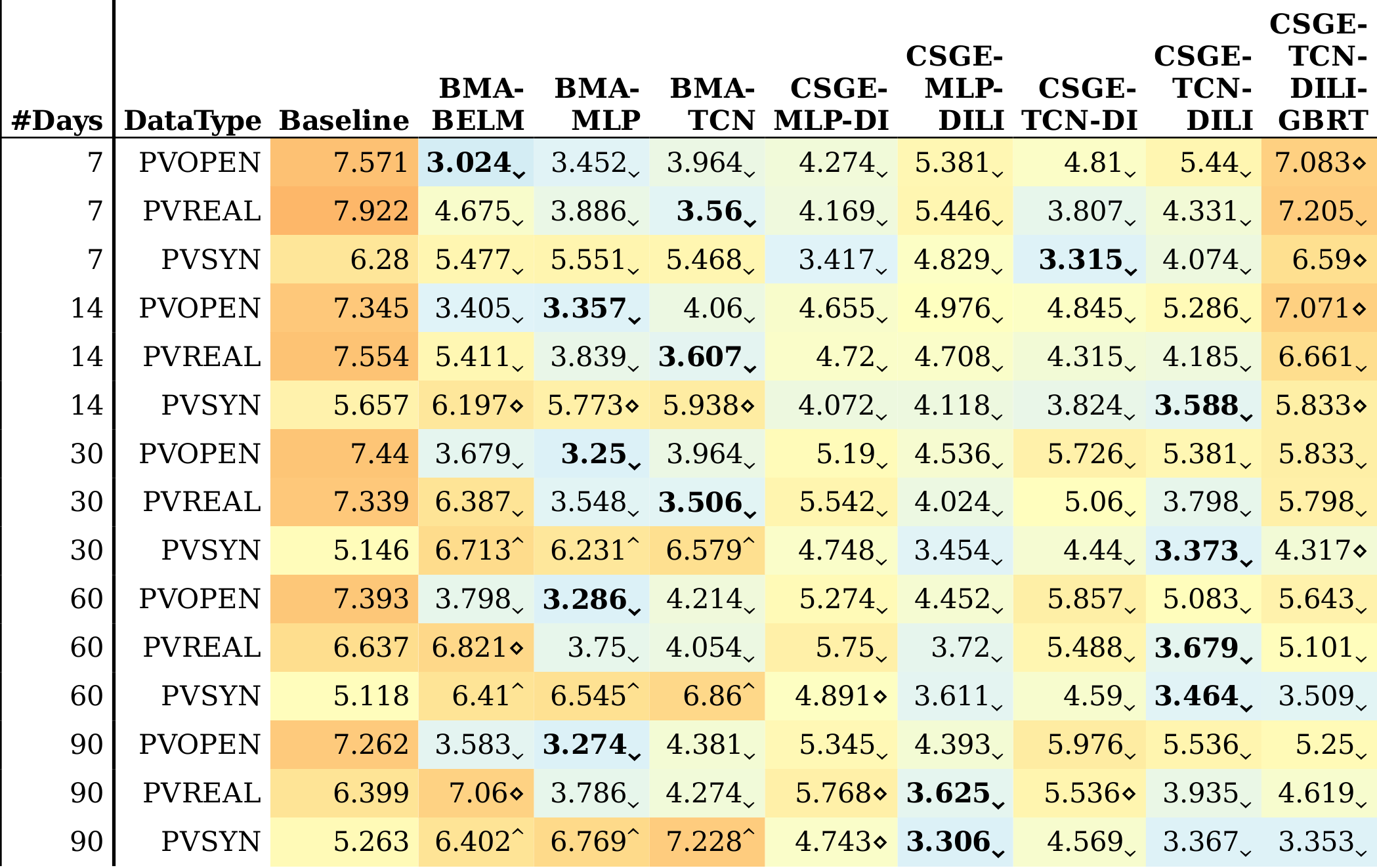}
\end{table*}
\begin{table*}[!ht]
    \centering
    \caption{Rank summary of ensembles on the wind datasets. Cf.~\TBL{tbl_model_selection_rank_pv_ensemble}.}\label{tbl_model_selection_rank_wind_ensemble}
    \includegraphics[width=0.92\textwidth,trim=0.1cm 0.09cm 0.0cm 0cm, clip]{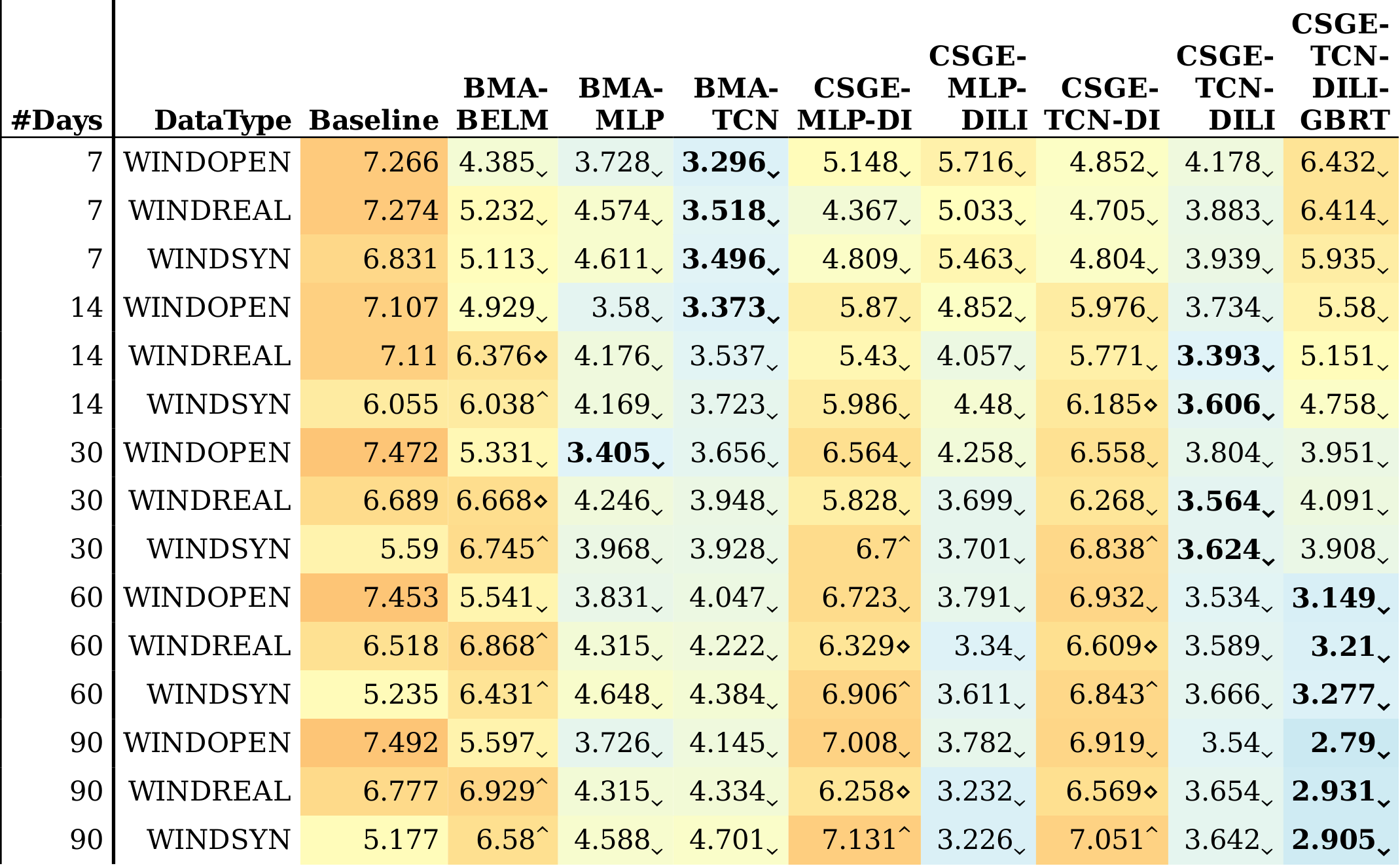}
\end{table*}
\begin{table*}[!ht]
    \centering
    \caption{Mean nRMSE of ensembles on the PV datasets. The best model, the TCN-EV-DILI, from the experiment in ~\SEC{sec_model_selection_model_selection_experiment} is the baseline. Cf.~\TBLs{tbl_model_selection_overview} and~\ref{tbl_model_selection_rank_pv}.}\label{tbl_model_selection_rmse_pv_ensemble}
    \includegraphics[width=0.92\textwidth,trim=0.1cm 0.09cm 0.0cm 0cm, clip]{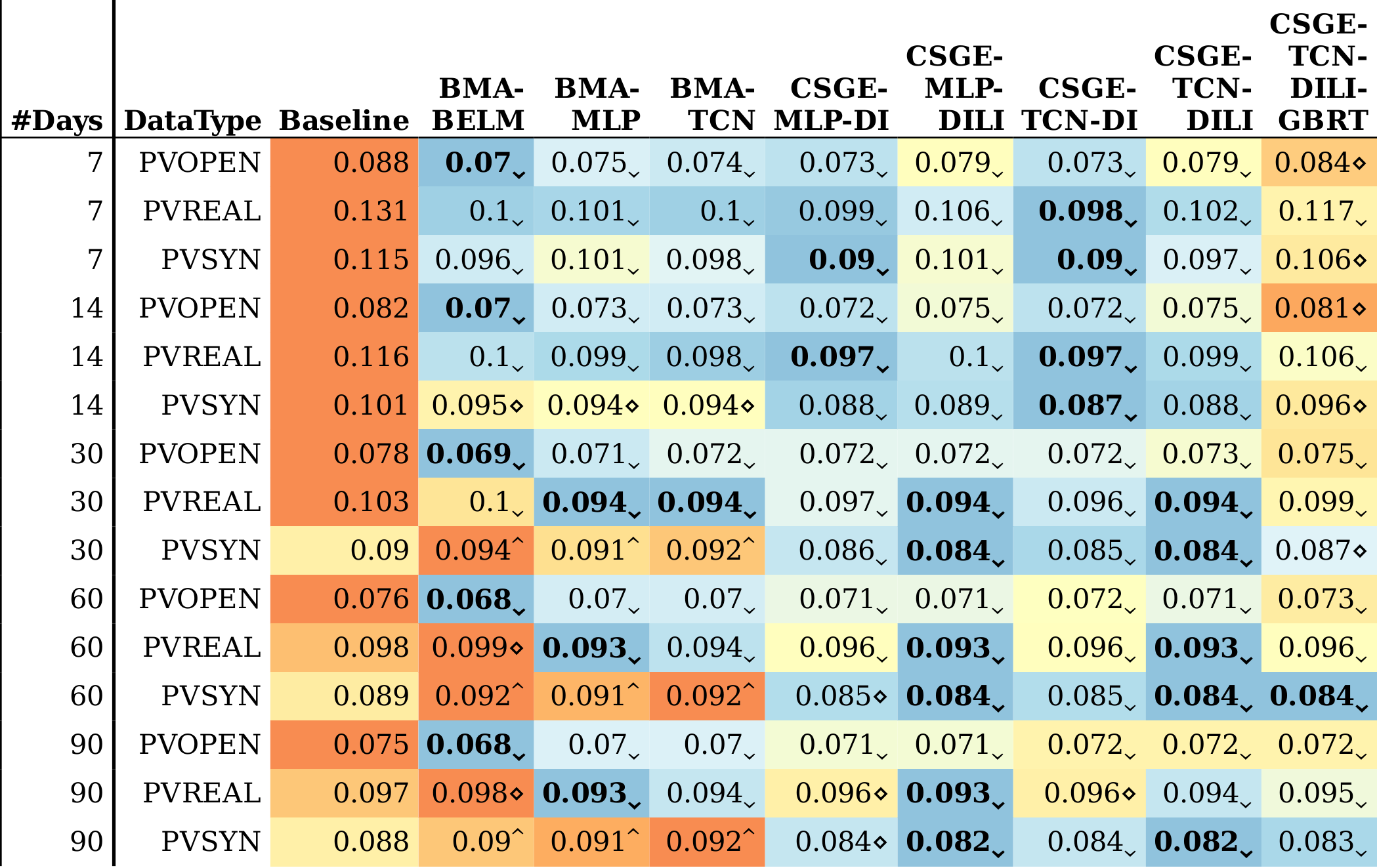}
\end{table*}
\begin{table*}[!ht]
    \centering
    \caption{Mean nRMSE of ensembles on the wind datasets. Cf.~\TBL{tbl_model_selection_rmse_pv_ensemble}.}\label{tbl_model_selection_rmse_wind_ensemble}
    \includegraphics[width=0.92\textwidth,trim=0.1cm 0.09cm 0.0cm 0cm, clip]{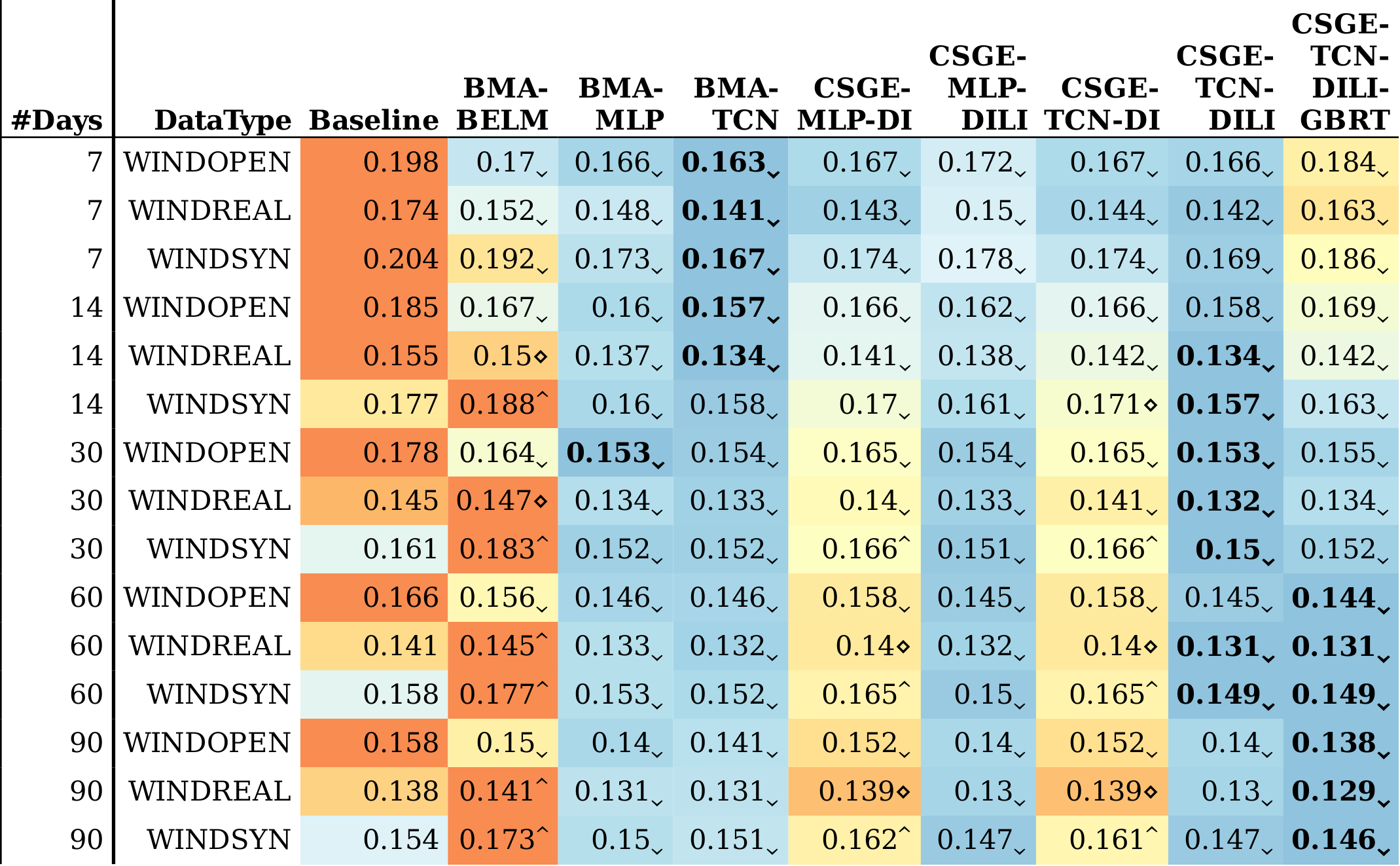}
\end{table*}

\setcounter{question}{0}
\subsection{Experiment on Model Selection and Model Adaptation}\label{sec_model_selection_model_selection_experiment}
This section conducts an experiment to answer research questions one and two simultaneously as a model selection technique can only be evaluated after the adaptation:
\begin{question}
    What is an appropriate similarity measure for model selection for a new target park from a model hub with pre-trained models?
\end{question}
\begin{question}
    What is the best adaptation strategy once a model is selected?
\end{question}

\subsubsection{Findings Questions 1 \& 2}
Model selection and adaptation strategies highly influence each other. With limited training data, between $7$ and $30$ days, selecting a model based on the forecast error with no adaptation has one of the best results. Replacing the final layer by a~\ac{blr} is superior with additional data.
None of the fine-tuning methods is among the best models.

\subsubsection{Experimental Setup}
The source models are those detailed in~\SEC{sec_itl_experiment_source_models}.
As adaptation strategies, we consider those mentioned in~\SEC{sec_itl_selection_method_adaption}.
For fine-tuning, we train for a single epoch and optimize hyperparameters through grid search on $30\%$ of the available target data.

For the weight decay adaptation, we optimize seven logarithmically spaced learning rates between $10^{-1}$ and $10^{-4}$, similar to~\cite{You2021LogMe}.
We take seven grid points for the amount of penalty $\lambda$ in the logarithmic space between $10^{-6}$ to $10^{-3}$, similar to~\cite{You2021LogMe}.
We use the same learning rate for the Bayesian tunning and weight decay source.
The amount of penalty $\lambda$ for the $L_{\text{Bayesian}}$ loss is one of $[0.1, 0.25, 0.5, 1, 2, 4, 8]$.
$\lambda$ is one of $[1, 0.1]$ for the weight decay source.
Note that during the training, we shuffle the data.
For other approaches, hyperparameter optimization is not required.

\subsubsection{Detailed Findings}

Results of the best techniques are summarized in~\TBL{tbl_model_selection_rank_pv} and \ref{tbl_model_selection_rank_wind}.
We only show models appearing at least once within the top four ranks for a dataset.
The~\ac{belm} is among the best models and outperforms the \baseline up to $30$ days of training data.
With less or equal than $14$ days of training data, it seems beneficial to directly utilize a model without any model adaptation.
Starting with $30$ days of training data utilizing a~\ac{blr} trained on extracted features from the source model and the historical power from the target is beneficial, especially for \WR and \PRNS.
This effect occurs as features extracted from a single model from a single prediction task do not generalize well enough for other parks.
Therefore a sufficient amount of data is required to train the~\ac{blr} to compensate for differences between a source and a target park.

These two observations let us conclude with two critical considerations for real-world applications.
First, due to the learning procedure of the gradient descent, there is a high risk of catastrophic forgetting that should be avoided for model hubs in safety-critical areas such as renewable power forecasts.
Second, the~\ac{blr} gives rise to optimal training due to the convex optimization problem, which reduces the risk of catastrophic forgetting.
Notably, neither a weight decay nor the Bayesian tuning adaptation strategy is within the best models in the evaluated scenarios.
This observation is surprising as this fine-tuning approach is common in various domains.
However, due to the source model's training on a single park approach, there is a high risk that even the best-selected source model causes catastrophic forgetting as the model is too specific.
For instance, catastrophic forgetting may appear due to slightly different weather conditions or physical characteristics such as the turbine type.

An additional study given in the~\ref{sec_appendix_exp_model_selection_and_adaption} shows that for fine-tuning techniques, the evidence selection strategy is superior for the~\ac{tcn} model regardless of the adaptation strategy.
For the~\ac{mlp}, a selection through the~\ac{nrmse} is preferable.
Most likely, the probabilistic approach of the evidence and, therefore, the more comprehensive treatment of similarity better captures the correlations between source and target for the convolutional layers in the~\ac{tcn}.
To update the weights of the final layer of the~\ac{mlp} the~\ac{nrmse} is sufficient.



\subsection{Experiment on Model Combination}\label{sec_model_selection_model_combination_experiment}
In this section, we conduct an experiment to answer research question three.
\begin{question}
    Compared to selecting and adapting a single model, are ensemble strategies beneficial for combining knowledge?
\end{question}

\subsubsection{Findings Research Question 3}
Ensembles improve results from the previous experiment significantly. For more straightforward problems, an approach utilizing~\ac{bma} is preferable. For more complex scenarios, an approach by the~\ac{csge} is superior.

\subsubsection{Experimental Setup}
The best model, the TCN-EV-DILI, from the experiment in ~\SEC{sec_model_selection_model_selection_experiment} is the \baseline.
For the~\ac{bma}, we first update all source models based on available target data as previously described.
For the~\ac{mlp} and~\ac{tcn} source models we replace the final layer through~\ac{blr} model(s), as detailed for the direct linear adaptation.
After this adaptation for the target, each model provides a predictive posterior distribtion according to~\EQ{eq_lin_predictive_posterior} that is combined by~\ac{bma} with~\EQ{eq_bayesian_model_averaging}.
We consider three variants for the~\ac{bma}, one for each of the three source model types.

For the~\ac{csge}, we calculate the global and forecast horizon-dependent error based on the~\ac{nrmse}.
We estimate the local error through a k-nearest neighbor approach.
Therefore, we first reduce the dimension of the feature space through~\ac{pca} to two components.
We consider three neighbors within this reduced features space to estimate the local error in the features space.
The hyperparameter $\eta$ is selected as either one or two through grid-search.
In total, we consider six variants of the~\ac{csge}:
Two for the~\ac{mlp} and~\ac{tcn} model where the source models are not updated for the target (CSGE-MLP-DI/CSGE-TCN-DI), two variants, where the final layer of the~\ac{mlp} and~\ac{tcn} source models are updated through~\ac{blr} (CSGE-MLP-DILI/CSGE-TCN-DILI), and these two variants are extended, where we utilize the~\ac{gbrt} as an additional source model (CSGE-MLP-DILI-GBRT/CSGE-TCN-DILI-GBRT).

\subsubsection{Detailed Findings}

Results are summarized in~\TBLs{tbl_model_selection_rank_pv_ensemble} and~\ref{tbl_model_selection_rank_wind_ensemble}.
For the~\ac{pv} datasets, the best~\ac{csge} variants outperform the \baseline in almost all cases.
At the same time, the~\ac{bma} achieves excellent results for the \PO dataset and the \PR dataset for up to 30 days of training data.
With minimal data, less than 30 days, the~\ac{bma} is among the best for the wind datasets.
With more training data, the~\ac{csge} with~\ac{tcn} source models, where the final layer is replaced by~\ac{blr} is the best.
We can also observe for these datasets that additionally considering the~\ac{gbrt} as the source model improves the results.

This observation also shows the flexibility of the~\ac{csge}.
Due to the combination through the forecast error, we can combine arbitrary models.
This flexibility is not given by the proposed~\ac{bma} approach.
However, the~\ac{bma} has the advantage that probabilistic forecasts are provided, which are not the focus of this article, but their evaluation is given in the appendix.

Another important consideration is that in almost all cases, the ensemble techniques outperform the~\baselineNS, which is the best model from the previous experiment.
These results show that a single source model's selection and adaptation process is highly uncertain because the model may be too specific for the target.
Due to specific characteristics of a single model, for example, the weather at the location or technical characteristics, selecting and adapting a single source model for the target is challenging.
In contrast, the combination of several models probably has a balancing effect on individual properties and improves the error significantly.

Besides the previous statistical discussion through the mean performance ranking, we also need to include an analysis of the forecast error for real-world implications.
Therefore, the forecast error, by means of the mean nRMSE, is summarized for this experiment in~\TBLs{tbl_model_selection_rmse_pv_ensemble} and \ref{tbl_model_selection_rmse_wind_ensemble}.
The best model from the previous experiment is again the \baselineNS.

In these tables, the error of the models decreases with increasing training data amount for all six datasets.
For the PV datasets, the best model has the largest error for the \PR dataset.
The best forecast error for this dataset is only $9.8$ percent with seven days of training data.
For the \PO with seven days of training data, the error is with $7$ percent error rate lower than results from~\cite{Schreiber2021}.

Also, for the~\WO dataset, the best models have similar error rates, between $16.3$ for seven days and $13.8$ percent for $90$ days of training data, similar to the result in~\cite{Schreiber2021}
The \WS dataset has the largest errors, between $16.7$ and $14.6$ percent, for the wind datasets.

Based on the analysis of the nRMSE, we can observe that even with a small amount of training data, good up to excellent prediction quality can be achieved. Furthermore, the mean nRMSE with more than 30 days often corresponds to error rates with a whole year of training data~\cite{Schreiber2021, VogtSynData2022}.
\section{Conclusion and Future Work}
\label{sec_itl_selection_conclusion}
We successfully evaluated several combinations of models, model selection, adaptation strategies, and two combination strategies on six datasets.
Our study's exhaustive evaluation is the most extensive evaluation for transfer learning in renewable power forecasts on real-world datasets.
We found that fine-tuning the final layer of a neural network, a well-known strategy, does not lead to convincing results in this setting.
Instead, replacing the layer with a Bayesian linear regression model trained with features extracted from the source and limited power measurements from the target task yields one of the best results, especially for a temporal convolutional neural network.
This result is best explained as source models in computer vision tasks are typically trained on many variations, e.g., various classification tasks. 
In contrast, renewable energy models are often trained on a single forecasting task.
This approach seems to generalize insufficiently for fine-tuning.
We suggest utilizing the forecast error with less than 30 days of training data for source model selection; the evidence is recommended with additional data.
We also showed how combining models leads to further significant improvements compared to considering a single model. 
The proposed coopetitive soft-gating ensemble combines source models solely based on the error of the target.
Our suggestion to utilize the Bayesian model averaging as an ensemble strategy is beneficial for minimal historical data.
To overcome the shortcomings of fine-tuning caused by limited data, we aim to augment the target data with synthetic data in the future. 
Likewise, we will expand our analysis for multi-task problems.

~\\
~\\
\noindent
{
    \footnotesize
    \textbf{Acknowledgments}
    This work results from the project TRANSFER (01IS20020B) funded by BMBF (German Federal Ministry of Education and Research).
    We thank Enercast GmBH for providing the PVREAL and WINDREAL datasets.
    We also thank Marek Herde, Mohammad Wazed Ali, and David Meier for their valuable input.
}
\bibliographystyle{elsarticle-num}
\bibliography{references}

\begin{thebibliography}{10}
\expandafter\ifx\csname url\endcsname\relax
  \def\url#1{\texttt{#1}}\fi
\expandafter\ifx\csname urlprefix\endcsname\relax\def\urlprefix{URL }\fi
\expandafter\ifx\csname href\endcsname\relax
  \def\href#1#2{#2} \def\path#1{#1}\fi

\bibitem{Schreiber2019a}
J.~Schreiber, {Transfer Learning in the Field of Renewable Energies - A
  Transfer Learning Framework Providing Power Forecasts Throughout the
  Lifecycle of Wind Farms After Initial Connection to the Electrical Grid}, in:
  Organic Computing - Doctoral Dissertation Colloquium, kassel university press
  GmbH, 2019, pp. 75--87.

\bibitem{Schwartz2019}
R.~Schwartz, J.~Dodge, N.~A. Smith, et~al., {Green AI}, CoRR (2019) 1--12ArXiv:
  1907.10597.

\bibitem{Schreiber2021}
J.~Schreiber, S.~Vogt, B.~Sick, {Task Embedding Temporal Convolution Networks
  for Transfer Learning Problems in Renewable Power Time-Series Forecast}, in:
  ECML, 2021, pp. 1--16.
\newblock \href {https://doi.org/10.1007/978-3-030-86514-6_8}
  {\path{doi:10.1007/978-3-030-86514-6_8}}.

\bibitem{You2021LogMe}
K.~You, Y.~Liu, J.~Wang, et~al., {LogME: Practical Assessment of Pre-trained
  Models for Transfer Learning}, in: ICML, 2021, pp. 12133--12143.
\newblock \href {http://arxiv.org/abs/2102.11005} {\path{arXiv:2102.11005}}.

\bibitem{You2021Hub}
K.~You, Y.~Liu, J.~Wang, et~al., {Ranking and Tuning Pre-trained Models: A New
  Paradigm of Exploiting Model Hubs}, CoRR arXiv:2110.10545 (2021) 1--45.
\newblock \href {http://arxiv.org/abs/2110.10545} {\path{arXiv:2110.10545}},
  \href {https://doi.org/10.48550/arXiv.2110.10545}
  {\path{doi:10.48550/arXiv.2110.10545}}.

\bibitem{Alkhayat2021}
G.~Alkhayat, R.~Mehmood, {A review and taxonomy of wind and solar energy
  forecasting methods based on deep learning}, Energy and AI 4 (2021) 100060.
\newblock \href {https://doi.org/10.1016/j.egyai.2021.100060}
  {\path{doi:10.1016/j.egyai.2021.100060}}.

\bibitem{VogtSynData2022}
S.~Vogt, J.~Schreiber, {Synthetic Photovoltaic and Wind Power Forecasting
  Data}, CoRR arXiv:2204.00411 (2022).

\bibitem{Li2020}
H.~Li, P.~Chaudhari, H.~Yang, et~al.,
  \href{http://arxiv.org/abs/2002.11770}{{Rethinking the Hyperparameters for
  Fine-tuning}}, in: ICLR, 2020, pp. 1--20.
\newblock \href {http://arxiv.org/abs/2002.11770} {\path{arXiv:2002.11770}}.
\newline\urlprefix\url{http://arxiv.org/abs/2002.11770}

\bibitem{Li2018}
X.~Li, Y.~Grandvalet, F.~Davoine, {Explicit inductive bias for transfer
  learning with convolutional networks}, in: ICML, Vol.~6, 2018, pp.
  4408--4419.
\newblock \href {http://arxiv.org/abs/1802.01483} {\path{arXiv:1802.01483}}.

\bibitem{Qureshi2019}
A.~S. Qureshi, A.~Khan, {Adaptive transfer learning in deep neural networks:
  Wind power prediction using knowledge transfer from region to region and
  between different task domains}, Computational Intelligence 35~(4) (2019)
  1088--1112.
\newblock \href {https://doi.org/10.1111/coin.12236}
  {\path{doi:10.1111/coin.12236}}.

\bibitem{Liu2021}
X.~Liu, Z.~Cao, Z.~Zhang, {Short-term predictions of multiple wind turbine
  power outputs based on deep neural networks with transfer learning}, Energy
  217 (2021) 119356.
\newblock \href {https://doi.org/10.1016/j.energy.2020.119356}
  {\path{doi:10.1016/j.energy.2020.119356}}.

\bibitem{Ju2020}
Y.~Ju, J.~Li, G.~Sun, {Ultra-Short-Term Photovoltaic Power Prediction Based on
  Self-Attention Mechanism and Multi-Task Learning}, IEEE Access 8 (2020)
  44821--44829.
\newblock \href {https://doi.org/10.1109/access.2020.2978635}
  {\path{doi:10.1109/access.2020.2978635}}.

\bibitem{Henze2020}
J.~Henze, J.~Schreiber, B.~Sick, {Representation Learning in Power Time Series
  Forecasting}, in: Deep Learning: Algorithms and Applications, Springer, Cham,
  2020, pp. 67--101.
\newblock \href {https://doi.org/10.1007/978-3-030-31760-7_3}
  {\path{doi:10.1007/978-3-030-31760-7_3}}.

\bibitem{Cao2018}
L.~Cao, L.~Wang, C.~Huang, X.~Luo, J.-H. Wang, {A Transfer Learning Strategy
  for Short-term Wind Power Forecasting}, in: Chinese Automation Congress,
  2018, pp. 3070--3075.
\newblock \href {https://doi.org/10.1016/j.renene.2015.06.034}
  {\path{doi:10.1016/j.renene.2015.06.034}}.

\bibitem{Cai2019}
L.~Cai, J.~Gu, J.~Ma, et~al., {Probabilistic wind power forecasting approach
  via instance-based transfer learning embedded gradient boosting decision
  trees}, Energies 12~(1) (2019) 159.
\newblock \href {https://doi.org/10.3390/en12010159}
  {\path{doi:10.3390/en12010159}}.

\bibitem{Liu2022}
Y.~Liu, J.~Wang, {Transfer learning based multi-layer extreme learning machine
  for probabilistic wind power forecasting}, Applied Energy 312 (2022) 118729.
\newblock \href {https://doi.org/10.1016/J.APENERGY.2022.118729}
  {\path{doi:10.1016/J.APENERGY.2022.118729}}.

\bibitem{Chen2020}
J.~Chen, Q.~Zhu, H.~Li, L.~Zhu, D.~Shi, Y.~Li, X.~Duan, Y.~Liu, {Learning
  Heterogeneous Features Jointly: A Deep End-to-End Framework for Multi-Step
  Short-Term Wind Power Prediction}, IEEE Transactions on Sustainable Energy
  11~(3) (2020) 1761--1772.
\newblock \href {https://doi.org/10.1109/TSTE.2019.2940590}
  {\path{doi:10.1109/TSTE.2019.2940590}}.

\bibitem{Sheng2022}
H.~Sheng, B.~Ray, J.~Shao, D.~Lasantha, N.~Das, {Generalization of solar power
  yield modelling using knowledge transfer}, Expert Systems with Applications
  (2022) 116992\href {https://doi.org/10.1016/J.ESWA.2022.116992}
  {\path{doi:10.1016/J.ESWA.2022.116992}}.

\bibitem{Yin2021}
H.~Yin, Z.~Ou, J.~Fu, Y.~Cai, S.~Chen, A.~Meng, {A novel transfer learning
  approach for wind power prediction based on a serio-parallel deep learning
  architecture}, Energy 234 (2021) 121271.
\newblock \href {https://doi.org/10.1016/J.ENERGY.2021.121271}
  {\path{doi:10.1016/J.ENERGY.2021.121271}}.

\bibitem{Khan2022}
M.~Khan, M.~R. Naeem, E.~A. Al-Ammar, W.~Ko, H.~Vettikalladi, I.~Ahmad, {Power
  Forecasting of Regional Wind Farms via Variational Auto-Encoder and Deep
  Hybrid Transfer Learning}, Electronics 11~(2) (2022) 206.
\newblock \href {https://doi.org/10.3390/ELECTRONICS11020206}
  {\path{doi:10.3390/ELECTRONICS11020206}}.

\bibitem{Almonacid2022}
G.~Almonacid-Olleros, G.~Almonacid, D.~Gil, J.~Medina-Quero, {Evaluation of
  Transfer Learning and Fine-Tuning to Nowcast Energy Generation of
  Photovoltaic Systems in Different Climates}, Sustainability 14~(5) (2022)
  3092.
\newblock \href {https://doi.org/10.3390/SU14053092}
  {\path{doi:10.3390/SU14053092}}.

\bibitem{Yan2019}
C.~Yan, Y.~Pan, C.~L. Archer, {A general method to estimate wind farm power
  using artificial neural networks}, Wind Energy 22~(11) (2019) 1421--1432.
\newblock \href {https://doi.org/10.1002/WE.2379} {\path{doi:10.1002/WE.2379}}.

\bibitem{Jens2019}
J.~Schreiber, A.~Buschin, B.~Sick, {Influences in Forecast Errors for Wind and
  Photovoltaic Power: A Study on Machine Learning Models}, in: INFORMATIK 2019,
  GI e.V., 2019, pp. 585--598.
\newblock \href {https://doi.org/10.18420/inf2019\_74}
  {\path{doi:10.18420/inf2019\_74}}.

\bibitem{Zhou2020}
S.~Zhou, L.~Zhou, M.~Mao, et~al., {Transfer learning for photovoltaic power
  forecasting with long short-term memory neural network}, in: International
  Conference on Big Data and Smart Computing (BigComp), 2020, pp. 125--132.
\newblock \href {https://doi.org/10.1109/BigComp48618.2020.00-87}
  {\path{doi:10.1109/BigComp48618.2020.00-87}}.

\bibitem{Ceci2017}
M.~Ceci, R.~Corizzo, F.~Fumarola, et~al., {Predictive Modeling of PV Energy
  Production: How to Set Up the Learning Task for a Better Prediction?}, IEEE
  TIL 13~(3) (2017) 956--966.
\newblock \href {https://doi.org/10.1109/TII.2016.2604758}
  {\path{doi:10.1109/TII.2016.2604758}}.

\bibitem{Shireen2018}
T.~Shireen, C.~Shao, H.~Wang, et~al., {Iterative multi-task learning for
  time-series modeling of solar panel PV outputs}, Applied Energy 212 (2018)
  654--662.
\newblock \href {https://doi.org/10.1016/j.apenergy.2017.12.058}
  {\path{doi:10.1016/j.apenergy.2017.12.058}}.

\bibitem{Tasnim2018}
S.~Tasnim, A.~Rahman, A.~Oo, et~al., {Wind power prediction in new stations
  based on knowledge of existing Stations: A cluster based multi source domain
  adaptation approach}, Knowledge-Based Systems 145 (2018) 15--24.
\newblock \href {https://doi.org/10.1016/j.knosys.2017.12.036}
  {\path{doi:10.1016/j.knosys.2017.12.036}}.

\bibitem{Vogt2019}
S.~Vogt, A.~Braun, J.~Dobschinski, et~al., {Wind Power Forecasting Based on
  Deep Neural Networks and Transfer Learning}, in: Wind Integration Workshop,
  Vol.~18, 2019, p.~8.

\bibitem{Schreiber2020}
J.~Schreiber, B.~Sick, {Emerging Relation Network and Task Embedding for
  Multi-Task Regression Problems}, in: ICPR, 2020, pp. 2663--2670.
\newblock \href {https://doi.org/10.1109/ICPR48806.2021.9412476}
  {\path{doi:10.1109/ICPR48806.2021.9412476}}.

\bibitem{Bishop2006}
C.~M. Bishop, {Pattern Recognition and Machine Learning}, Springer, 2006.

\bibitem{deisenroth2020mathematics}
M.~Borthwick, {Math for Machine Learning}, Cambridge University, 2019.
\newblock \href {https://doi.org/10.1515/9781400843909-001}
  {\path{doi:10.1515/9781400843909-001}}.

\bibitem{Hoeting1999}
J.~A. Hoeting, D.~Madigan, A.~E. Raftery, et~al., Bayesian model averaging: a
  tutorial, MathSciNet 14~(4) (1999) 382--417.
\newblock \href {https://doi.org/10.1214/SS/1009212519}
  {\path{doi:10.1214/SS/1009212519}}.

\bibitem{GS18}
A.~Gensler, B.~Sick, A multi-scheme ensemble using coopetitive soft gating with
  application to power forecasting for renewable energy generation, CoRR
  arXiv:1803.06344 (2018).

\bibitem{Gensler2018}
A.~Gensler, {Wind power ensemble forecasting}, Ph.D. thesis, University of
  Kassel (2018).

\bibitem{DBS+18}
S.~Deist, M.~Bieshaar, J.~Schreiber, A.~Gensler, B.~Sick, Coopetitive soft
  gating ensemble, in: Workshop on Self-Improving System Integration (SISSY),
  Trento, Italy, 2018, pp. 190--197.

\bibitem{maartenPHD}
M.~Bieshaar, Cooperative intention detection using machine learning. Advanced
  cyclist protection in the context of automated driving, kassel university
  press, 2021.

\bibitem{GS16}
A.~Gensler, B.~Sick, Forecasting wind power - an ensemble technique with
  gradual coopetitive weighting based on weather situation, in: IJCNN,
  Vancouver, BC, 2016, pp. 4976--4984.

\bibitem{Sain2006}
S.~R. Sain, V.~N. Vapnik, {The Nature of Statistical Learning Theory},
  Springer-Verlag New York, 2006.
\newblock \href {https://doi.org/10.1007/978-1-4757-3264-1}
  {\path{doi:10.1007/978-1-4757-3264-1}}.

\bibitem{Mclean2008}
F.~Van~Hulle, J.~O. Tande, K.~Uhlen, et~al., {Further Developing Europe's Power
  Market for Large Scale Integration of Wind Power - Equivalent Wind Power
  Curves}, European Wind Energy Association (EWEA) (2008).

\bibitem{Schreiber2020Prophesy}
J.~Schreiber, M.~Siefert, K.~Winter, A.~Wessel, R.~Fritz, G.~Good, A.~Schella,
  J.~Muraschko, S.~Staedler, {Abschlussbericht Projekt Prophesy :
  Prognoseunsicherheiten von Windenergie und Photovoltaik in zuk{\"{u}}nftigen
  Stromversorgungssysteme}, German National Library of Science and Technology
  (2020) 168\href {https://doi.org/10.2314/KXP:176026797X}
  {\path{doi:10.2314/KXP:176026797X}}.

\bibitem{GS17}
A.~Gensler, B.~Sick, Probabilistic wind power forecasting: A multi-scheme
  ensemble technique with gradual coopetitive soft gating, in: SSCI, Honolulu,
  HI, 2017, pp. 1--10.

\end{thebibliography}
\onecolumn
\clearpage
\appendix
\setcounter{footnote}{0}
\pagenumbering{roman} 
\setcounter{page}{1}

\glsresetall
\numberwithin{equation}{section}
\numberwithin{figure}{section}
\numberwithin{table}{section}


\section{Data}\label{sec_appendix_model_selection_data}
%

\begin{figure}[htb]
         \centering
          \includegraphics[width=0.98\textwidth]{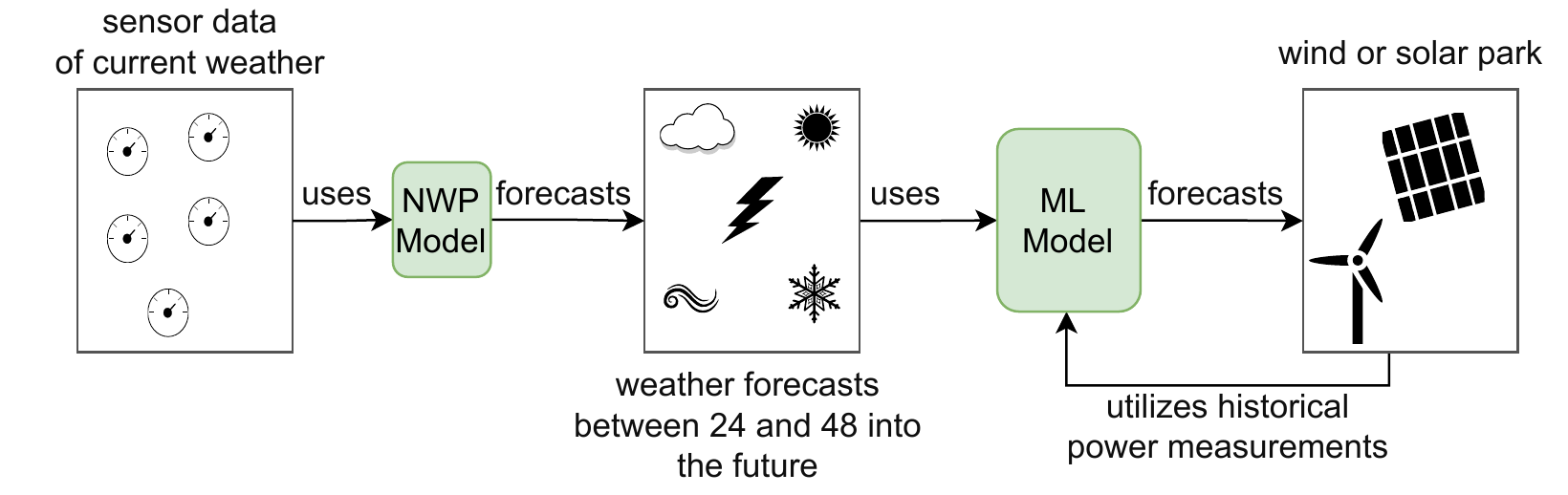}
         \caption{Overview of renewable power forecast process.}
        \label{fig_nwp_ml_overview}
\end{figure}

The following section details challenges with the different datasets.
We also include a description on the overall process for generating power forecasts for a wind or~\ac{pv} park.
\FIG{fig_nwp_ml_overview} summarizes this process.
Due to the weather dependency of renewable power plants, we require weather predictions from so called~\ac{nwp} models.
The~\ac{nwp} model has input from sensors that approximate the current weather situation.
Based on the last sensory data, a so-called model run is calculated.
This model run is, e.g., 0.00 am.
Due to complex and manifold stochastic differential equations involved in predicting the weather, such a model run typically requires about six hours until it is finished.
Afterward, the~\ac{nwp} provides forecasts, e.g., up to $72$ hours into the future.
In our case, we are interested in so-called day-ahead power forecasts based on weather forecasts between $24$ and $48$ hours into the future.
Based on these weather forecasts and historical power measurements, we can train a~\ac{ml} model to predict the expected power generation for day-ahead forecasting problems.

\begin{figure}
     \centering
     \subfloat[PVSYN datastet.]{
         \centering
          \includegraphics[width=0.47\textwidth]{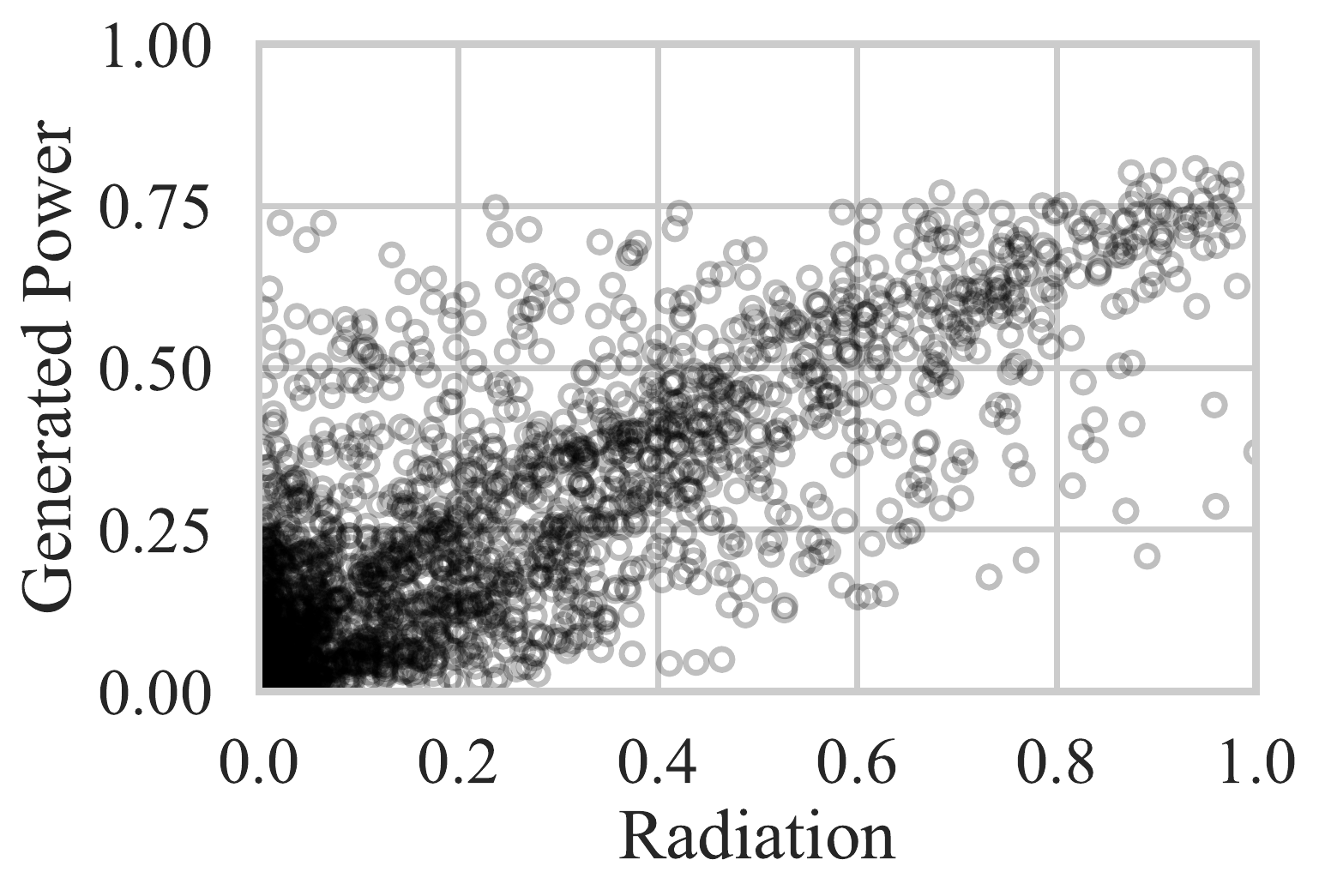}
    }
    \subfloat[WINDSYN datastet.]{
         \centering
          \includegraphics[width=0.47\textwidth]{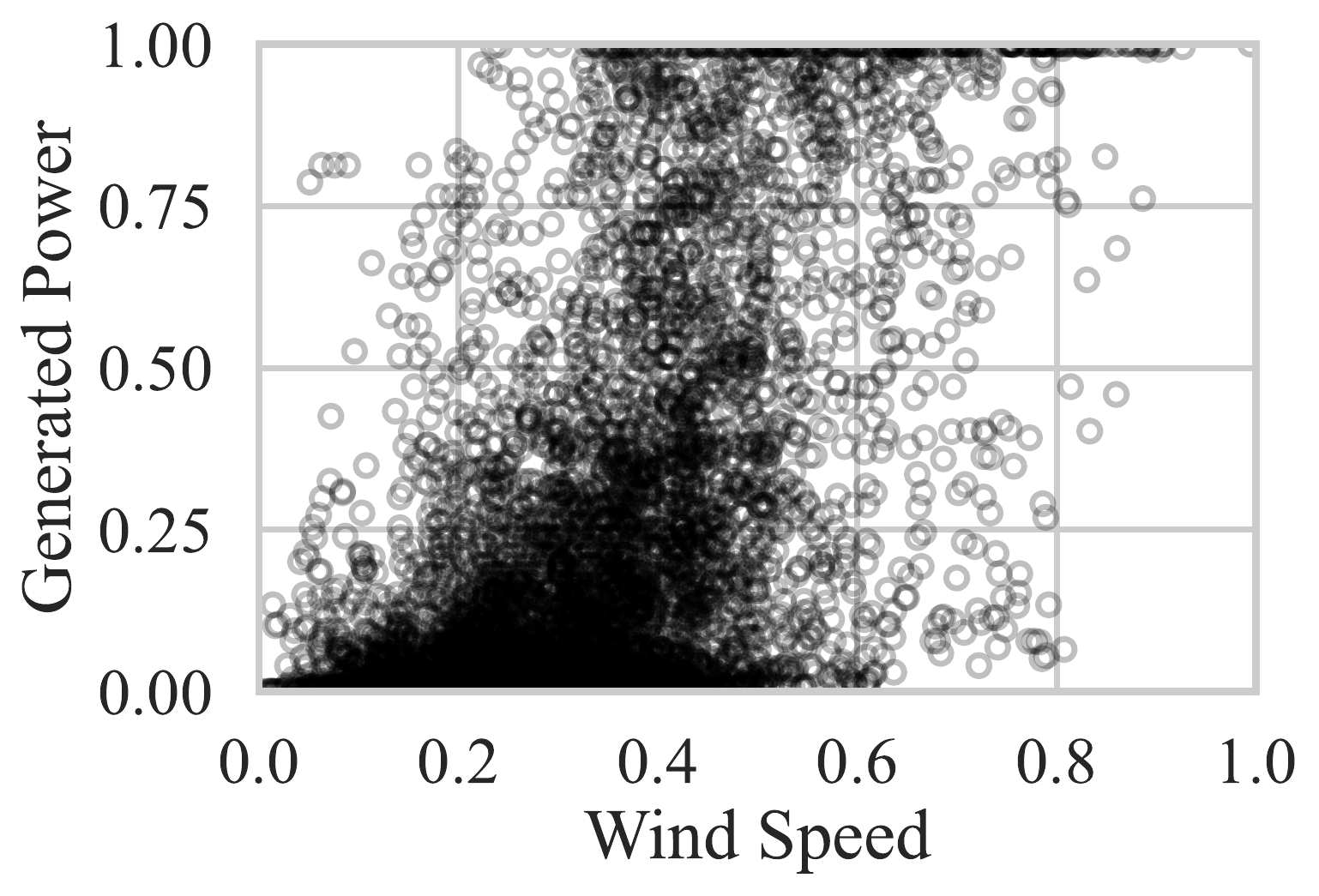}
    }
    \hfill
    \centering
    \subfloat[PVOPEN datastet.]{
         \centering
          \includegraphics[width=0.48\textwidth]{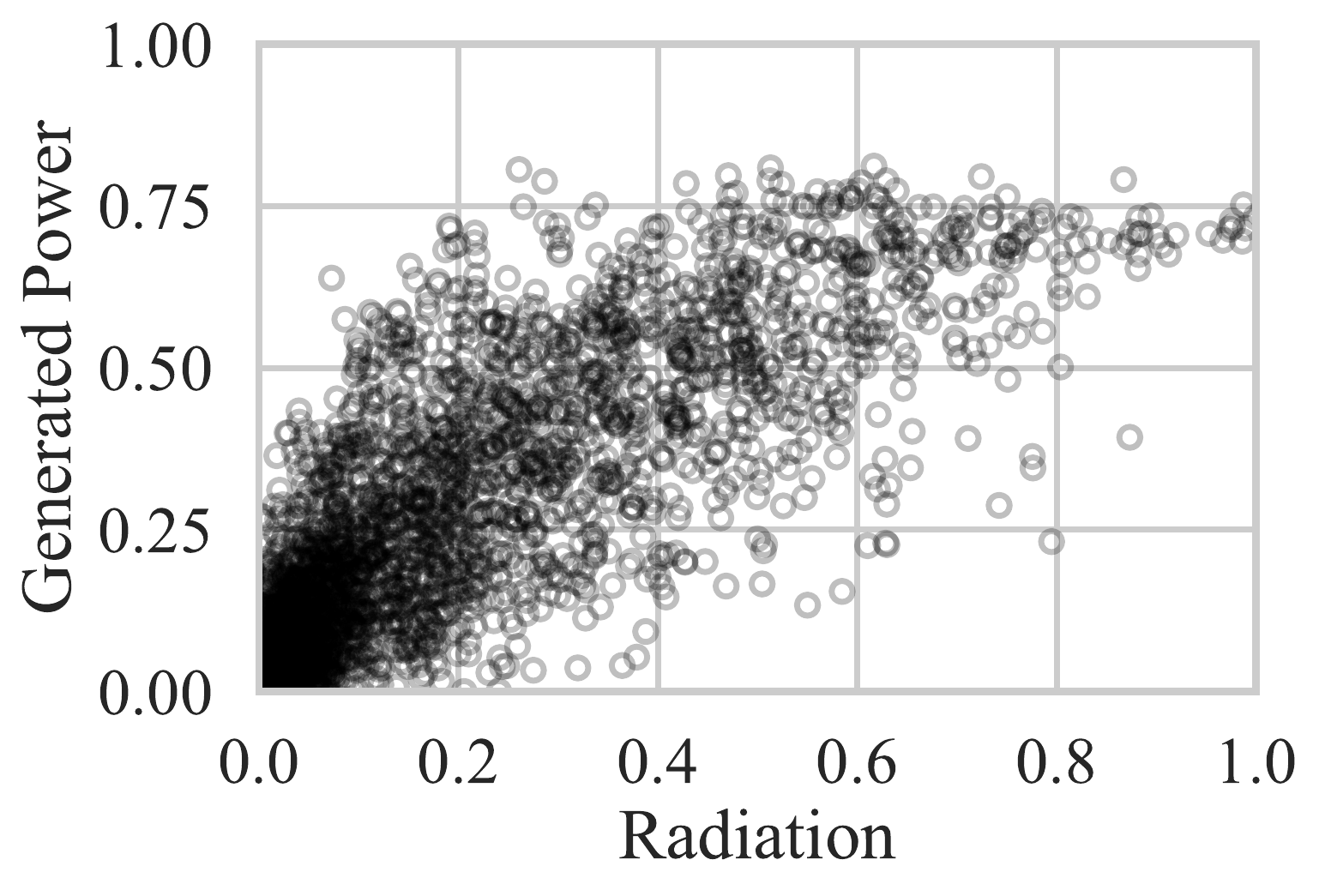}
    }
    \subfloat[WINDOPEN datastet.]{
         \centering
          \includegraphics[width=0.48\textwidth]{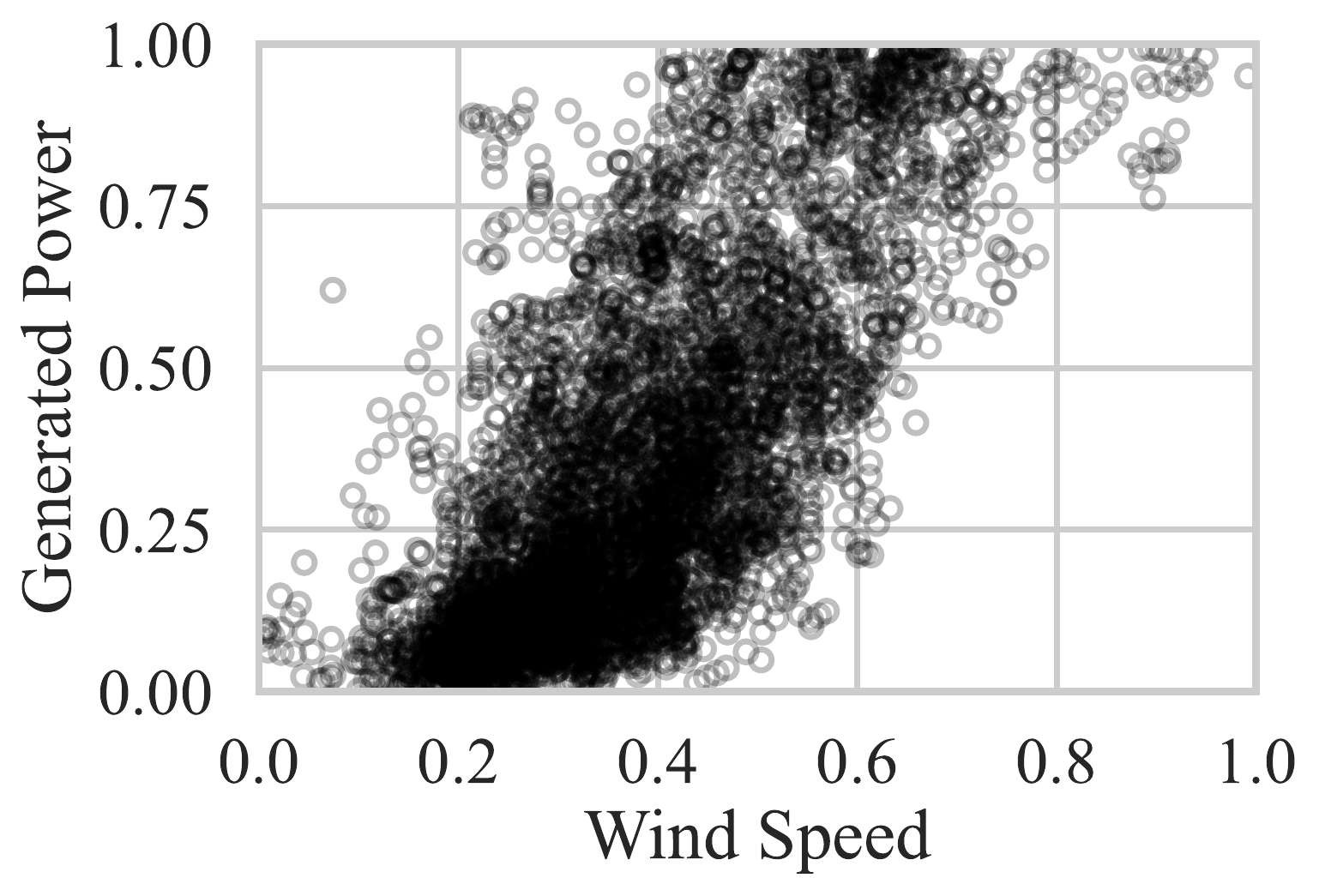}
     }
    \hfill
    \subfloat[PVREAL datastet.]{
         \centering
          \includegraphics[width=0.48\textwidth]{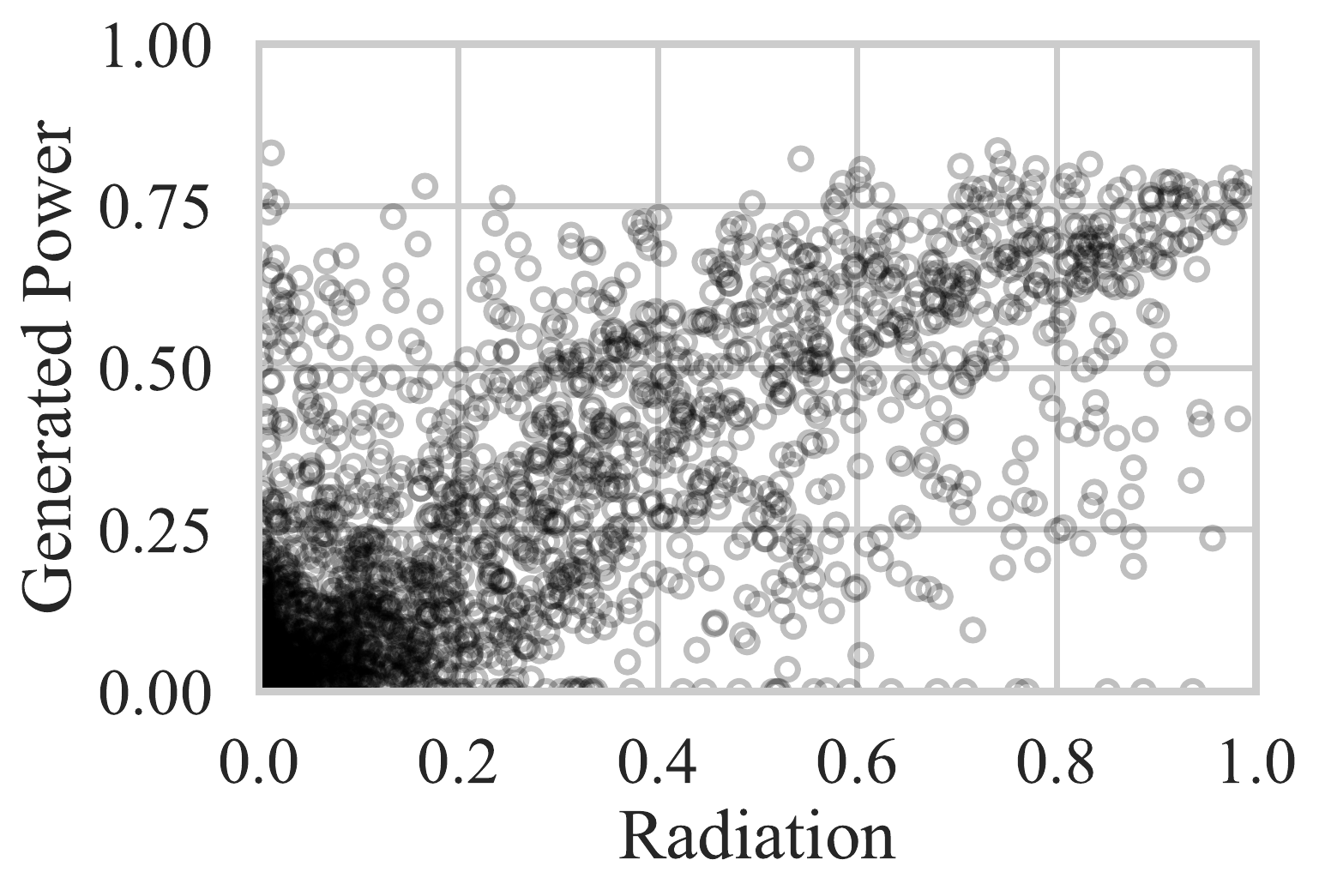}
    }
    \subfloat[WINDREAL datastet.]{
         \centering
          \includegraphics[width=0.48\textwidth]{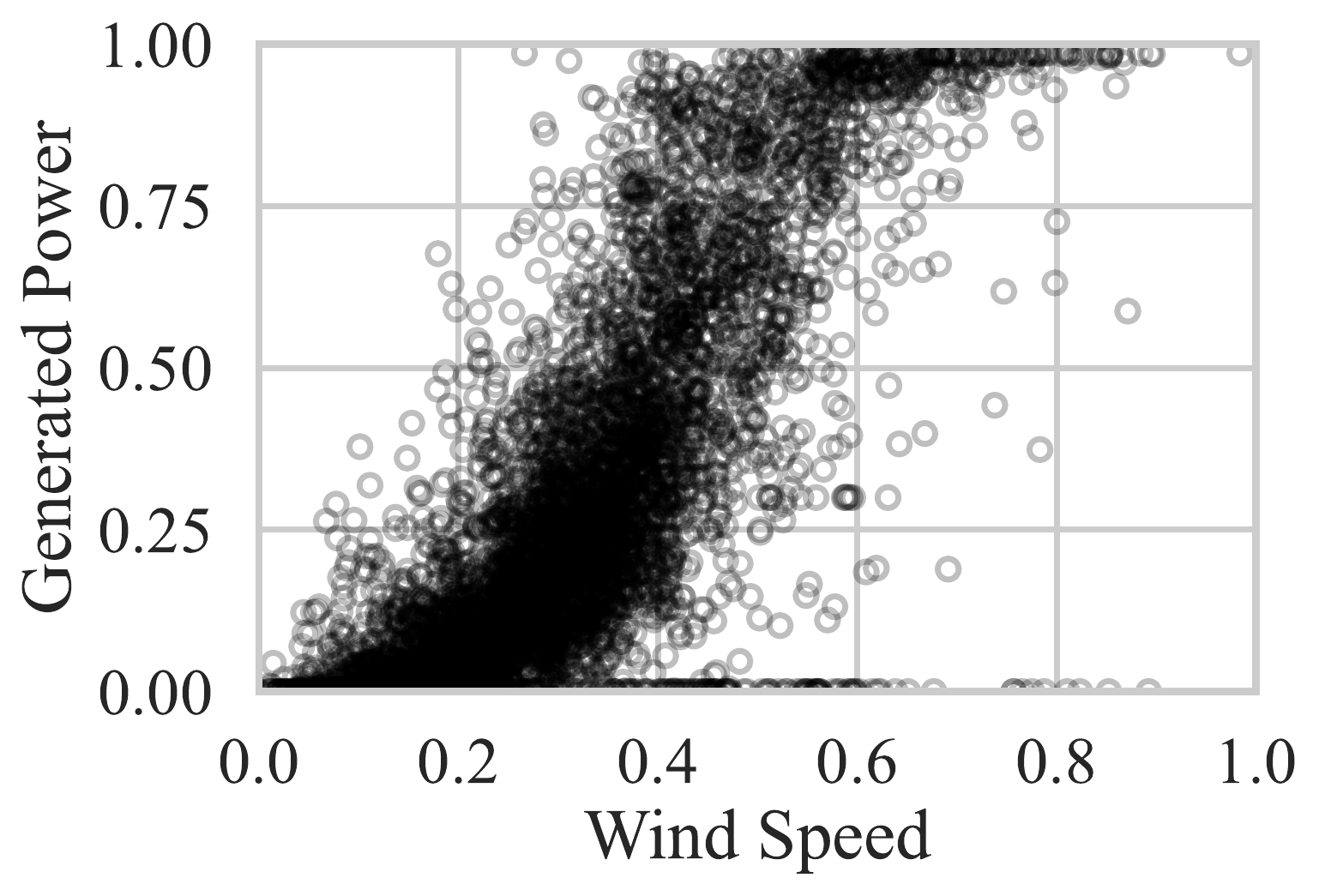}
    }
    \caption{Scatter plots of most relevant features for power forecasts from day-ahead weather forecasts and the historical power measurements for an exemplary park from all datasets.}
    \label{fig_scatter_all_datasets}
\end{figure}

\begin{figure}
     \centering
    \subfloat[PVSYN dataset.]{
         \centering
          \includegraphics[width=0.98\textwidth]{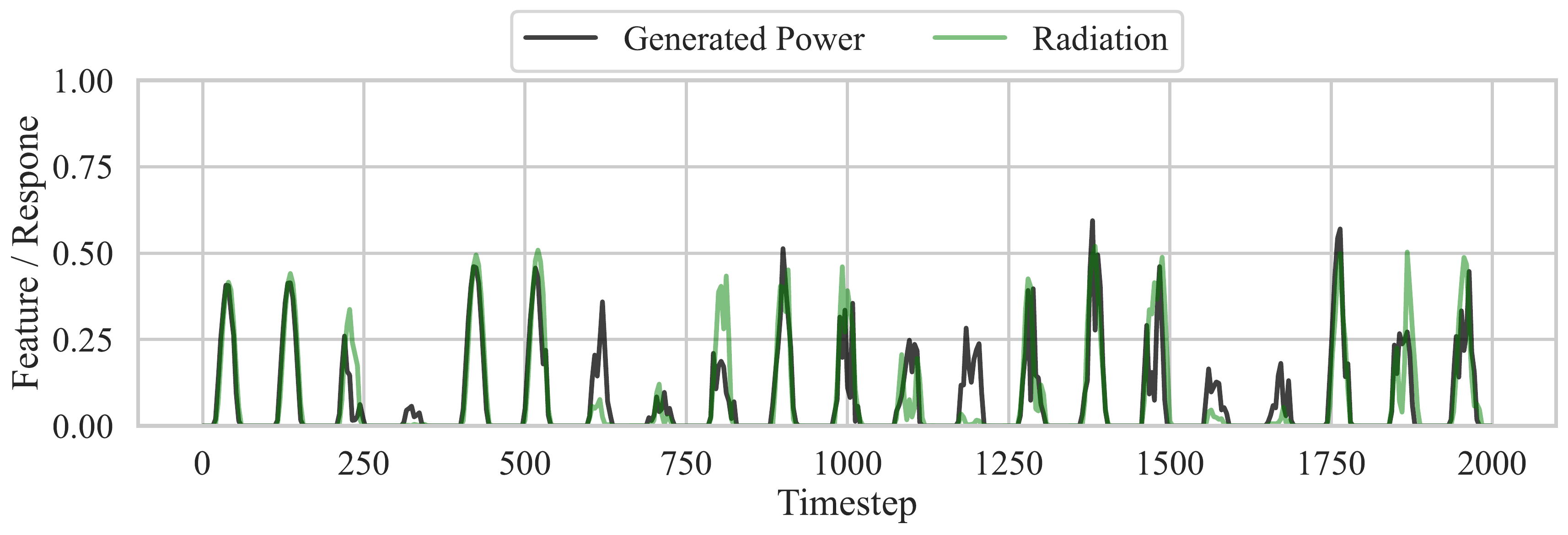}
    }
     
     \subfloat[PVOPEN dataset.]{
         \centering
          \includegraphics[width=0.98\textwidth]{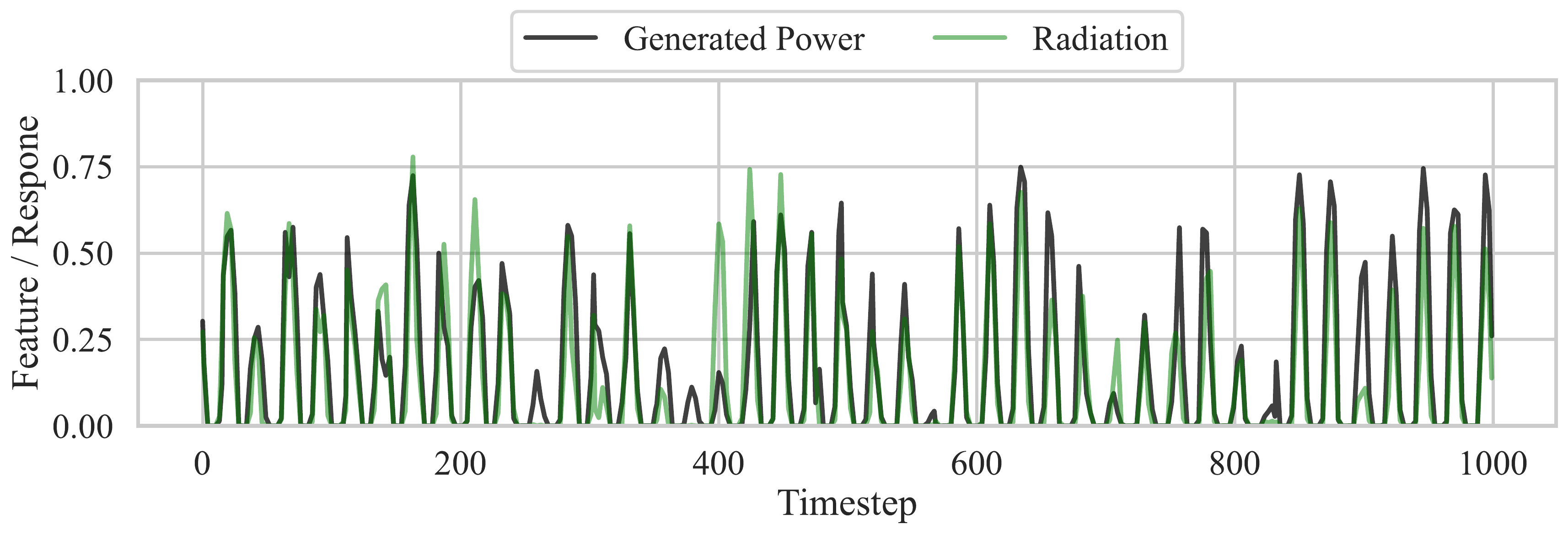}
    }
    \hfill
     \centering
    \subfloat[PVREAL dataset.]{
         \centering
          \includegraphics[width=0.98\textwidth]{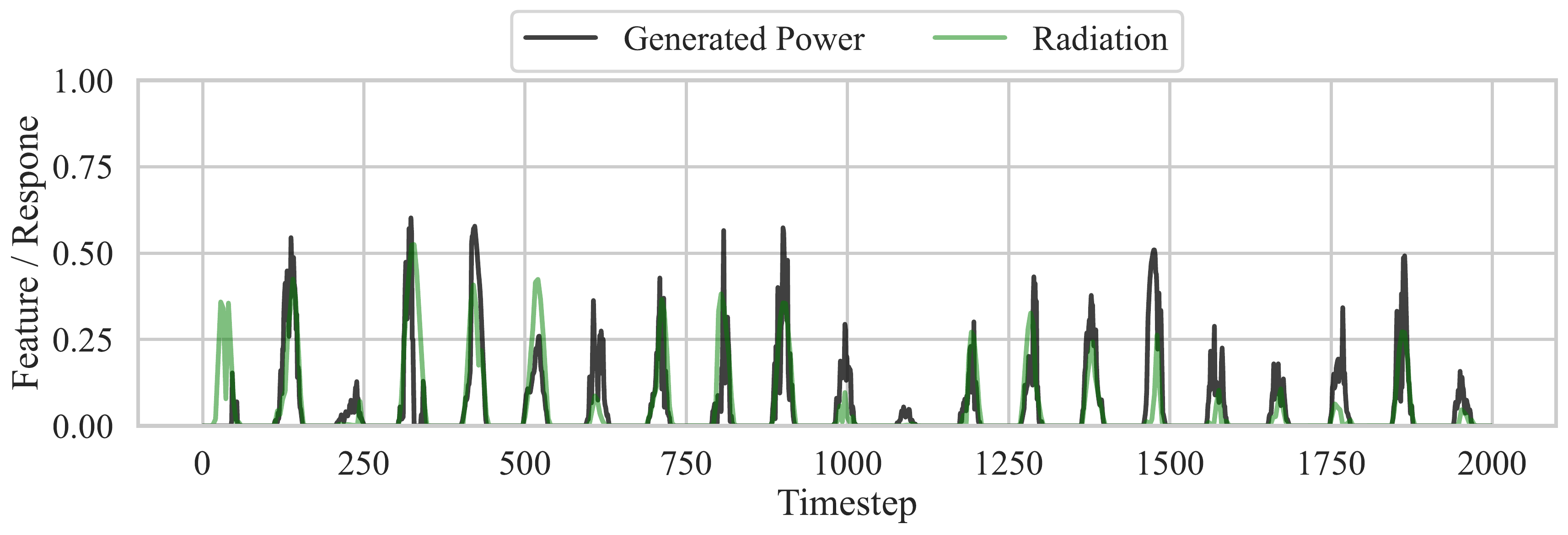}
    }
     \hfill
    \caption{Timeseries plots of~\ac{pv} datasets.}
    \label{fig_timeseries_all_pv_datasets}
\end{figure}

\begin{figure}
     \centering
    \subfloat[WINDSYN dataset.]{
         \centering
          \includegraphics[width=0.98\textwidth]{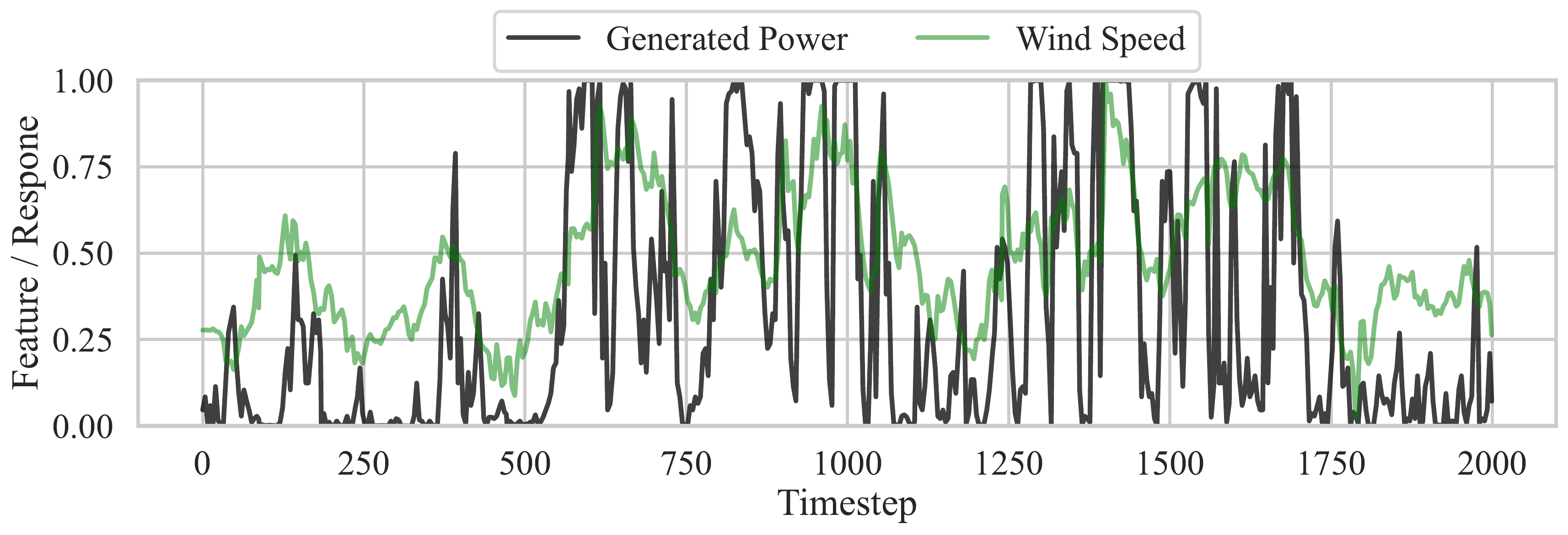}
    }
     \hfill
    \subfloat[WINDOPEN dataset.]{
         \centering
          \includegraphics[width=0.98\textwidth]{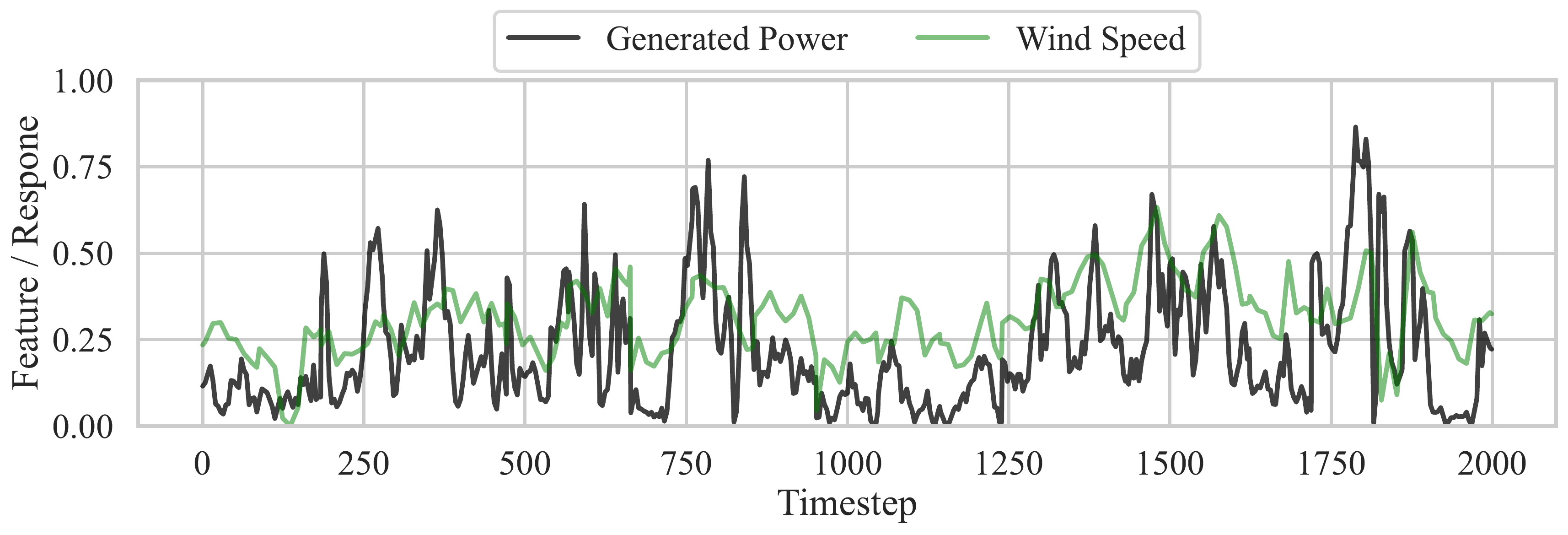}
    }
    \hfill
     \centering
    \subfloat[WINDREAL dataset.\label{fig_windreal_ts_sample}]{
         \centering
          \includegraphics[width=0.98\textwidth]{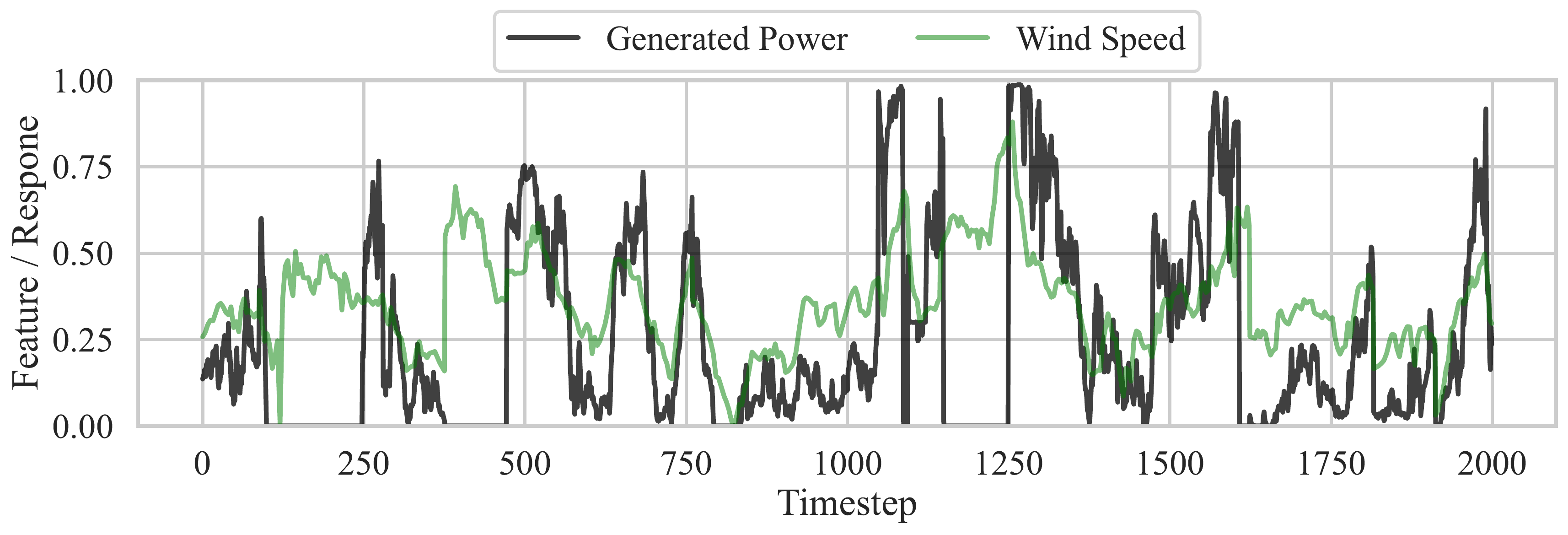}
    }
     \hfill
    \caption{Timeseries plots of wind datasets.}
    \label{fig_timeseries_all_wind_datasets}
\end{figure}

However, due to the dependency of renewable power forecasts on weather forecasts as input features, substantial uncertainty is associated with these forecasts making it a challenging problem.
At the same time, weather forecasts are valid for larger grid sizes, e.g., three kilometers and a mismatch between these grids and the location of a wind or~\ac{pv} causes additional uncertainty in the power forecasts.
These mismatches and the non-linearity of the forecasting problem are visually accessible in \FIG{fig_scatter_all_datasets}.
We can observe various mismatches between the predicted wind speed (or radiation) with historical power measurements.
For instance, we can observe (outliers) where a large amount of power is generated for low values of those features.
Those mismatches are also visually accessible in the time-series plots in \FIG{fig_timeseries_all_pv_datasets} and~\ref{fig_timeseries_all_wind_datasets}.
This observation indicates that the weather forecasts were wrong.
Various examples are also present where we observe a large value of wind speed or radiation, but no or little power is generated.
A wrong weather forecast can cause this problem. However, often, it is associated with regular interventions.
For instance, in some regions in Germany, wind turbines need to limit the rotation speed during the night.
Also, there is a large portion of feed-in management interventions in Germany.
These interventions are used to stabilize the electrical grid.
A typical pattern for those interventions is given in~\FIG{fig_windreal_ts_sample}. 
We can observe an initial large power production associated with large wind speed values.
At this point, the power generation drops to zero, while the wind speed remains high.
Such a sudden drop is typically associated with feed-in management interventions.
As those interventions depend on the power grid's state, we typically have no information about such drops, making the forecasting problem even more challenging.

\clearpage

\section{Method}\label{sec_appendix_model_selection_method}
The following sections detail proposed methods and utilized approaches.

\subsection{Coopetitive Soft Gating Ensemble for Model Combination}

To adapt source models from a model hub for a target, it is essential that a learning strategy only adapts relevant knowledge for the target.
For instance, a learning strategy should assure that a model is not adapted to seasonal specific behavior when only limited data of a specific season is available.
One way to achieve this is through ensembles.
The~\ac{bma} ensemble, for instance, allows combining source models by their marginal likelihood on the target.
However, in practice, Bayesian models are often not available.
While the~\ac{bma} combines models through the posterior, we propose to utilize the~\ac{csge}, which combines ensemble members based on the error scores.
Recently, the~\ac{csge} has proven successful in renewable power~\cite{DBS+18,Gensler2018,GS18} and trajectory forecasts~\cite{maartenPHD}.
The following section details this method in the context of~\ac{itl}.

Ensemble approaches are categorized as either weighting or gating strategies to combine ensemble members.
In the case of a gating strategy, a single ensemble member provides a prediction and other source models are neglected.
In the case of a weighting strategy, several ensemble members are combined.
The~\ac{csge} incorporates both principles through the soft gating principle~\cite{GS16}.
The idea of the CSGE is to link the weights to the ensemble members' performance, i.e., good source models are weighted stronger than weaker ones.
The overall weight of a source model for the target task is characterized using three aspects:

\begin{itemize}
    \item The \textit{global weight} is defined by how well a source model performs with the available training data on the target task.
    \item The \textit{local weight} is defined by how well a source model performs on the target tasks for different areas in the feature space. For example, in the case of wind, one model might performs well for low wind speeds, while another source model might performs well for larger wind speeds on the target.
    \item The \textit{forecast horizon-dependent weight} is defined by how well a source model performs for different lead times on the target task.
\end{itemize}

\begin{figure}[htb]
    \centering
    {
    \large
    \begin{center} 
    \resizebox{0.75\textwidth}{!}{%
        \tikzstyle{block} = [rectangle, rounded corners, text centered, draw=black]
        \tikzstyle{clear} = [rectangle, rounded corners, text centered]
        \tikzstyle{arrow} = [thick,->,>=stealth]
        \tikzstyle{connect} = [thick,-,>=stealth]
            \begin{tikzpicture}[node distance=1cm]
            
            \node [block, text width=2cm, text height=6cm,] (cw) {};
            \node [block, text width=2cm, text height=6cm, right of=cw, xshift=2.5cm] (nw) {};
            \node [block, text width=2cm, text height=6cm, right of=nw, xshift=4cm] (ap) {};
            \node[text width=2.6cm] at (0.3,0.0) {\textbf{Compute Weights}};
            \node[text width=2.6cm] at (3.75,0.0) {\textbf{Normalize Weights}};
            \node[text width=2.3cm] at (8.6,0.0) {\textbf{Aggregate Forecasts}};

            \node [block, text width=1.5cm, text height=0.25cm, above of=cw, opacity=0.75, yshift=1.75cm, dashed] (global) {Global};
            \node [block, text width=1.5cm, text height=0.25cm, below of=global, opacity=0.75, yshift=-0.5cm, dashed] (local) {Local};
            \node [block, text width=1.5cm, text height=0.25cm, below of=local, opacity=0.75, yshift=-1.8cm, dashed] (fh) {Forecast Horizon};
            
            \node [block, text width=3cm, text height=0.5cm, left of=cw, xshift=-3cm, yshift=2.7cm] (sm1) {Source Model (1)};
            \node [block, text width=3cm, text height=0.5cm, below of=sm1, yshift=-0.5cm] (sm2) {Source Model (2)};
            \node [clear, text width=3cm, text height=0.5cm, below of=sm2, yshift=-0.5cm] (sdots) {...};
            \node [block, text width=3cm, text height=0.5cm, below of=sdots, yshift=-0.5cm] (smj) {Source Model (M)};
            
            \node [clear, text width=0.5cm, text height=0.5cm, left of=cw, xshift=-6.5cm] (x) {};
            \node [clear, text width=0.0cm, text height=0.0cm, right of=x] (xa) {};
            \draw [connect] (x.east) -- node[anchor=north, yshift=0.6cm, xshift=-0.25cm] {$\inputFeatureVector_{t + k | t}$} (xa.east);
            \draw [arrow] (xa.east) |- node[] {} (sm1);
            \draw [arrow] (xa.east) |- node[] {} (sm2);
            \draw [arrow] (xa.east) |- node[] {} (smj);
            
            \node [clear, text width=0.5cm, text height=0.5cm, right of=ap, xshift=2cm] (y) {};
            \draw [arrow] (ap.east) -- node[anchor=north, yshift=0.75cm] {$\bar{\outputPredictionVector}_{t+k|t}$} (y.west);
            
            \node [clear, right of=sm1, xshift=3cm, text width=2cm] (cw1) {};
            \node [clear, right of=sm2, xshift=3cm, text width=2cm] (cw2) {};
            \node [clear, right of=smj, xshift=3cm, text width=2cm] (cwj) {};
            \draw [arrow] (sm1.east) -- node[anchor=north, yshift=0.8cm] {$\outputPredictionVector^{(1)}_{t+k|t}$} (cw1.west);
            \draw [arrow] (sm2.east) -- node[anchor=north, yshift=0.8cm] {$\outputPredictionVector^{(2)}_{t+k|t}$} (cw2.west);
            \draw [arrow] (smj.east) -- node[anchor=north, yshift=0.8cm] {$\outputPredictionVector^{(\numberSourceModels)}_{t+k|t}$} (cwj.west);
            
            \node [clear, right of=cw1, xshift=2.5cm, text width=2cm] (nw1) {};
            \node [clear, right of=cw2, xshift=2.5cm, text width=2cm] (nw2) {};
            \node [clear, right of=cwj, xshift=2.5cm, text width=2cm] (nwj) {};
            \draw [arrow] (cw1.east) -- node[anchor=north, yshift=0.8cm] {$\csgeWeightUnnormalized^{(1)}_{t+k|t}$} (nw1.west);
            \draw [arrow] (cw2.east) -- node[anchor=north, yshift=0.8cm] {$\csgeWeightUnnormalized^{(2)}_{t+k|t}$} (nw2.west);
            \draw [arrow] (cwj.east) -- node[anchor=north, yshift=0.8cm] {$\csgeWeightUnnormalized^{(\numberSourceModels)}_{t+k|t}$} (nwj.west);
            
            \node [clear, right of=nw1, xshift=4cm, text width=2cm] (ap1) {};
            \node [clear, right of=nw2, xshift=4cm, text width=2cm] (ap2) {};
            \node [clear, right of=nwj, xshift=4cm, text width=2cm] (apj) {};
            \draw [arrow] (nw1.east) -- node[anchor=north, yshift=0.8cm] {$\csgeWeightNormalized^{(1)}_{t+k|t} \cdot \outputPredictionVector^{(1)}_{t+k|t}$} (ap1.west);
            \draw [arrow] (nw2.east) -- node[anchor=north, yshift=0.8cm] {$\csgeWeightNormalized^{(2)}_{t+k|t} \cdot \outputPredictionVector^{(2)}_{t+k|t}$} (ap2.west);
            \draw [arrow] (nwj.east) -- node[anchor=north, yshift=0.8cm] {$\csgeWeightNormalized^{(\numberSourceModels)}_{t+k|t} \cdot \outputPredictionVector^{(\numberSourceModels)}_{t+k|t}$} (apj.west);
            
            \node [clear, right of=sm1, xshift=1.25cm] (sc1) {};
            \node [clear, right of=cw1, xshift=2.5cm, yshift=-0.5cm, text width=2cm] (nw1h) {};
            \draw [arrow, opacity=0.5] (sc1.center) |- node[] {} (nw1h.west);
            \node [clear, right of=sm2, xshift=1.25cm] (sc2) {};
            \node [clear, right of=cw2, xshift=2.5cm, yshift=-0.5cm, text width=2cm] (nw2h) {};
            \draw [arrow, opacity=0.5] (sc2.center) |- node[] {} (nw2h.west);
            \node [clear, right of=smj, xshift=1.25cm] (scj) {};
            \node [clear, right of=cwj, xshift=2.5cm, yshift=-0.5cm, text width=2cm] (nwjh) {};
            \draw [arrow, opacity=0.5] (scj.center) |- node[] {} (nwjh.west);
            
            \node [clear, text width=3cm, text height=0.5cm, right of=sdots, xshift=1.25cm] (ydots) {...};
            \node [clear, text width=3cm, text height=0.5cm, right of=sdots, xshift=4.75cm] (wdots) {...};
            \node [clear, text width=3cm, text height=0.5cm, right of=sdots, xshift=9cm] (wydots) {...};

            \end{tikzpicture}
    }
    \end{center}
}
    \caption{The architecture of the CSGE. The source models' predictions $\outputPredictionVector^{(\indexSourceModel)}_{t + k | t}$ for the input $\inputFeatureVector$ are passed to the CSGE. The ensemble member's weights are given by the aggregation of the respective global-, local- and forecast horizon-dependent weights. The weights are normalized. In the final step, the source models' predictions are weighted and aggregated.}
    \label{fig:coop_int_fusion:csge_overview_graphic_appendix}
\end{figure}

Fig.~\ref{fig:coop_int_fusion:csge_overview_graphic} provides an overview of the~\ac{csge}.
In the following, we treat ensemble members equivalently to source models from a model hub.
The CSGE includes $\numberSourceModels$ ensemble members, with $\indexSourceModel \in \{1, ..., \numberSourceModels\}$.
Each of these ensemble members is a source model with a predictive function $f_m$. 
Each source model forecasts uni-variate estimates $\outputPredictionVector^{(\indexSourceModel)}_{t + k | t} \in \mathbb{R}$ for the input $\inputFeatureVector_{t + k | t} \in \mathbb{R}^{\dimFeature}$ of a target task $T$, we omit the subscript $T$ for reasons of clarity and comprehensibility.
Let $\dimFeature$ be the dimension of the input feature vector $\inputFeatureVector_{t + k | t}$.
Then, $k$ denotes the forecast horizon, denoted by the subscript, for the forecast origin $t$.
Furthermore, let $\outputPredictionVector^{(\indexSourceModel)}_{t + k | t}$ the $\indexSourceModel$-th source model's prediction.
For each prediction of each source model, we compute an aggregated weight $\csgeWeightNormalized^{(m)}_{t + k | t}$, based on the global, local, and forecast horizon-dependent weighting.
Finally, the weight and the prediction of each model for each forecast horizon are aggregated as a final prediction through

\begin{align}
    \label{eq:coop_int_fusion:csge:ensemble_pred}
    \bar{\outputPredictionVector}_{t + k | t} = \sum^{\numberSourceModels}_{\indexSourceModel=1} \csgeWeightNormalized^{(\indexSourceModel)}_{t + k | t} \cdot \outputPredictionVector^{(\indexSourceModel)}_{t + k | t},
\end{align}
with the following constraints
\begin{align}
    \label{eq:coop_int_fusion:csge:formal_norm}
    \sum_{\indexSourceModel=1}^{\numberSourceModels} w^{(\indexSourceModel)}_{t | t + k}=1 \qquad \forall k \in \naturalNumberPos.
\end{align}

The constraints ensure that the weights of all $\numberSourceModels$ ensemble members, where $\csgeWeightNormalized^{(\indexSourceModel)}_{t | t + k}$ denotes the weight of the $m$-th ensemble member's prediction with lead time $k$, are properly normalized.
The weight incorporating the global, local, and forecast horizon-dependent weight, for a single source model and lead time, is given by
\begin{align}
    \label{eq:coop_int_fusion:csge:overallweights_appendix}
    \bar{\csgeWeightNormalized}^{(m)}_{t + k | t} = \csgeWeightNormalized^{(m)}_g \cdot \csgeWeightNormalized^{(m,t)}_l \cdot \csgeWeightNormalized^{(m,k)}_{h}.
\end{align}
We achieve a normalization of $\bar{\csgeWeightNormalized}^{(m)}_{t + k | t}$ through the division of the sum of all weights for all source models and lead times:

\begin{align}
    \label{eq:coop_int_fusion:csge:overallWeighingFormular}
    \csgeWeightNormalized^{(m)}_{t + k | t} = \bar{\csgeWeightNormalized}^{(m)}_{t + k | t} \slash \sum_{\tilde{m}=1}^M \bar{\csgeWeightNormalized}^{(\tilde{m})}_{t + k | t}.
\end{align}

\subsubsection{Soft Gating Principle}
The soft gating principle to calculate a source model's weight has several theoretical advantages in the context of an ensemble for a target task from a model hub.
These advantages are all due to the compromise between the weighting and the gating principle of the~\ac{csge}.
On the one hand, we want gating, as properly many source models do not apply to a target task and should therefore not be considered.
On the other hand, those applicable source models should be combined through weighting.
In a sense, this is comparable to an~\ac{bma} approach, where models with a low posterior are neglected (gated), while those with a larger posterior get a larger weight.

To calculate the weights $\csgeWeightNormalized^{(\indexSourceModel)}_{t + k | t}$, we utilize the definition of the inverse similarity measurement $\similarityMeasure^{-1}$ from~\SEC{sec_model_selection_definition} and the coopetitive soft gating principle from~\EQ{eq_soft_gating_appendix}.
By calculating the weighting through the inverse $\similarityMeasure^{-1}$, here the~\ac{nrmse}, we directly measure how well a source model performs on the target.
Therefore, let us assume that $\mathbf{\Phi} \in \realNumber^{J}$ contains all $J \in \naturalNumberPos$ estimates based on the~\ac{nrmse} and $\phi$ is an arbitrary element from $\mathbf{\Phi}$.
Then, $\eta \geq 0$ depicts the amount of exponential weighting and the small constant $\epsilon > 0$ avoids division by zero.
For greater $\eta$, the \ac{csge} tends to work as a gating ensemble and, thereby, considering only a few source models.
For smaller $\eta$ result in a weighting ensemble.
After calculating all weights from $\mathbf{\Phi}$ through~\EQ{eq_soft_gating}, we normalize the results to sum up to one estimating the final weights.

\begin{align}\label{eq_soft_gating_appendix}
    \varsigma^{'}_\eta(\mathbf{\Phi}, \phi) = \frac{\sum_{j=1}^{J} \mathbf{\Phi}_j}{\phi^{\eta} + \epsilon},
\end{align}

Finally, to assure that calculated weights for each ensemble member are normalized, we use the function $\varsigma^{}_\eta : (\mathbb{R}^+)^{M+1} \rightarrow [0,1]$ with

\begin{align}
    \varsigma^{}_\eta(\mathbf{\Phi}, \phi) = \frac{ \varsigma^{'}_\eta(\mathbf{\Phi}, \phi, ) }{ \sum_{m=1}^{M}\varsigma^{'}_\eta(\mathbf{\Phi}, \phi_{m})}
\end{align}

\subsubsection{Global Weighting}
\label{sec:global}
The global weight expresses the overall performance of the different source models on the target based on the available training data.
The average $r^{(\indexSourceModel)} \in \realNumberPos$ of a single source model is given by
\begin{align}\label{eq:coop_int_fusion:csge:formelGlobaleGewichtungR}
    r^{(\indexSourceModel)} = & \, \frac{1}{\nSamples} \cdot \sum_{\indexSamples=1}^{\nSamples} \similarityMeasure^{-1} ( \outputPredictionVector^{(\indexSourceModel)}_\indexSamples, o_{\indexSamples} ),
\end{align}
where $\nSamples$ refers to the number of available training data, $\similarityMeasure^{-1}$ is the inverse similarity for the prediction $\outputPredictionVector^{(\indexSourceModel)}_\indexSamples$ and the respective observations $o_{\indexSamples}$.
Then, the set of all errors, from all source models, is given by $R =\, ( r^{(1)}, \ldots , r^{(\indexSourceModel)}, \ldots , r^{(\numberSourceModels)})$.
This set allows us to calculate the global weight by
\begin{align}
    w_g^{(m)} = \varsigma^{}_{\eta_{g}}(R, r^{(m)}),
\end{align}
with the hyperparameter $\eta_{g}$ for the soft gating formula.

\subsubsection{Local Weighting}
\label{sec:local}
The local weighting expresses how well a source model performs for different areas in the feature space.
For instance, one source model might perform well on the target for cloudy days while another performs well for sunny days.
To calculate an expectation of the error under certain conditions in the input features $\inputFeatureVector$, we train a regression model for each source model based on the available training data.
Once the regression model is trained, this function with $L^{(\indexSourceModel)}: \realNumber^{\dimFeature} \rightarrow \realNumberNonNeg$ lets us predict the expected error $\hat{q}^{(m)}_{t + k | t}$ for each ensemble member and each $k$ with

\begin{align}\label{knn_variante}
    \hat{q}^{(m)}_{t + k | t} = L^{(m)}(\inputFeatureVector_{t + k | t}).
\end{align}
Similar, to the global weighting, the set of all local errors 
\begin{equation*}
Q = ( \hat{q}^{(1)}_{t + k_\text{min} | t}, \ldots , \hat{q}^{(m)}_{t + k | t}, \ldots, \hat{q}^{(M)}_{t + k_\text{max} | t}),  
\end{equation*}
$k_\text{min}$ refers to the smallest forecast horizon and $k_\text{max}$ refers to the maximum forecast horizon, lets us calculate the local weight $w_l^{(j)}$ with
\begin{align}
    w_l^{(t,k)} =  \varsigma^{}_{\eta_{l}}( Q, \hat{q}^{(m)}_{t + k | t}),
\end{align}
where $\eta_{l}$ is the hyperparameter of the soft gating formula for the local weighting.

\subsubsection{Forecast horizon-Dependent Weighting}
\label{sec:time}
The time-dependent weighting considers that ensemble members may perform differently for different forecast horizons $k$, e.g., some source models might be better in the short-term while others are better in the long term.
Similar to global weighting, we calculate this weighting during training.
But instead of averaging across all training samples, we calculate the mean with

\begin{align}
    p^{(\indexSourceModel)}_k = & \, \frac{1}{\nSamples_k} \cdot \sum_{\indexSamples_k=1}^{\nSamples_k} \similarityMeasure^{-1} ( \outputPredictionVector^{(\indexSourceModel)}_{\indexSamples_k}, o_{\indexSamples_k} )
\end{align}
to obtain the forecast horizon-dependent error $p^{(\indexSourceModel,k)}$, where $\nSamples_k$ refers to the number of samples for a forecast horizon $k$.
All errors of all ensemble member are the combined through the set $P_k =\, ( p_k^{(1)} , \ldots, p_k^{(\indexSourceModel)},  \ldots, p_k^{(\numberSourceModels)} )$ to calculate the time-dependent weight through
\begin{align}
    w_{lt}^{(\indexSourceModel,k)} = & \, \varsigma^{}_{\eta_{lt}}(P_k, p_k^{(m)}).
\end{align}
The forecast horizon-dependent hyperparameter is here denoted with $\eta_{lt}$.

\clearpage
\section{Experiment}\label{sec_appendix_model_selection_experiment}

In the following, we detail our hyper-parameters for model training, compare them to a physical model for the REAL datasets, provide additional evaluations and example forecasts for the most promising methods.


\subsection{Hyperparameter Tuning of Multi-Layer Perceptron and Temporal Convolution Network Source Models} As source models for the model hub, we utilize an~\ac{mlp} and a~\ac{tcn}.
The layers' initialization strategy is the default one of the utilized libraries (\textit{pytorch\footnote{\url{https://pytorch.org/docs/stable/index.html}, accessed 2021-02-28}}, \textit{fastai\footnote{\url{https://docs.fast.ai/}, accessed 2021-02-28}}).
For all source models, we conducted a hyperparameter search through an tree-structured Parzen sampler for $200$ samples through the \textit{optuna\footnote{\url{https://optuna.org/}, accessed 2021-02-28}} library.
The best hyperparameters are selected based on the smallest~\ac{nrmse} on the validation dataset.
We initially transform the model's number of input features in a higher dimension through a factor $k \in \{1, \ldots, 20\}$.
Note that initially transforming the input features in a higher-dimensional space typically improves the performance~\cite{Sain2006}.
Afterward, the features are reduced by $50$ percent in each layer to a minimum of $11$ before the final output.
The final two layers of all networks have sizes $3$ and $1$.
The learning rate of the \textit{adam} optimizer is sampled from a logarithmic uniform distribution $\mathcal{U}(\ln(\SI{e-6}), \ln(\SI{e-1}))$.
For each model, we sample the number of training epochs from the set $\{10,\ldots,100\}$.
The dropout is sampled from $\mathcal{U}(0, 0.9)$ and the weight decay from $\{0.0,0.1,0.2,0.4\}$.
For the~\ac{mlp} the batch size is sampled from the set $\{1024,2048,4096\}$.
In the case of the~\ac{tcn} network architecture, these batch sizes are divided by the number of samples per day, $24$ for the \PO dataset and $96$ for the remaining datasets.

\subsection{Hyperparameter Tuning Bayesian Extreme Learning Machine Source Models}
The~\ac{belm} is optimized by sampling $200$ times through the tree-structured Parzen sampler and the best parameters are selected by the largest evidence on the validation data.
The number of hidden neurons is from $\mathcal{U}(10, 1000)$ to generate the random features.
As an activation function for the random features, one of \ac{relu}, sigmoid, \ac{relu} and sigmoid is sampled.
In the latter case, the results of the two activation functions are concatenated to generate a maximum of $2000$ random features.
Additionally to these random features, we give the hyperparameter tuning the option to include the original~\ac{nwp} features.
Alpha and beta of the Bayesian linear regression are optimized through empirical Bayes~\cite{Bishop2006}.

\subsection{Gradient Boosting Regression Tree Target Model}
\label{sec_selection_hyperparameters_gbrt}
To have a strong \baseline that generalizes well with the limited amount of data and is known to mitigate the effects of overfitting, we train a~\ac{gbrt} for each target model.
We optimize the~\ac{gbrt} for each park, season, and number of training data through a grid search by three-fold cross-validation.
Thereby, the learning rate is evaluated at $[\SI{e-6}, \num[round-precision=1]{3.1e-6}, \ldots,\num[round-precision=1]{3.1e-1}, 1]$.
The number of estimators is $300$ and the maximum depth is one of $[2,4,6,8]$.
Other values are the default ones of scikit-learn\footnote{\url{https://scikit-learn.org/}, accessed 2022-02-28}.

\subsection{Target Training Extreme Learning Machine}:
\label{sec_doe_target_belm}
For the adaptation of the~\ac{belm} model, we assume that given an appropriate model selection strategy, the prior $\alpha$ and $\beta$ of the Bayesian linear regression is also valid for the target.
In principle, this is a relatively strong assumption.
While the aleatoric uncertainty might be the same between a source and a target, the knowledge of the source model for the target is low and therefore, the epistemic uncertainty should be high.
We argue that the problem is reduced as we update the mean vector and the precision matrix, given the available target data through \EQ{eq_blr_online_update_mean}, where the previous posterior, from the source model, acts as prior for the target.
Another interesting consideration is that the insufficient evidence of the source model on the target requires that an update of the source model needs a large amount of target data which contracts to the training from the source model.
Therefore, it reduces the risk of catastrophic forgetting.

We first update all source models by the given target data to select a source model.
We train and select the source model with all the available data for the evidence selection strategy.
For the \ac{nrmse} selection, we use $70\%$ of the data for training and the remaining data for the selection.

\subsection{Target training of Multi-Layer Perceptron and Temporal Convolution Network}
\label{sec_doe_target_ann}
As pointed out earlier, we deploy different model selection and adaptation strategies, as summarized in~\TBL{tbl_model_selection_overview}.
In case we fine-tune a source model, e.g., through weight decay, we conduct a grid search for different hyperparameters and select the best one based on the lowest \ac{nrmse} through $30\%$ of the training data as a validation dataset.
The learning rate and the amount of penalty $\lambda$ are optimized for all adaptation strategies, as detailed earlier in this section.
For all optimizations, we use the stochastic gradient descent algorithm without momentum as suggested in~\cite{Li2020} for~\ac{tl}.
For the Bayesian tunning, we select the ten most relevant models selected through the evidence for regularization.
In all cases, the batch size is selected to have about ten iterations within an epoch and we train for a single epoch to limit the amount of training time.

For the direct adaptation, we utilize the source model without any adaptation.
In the case of the direct linear adaptation, we replace the final layer through a~\ac{blr} model(s).
These~\acp{blr} are trained through empirical Bayes.
For the~\ac{tcn} source model, we train one~\ac{blr} for each forecast horizon based on the extracted features from the source model from a specific forecast horizon and its response.



\subsection{Comparison to Physical Models}
As pointed out in~\SEC{sec_itl_selection_experiment} the \WS, \PS, \WO, \PO have been previously mentioned in~\cite{Schreiber2021} and \cite{VogtSynData2022}.
The \WR and \PR have not been mentioned previously.
Therefore, to assure that the~\ac{gbrt} \baseline is superior to physical models that are commonly utilized for parks with limited data, we detail a comparison in this section.
Generally, the experimental set-up is as detailed in~\SEC{sec_itl_selection_experiment}.
Hyper-parameters of the GBRT model are detailed in~\SEC{sec_selection_hyperparameters_gbrt}.
For the \WR dataset, we compare the GBRT results for the different amounts of training data to the empirical \textit{McLean} power curve~\cite{Mclean2008}.
This power curve is well known and considers typical influences by different terrains.
For the \PR dataset we created forecasts through the \textit{PVLib}\footnote{\url{https://pvlib-python.readthedocs.io/}, accessed 2022-02-28}.
For both datasets, it is not feasible to utilize more specific physical models, as details on the physical characteristics are not given.

\begin{figure}[ht]
    \centering
    \includegraphics[width=\textwidth]{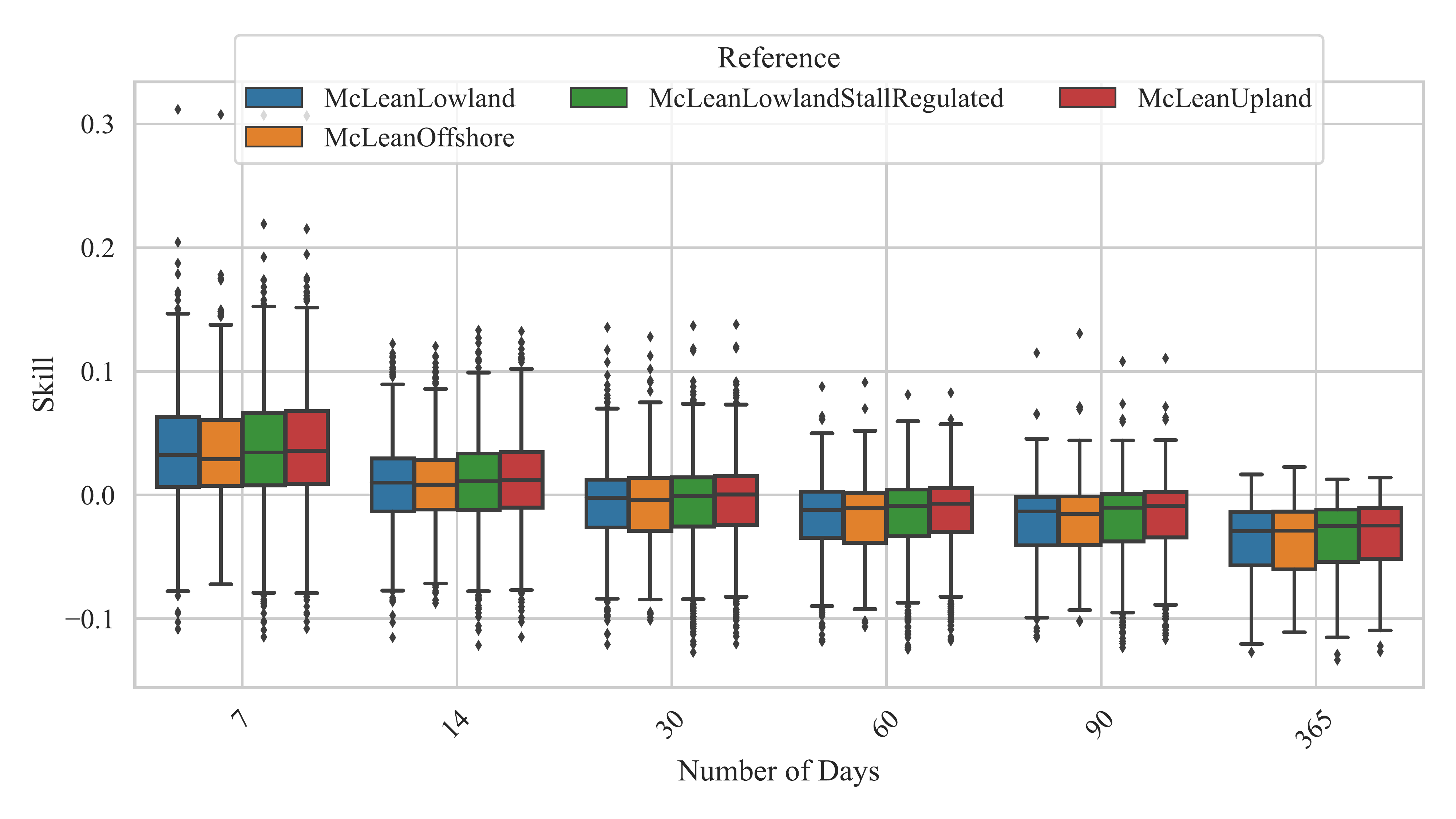}
    \caption{Comparison of physical model and the GBRT model for the WINDREAL dataset for different amounts of number of days training data through the skill. Values below zero indicate an improvement of the GBRT over the physical model.}
    \label{windreal_physical_skill_phyiscal}
\end{figure}

\begin{figure}[ht]
    \centering
    \includegraphics[width=\textwidth]{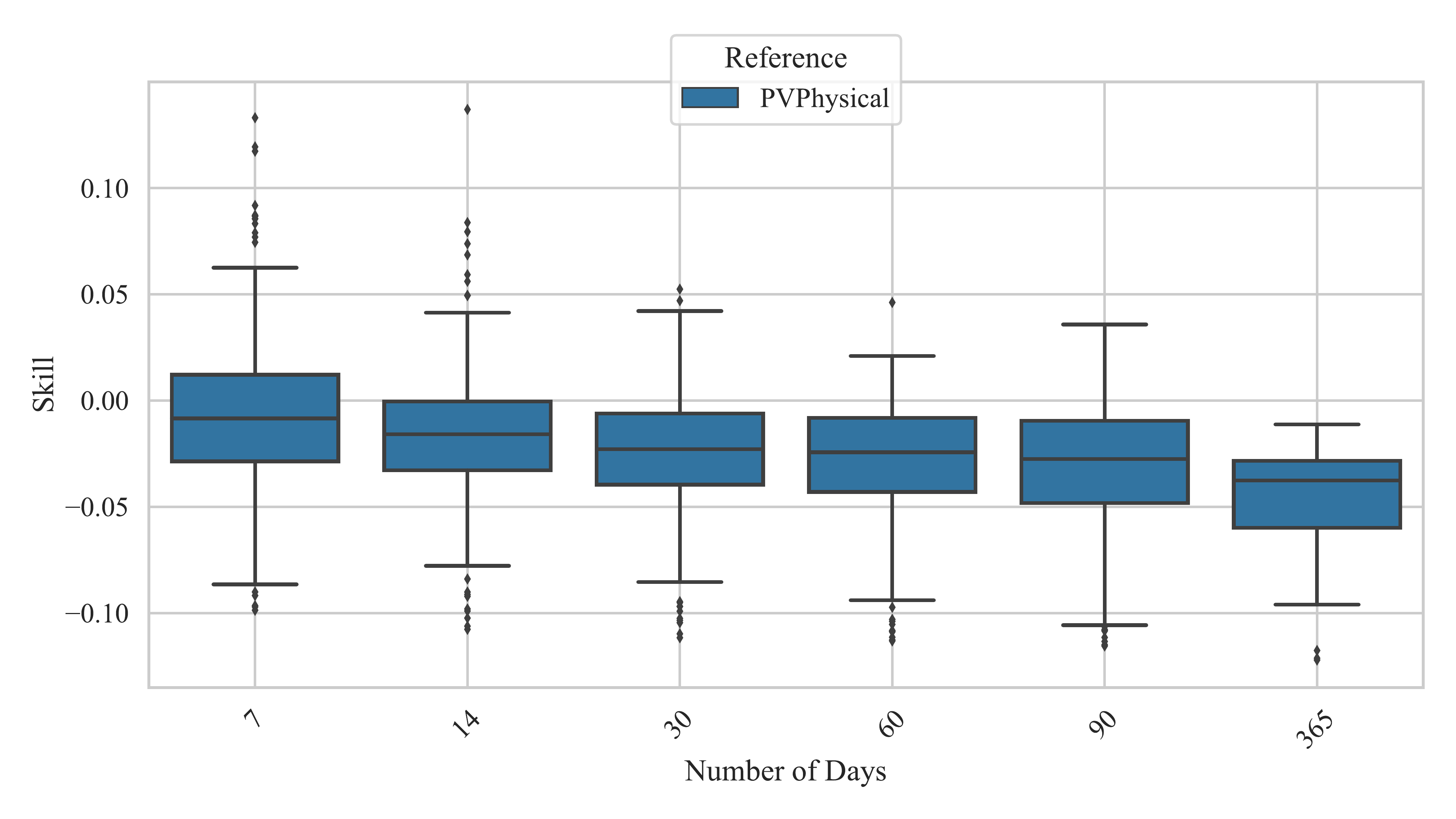}
    \caption{Comparison of physical model and the GBRT model for the PVREAL dataset for different amounts of number of days training data through the skill. Values below zero indicate an improvement of the GBRT over the physical model.}
    \label{pvreal_physical_skill_phyiscal}
\end{figure}

\FIG{windreal_physical_skill_phyiscal} compares the physical models to the GBRT model by means of the skill, for the \WR dataset, given by
\begin{equation}
    \text{Skill} = \text{nRMSE}_{gbrt}-\text{nRMSE}_{phy},
\end{equation}
where $\text{nRMSE}_{gbrt}$ is the \ac{nrmse} of the~\ac{gbrt} model and $\text{nRMSE}_{phy}$ of the physical model, respectively.
We can observe that the~\ac{gbrt} leads to improvements for the \WR dataset starting with $30$ days of training data.
With more than $30$ days of training data, there is a substantial improvement compared to all types of the McLean power curve.
The mean \ac{nrmse} for the \WR dataset is given in~\TBL{windreal_physical_nrmse}.
In comparison to the analysis of forecast errors in~\cite{Schreiber2020Prophesy}, we observe a similarly mean~\ac{nrmse} with $30$ days of training data.
Note that the experimental set-up in~\cite{Schreiber2020Prophesy} utilizes more than $365$ days of training data.
Results for the \PR dataset are given~\FIG{pvreal_physical_skill_phyiscal} and~\TBL{pvreal_physical_nrmse}.
Substantial improvements of the~\ac{gbrt} model over the physical PV model is given by more than seven days.
The mean forecast errors are similar to those in~\cite{Schreiber2020Prophesy} between $30$ and $60$ days of training data.
Due to forecast errors being similar to results in~\cite{Schreiber2020Prophesy} with limited data, we can conclude that the~\ac{gbrt} is a strong \baseline for~\ac{itl} for the \WR and \PR dataset.

\begin{table}[tb]
    \centering
    \caption{Mean nRMSE values for the GBRT model and McLean power curves for the WINDREAL dataset. Best values for the number of available training data in days is highlighted in bold.}
    \label{windreal_physical_nrmse}
    \begin{tabular}{|l | r | r| r| r| r|}
\toprule
{} &   GBRT &  McLean- &  McLean- &  McLean- &  McLean- \\
{} &    &  Lowland &  LowlandStallRegulated &  Offshore &  Upland \\
\#Days &        &                &                              &                 &               \\
\midrule
7              &  0.190 &          0.155 &                        0.154 &           0.156 &         \textbf{0.152} \\

14             &  0.164 &          0.155 &                        0.154 &           0.156 &         \textbf{0.152} \\

30             &  \textbf{0.149} &          0.155 &                        0.154 &           0.156 &         0.152 \\

60             &  \textbf{0.140} &          0.157 &                        0.156 &           0.158 &         0.154 \\

90             &  \textbf{0.137} &          0.157 &                        0.155 &           0.157 &         0.153 \\

365            &  \textbf{0.122} &          0.159 &                        0.157 &           0.160 &         0.155 \\
\bottomrule
\end{tabular}

\end{table}
\begin{table}[ht]
    \centering
    \caption{Mean nRMSE values for the GBRT and physical model for the PVREAL dataset. Best values for the number of available training data in days is highlighted in bold.}
    \label{pvreal_physical_nrmse}
    \begin{tabular}{|l|r|r|}
\toprule
{} &   GBRT &  PVPhysical \\
\#Days &        &             \\
\midrule
7              &  \textbf{0.133} &       0.137 \\
14             &  \textbf{ 0.120} &       0.137 \\
30             &   \textbf{0.112} &       0.137 \\
60             &   \textbf{0.107} &       0.137 \\
90             &   \textbf{0.105} &       0.137 \\
365            &   \textbf{0.091} &       0.137 \\
\bottomrule
\end{tabular}

\end{table}

\clearpage
\subsection{Mean Forecast Errors}
\begin{figure}
    \centering
    \includegraphics[height=0.95\textheight,trim=6cm 0cm 8.0cm 12cm, clip]{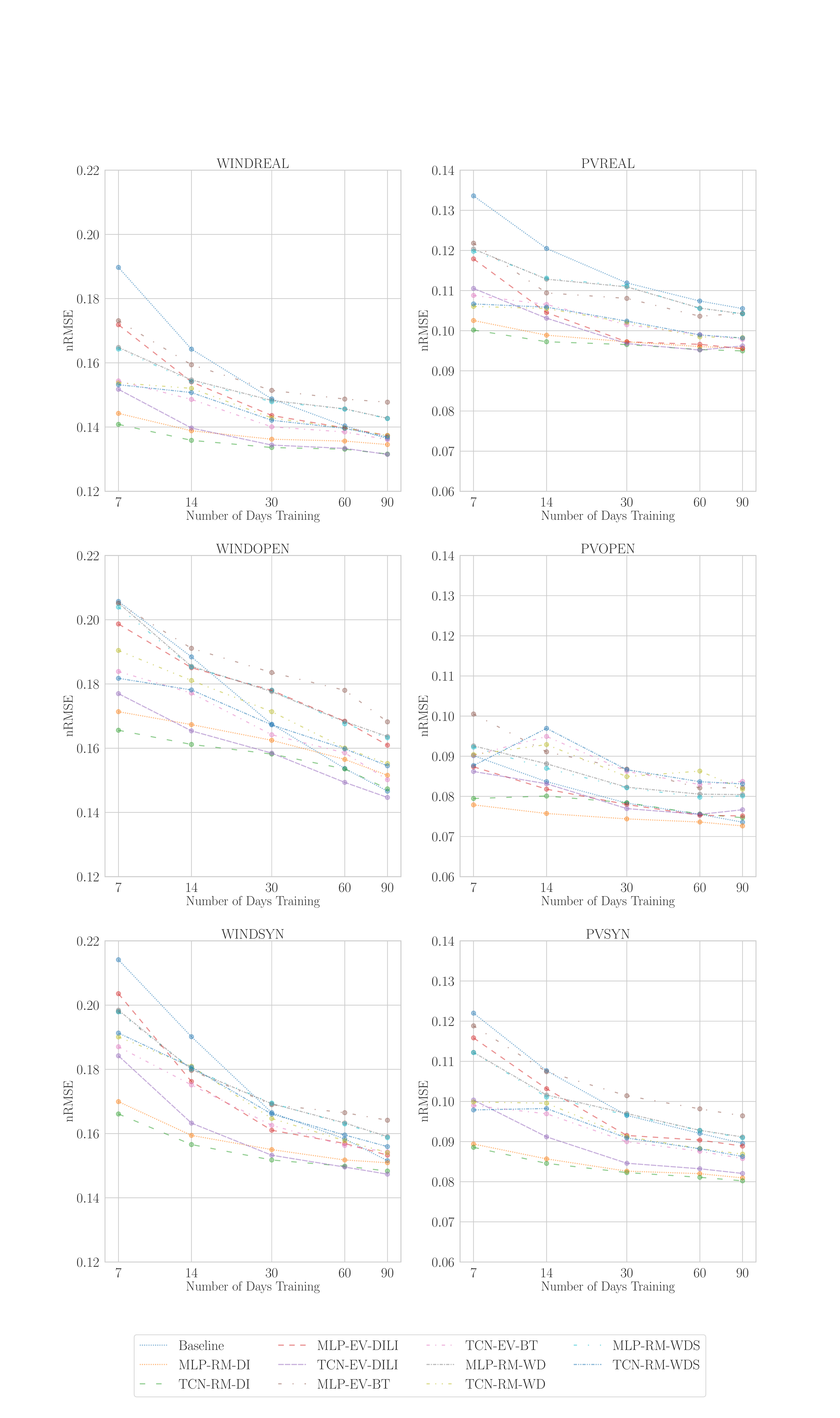}
    \caption{Mean nRMSE of all folds for most relevant techniques for the model selection and adaptation experiment.}
    \label{fig_model_selection_and_adaption_mean_nrmse}
\end{figure}

\FIG{fig_model_selection_and_adaption_mean_nrmse} summarizes the mean nRMSE values for the model selection and adaptation experiment for all six datasets.
For clarity of the plots, we show only the most relevant models.
All models decrease with increasing available training data for all six datasets.
For the wind datasets, the \WS dataset has the largest errors.
For the PV datasets, the \PR dataset has the largest error.
Except for the \PR dataset, the GBRT \baseline has a similar nRMSE with $14$ days of training data, showing that this model is a strong \baseline.
With increasing training samples, all models, selection strategies, and adaptation strategies converge to a similar minimum.
The difference between the best and the worse model can be up to five percent for seven days of training data.
With $90$ days of training data, the difference is between one and two percent. 
Models without any adaptation, the \textit{direct} adaptation strategy, start with a low forecast error but converge to a plateau in most cases with $30$ days of training samples.
In contrast, the models with adaptation strategies start with a larger mean forecast error but improve upon the direct adaptation with equal or more than $30$ days.
We can also observe that fine-tuning adaptation strategies require more data to achieve good results.
This observation is potentially caused by the problem that knowledge from a single source task is too specific and causes catastrophic forgetting in various cases.

\begin{figure}
    \centering
    \includegraphics[height=0.95\textheight,trim=6cm 0cm 8.0cm 12cm, clip]{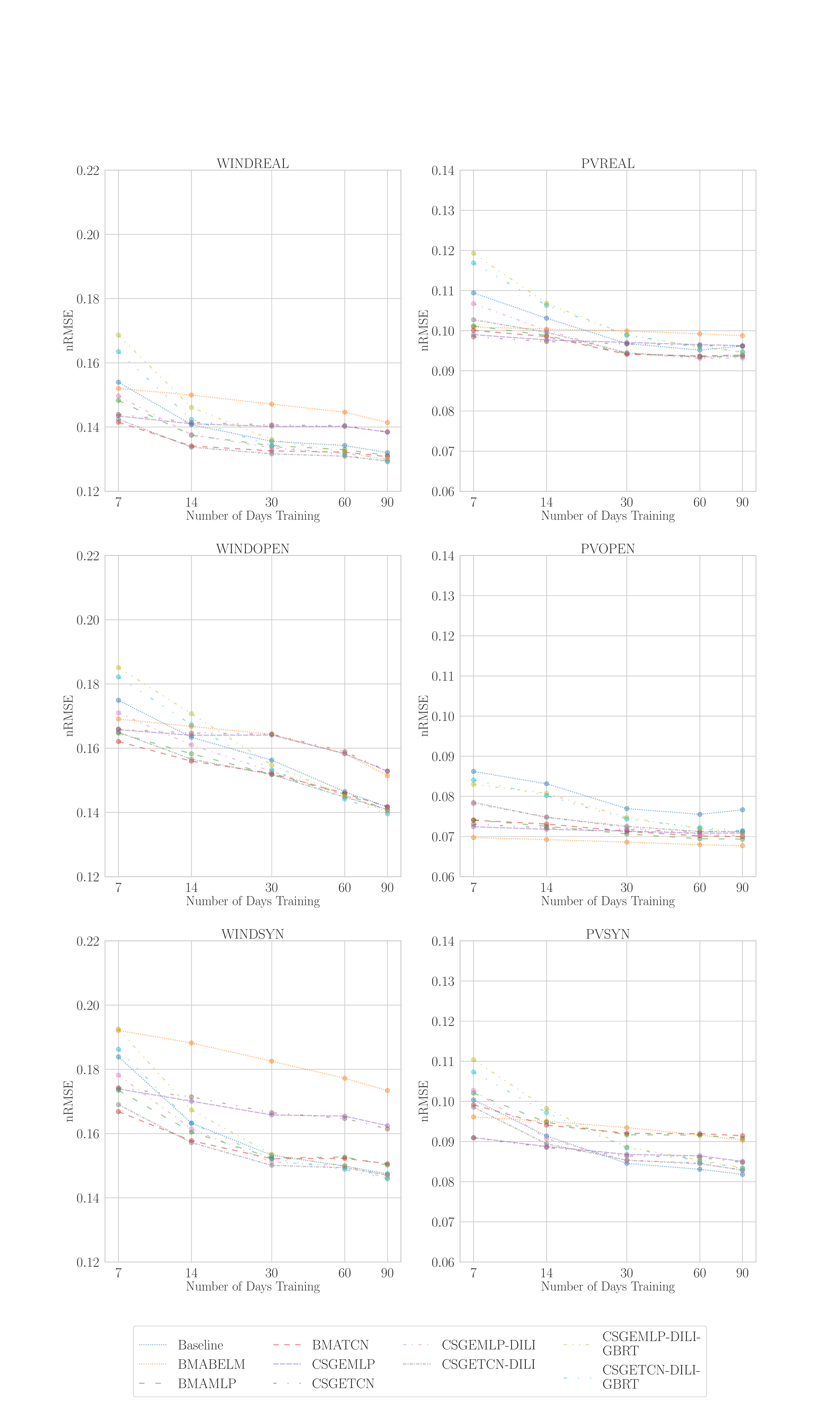}
    \caption{Mean nRMSE of all folds for the model combination experiment.}
    \label{fig_model_combination_mean_nrmse}
\end{figure}

The improvements of the mean~\ac{nrmse} through model combination is summarized in~\FIG{fig_model_combination_mean_nrmse}.
The \baseline is the TCN-EV-DILI model from the previous experiment.
Compared to the ensemble techniques results, the \baseline results vary depending on the dataset and the amount of available training data.
Nonetheless, often the \baseline has at least the fifth-most minor forecast error.
Comparing the worst models of the two experiments, we can observe improvements of more than two percent and at least one percent in most cases.
The forecast errors between the different ensemble techniques are less than the differences in techniques in the previous experiment.


An interesting comparison of our best ensemble techniques to the state-of-the-art for the \WO and \PO datasets are in~\TBL{tbl_model_seclection_comparions_mtl}.
The authors of these results~\cite{Schreiber2021} utilize an~\ac{mtl} approach.
As also explained in~\SEC{sec_itl_selection_related_work},~\ac{mtl} approaches are not yet common in the industry, and therefore it is necessary to make the best use of existing~\ac{stl} models.
The table shows that the error is in a similar range for the \WO and \PO datasets.
For the \PO dataset, we have better results, while with more than $30$ days of training data, the~\ac{mtl} approach is better for the \WO dataset.
However, note that the table compares our mean~\ac{nrmse} from all seasons to values from results for the spring seasons given in~\cite{Schreiber2021}, as other values are not given.
Nonetheless, the similarity of the forecast errors suggests that our approach leads to similar good results.

\begin{table}
\centering
\caption{Results of our best ensemble techniques, given in parenthesis, in comparison to the current state of the art~\cite{Schreiber2021} in inductive transfer learning for the WINDOPEN and PVOPEN datasets. The authors of~\cite{Schreiber2021} utilize a multi-task learning approach and mean nRMSE values from the spring season in~\cite{Schreiber2021} are compared to the mean values across all seasons from our experiments. $\downarrow$ indicates that our best ensemble model has a smaller mean nRMSE and $\uparrow$ indicates that our best ensemble technique has a worse error. }
\label{tbl_model_seclection_comparions_mtl}
\begin{tabular}{l|l|l|l|l|l|l|} 
\cline{2-7}
 & \diagbox{{}DataType}{{}\#Days} & 7     & 14    & 30    & 60    & 90     \\ 
\cline{2-7}
\cline{2-7}
 & PVOPEN                             & 0.096& 0.083 (0.074)$\downarrow$ & 0.078 & 0.076& 0.077 \\
 &                              &  (0.075) $\downarrow$ &  (0.074) $\downarrow$ &  (0.073) $\downarrow$& (0.072) $\downarrow$ &  (0.071) $\downarrow$ \\
 & WINDOPEN                           & 0.161 & 0.173 & 0.152  & 0.141& 0.137  \\ 
 &                            & (0.164) $\uparrow$  &  (0.160) $\downarrow$ &  (0.167) $\uparrow$ & (0.148) $\uparrow$  & (0.143) $\uparrow$ \\ 
\cline{2-7}
\end{tabular}
\end{table}

\clearpage
\subsection{Model Selection and Adaptation}\label{sec_appendix_exp_model_selection_and_adaption}
Previously in~\SEC{sec_itl_selection_experiment}, we discussed the best model selection and adaptation strategies through the mean rank.
\TBL{tbl_model_selection_rank_pv_worse} and \ref{tbl_model_selection_rank_wind_worse} lists those combination, from the overview in~\TBL{tbl_model_selection_overview}, previously not discussed.
We additionally include the \textit{TCN-EV-WD} model, as it allows a better comparison across all evaluated fine-tuning techniques.
We calculate the mean rank separate from the best models to better quantify the quality within the worse models.
Again, we test for a significant improvement compared to the~\ac{gbrt} \baseline.
We evaluate the influence of available training data on the different strategies.

For all datasets, we observe that~\ac{tcn} based models have a smaller rank than~\ac{mlp} based models and adaptations.
For both datasets, only one model is included that is not considering fine-tuning, the \textit{MLP-EV-DI} model, which is not adapted to the target.
This model is the best for PV datasets with limited data for up to $14$ days.
Additional results for the PV datasets can be summarized as follows:
\begin{itemize}
    \item The~\ac{mlp} model has the best results when the weight decay source adaptation strategy is utilized for PV datasets. A weight decay concerning the origin has a larger rank in most cases.
    \item For the \PO dataset, models are not significantly better than the \baseline with more than $14$ days of training data.
    \item For \PS and \PR, the TCN model with the evidence selection strategy and weight decay source adaptation leads to the best significant improvements.
    \item The Bayesian tuning is among the best for the~\ac{tcn} model but leads to one of the worst results for the~\ac{mlp} model.
\end{itemize}
~\newline
The results for the wind datasets can be summarized as follows
\begin{itemize}
    \item For the MLP and the selection through~\ac{nrmse} there is no clear winner of the adaptation strategies.
    \item For the MLP and the selection through~\ac{nrmse} with limited data, the weight decay source is preferable; with more data, the weight decay leads to better results.
    \item For the MLP and the adaptation through Bayesian tunning, results are worse than those from other selection and adaptation strategies.
    \item For the TCN, the Bayesian tuning has the best results.
    \item The second best technique for the TCN is selecting through the evidence and a weight decay with respect to the origin.
    \item No model, selection, and adaptation is superior to the \baseline with $90$ days of training data.
\end{itemize}

Generally, we can observe that for the~\ac{mlp} model, a selection through the~\ac{nrmse} is preferable over the evidence, regardless of the fine-tuning technique.
The opposite is the case for the~\ac{tcn} model.
Here, we observe that in most cases, it is preferable to select a source model based on the evidence instead of the~\ac{nrmse}
This observation suggests that the~\ac{nrmse} correlates better with the transferability of the final layer than for the evidence for the~\ac{mlp}.
For the~\ac{tcn}, the final layer is a residual block and the evidence better captures this complexity.

\begin{table}[tb]
    \centering
    \caption{Rank summary for different source models, selections, and adaptation strategies on the PV datasets. Only those that are \textit{not} within the top four ranks for a dataset are included. GBRT is the baseline and all models are tested if the forecasts error is significantly better ($\vee$), worse ($\wedge$), or not significantly different ($\diamond$). EV indicates the model selection by the evidence, RM the one by the nRMSE. DI shows that a source model is directly applied to the target without any adaptation. DILI indicates the adaptation though the~\ac{blr}, WD indicates fine-tuning regularized by weight decay w.r.t. the origin, WDS indicates fine-tuning regularized by weight decay w.r.t. the source models parameter, and BT indicates Bayesian tunning.}\label{tbl_model_selection_rank_pv_worse}
    \includegraphics[width=1\textwidth,trim=0.1cm 0cm 0.0cm 0cm, clip]{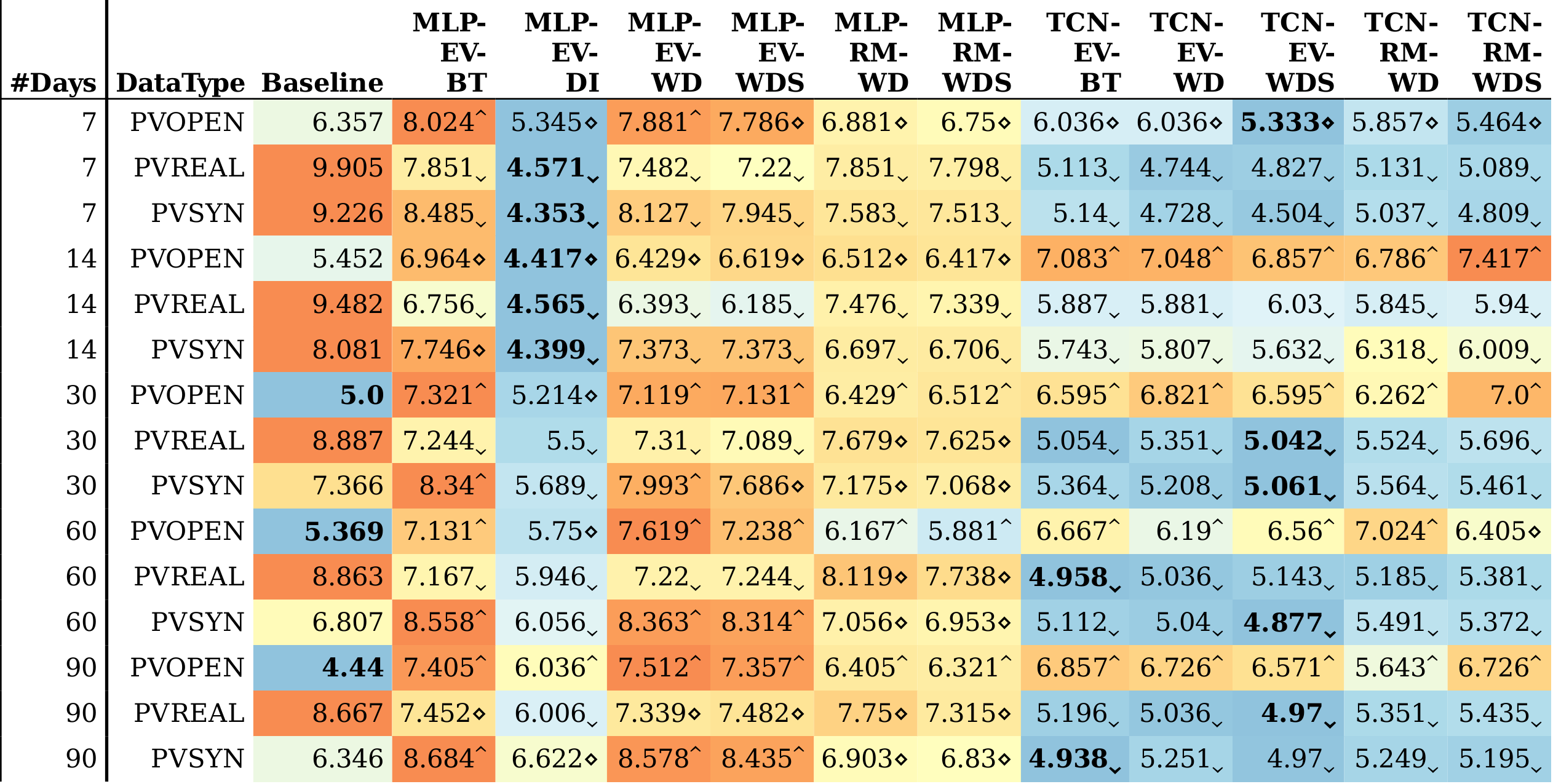}
\end{table}
\begin{table}[ht]
    \centering
    \caption{Rank summary for the wind datasets. Compare~\TBL{tbl_model_selection_rank_pv_worse}.}\label{tbl_model_selection_rank_wind_worse}
    \includegraphics[width=1\textwidth,trim=0.1cm 0cm 0.0cm 0.cm, clip]{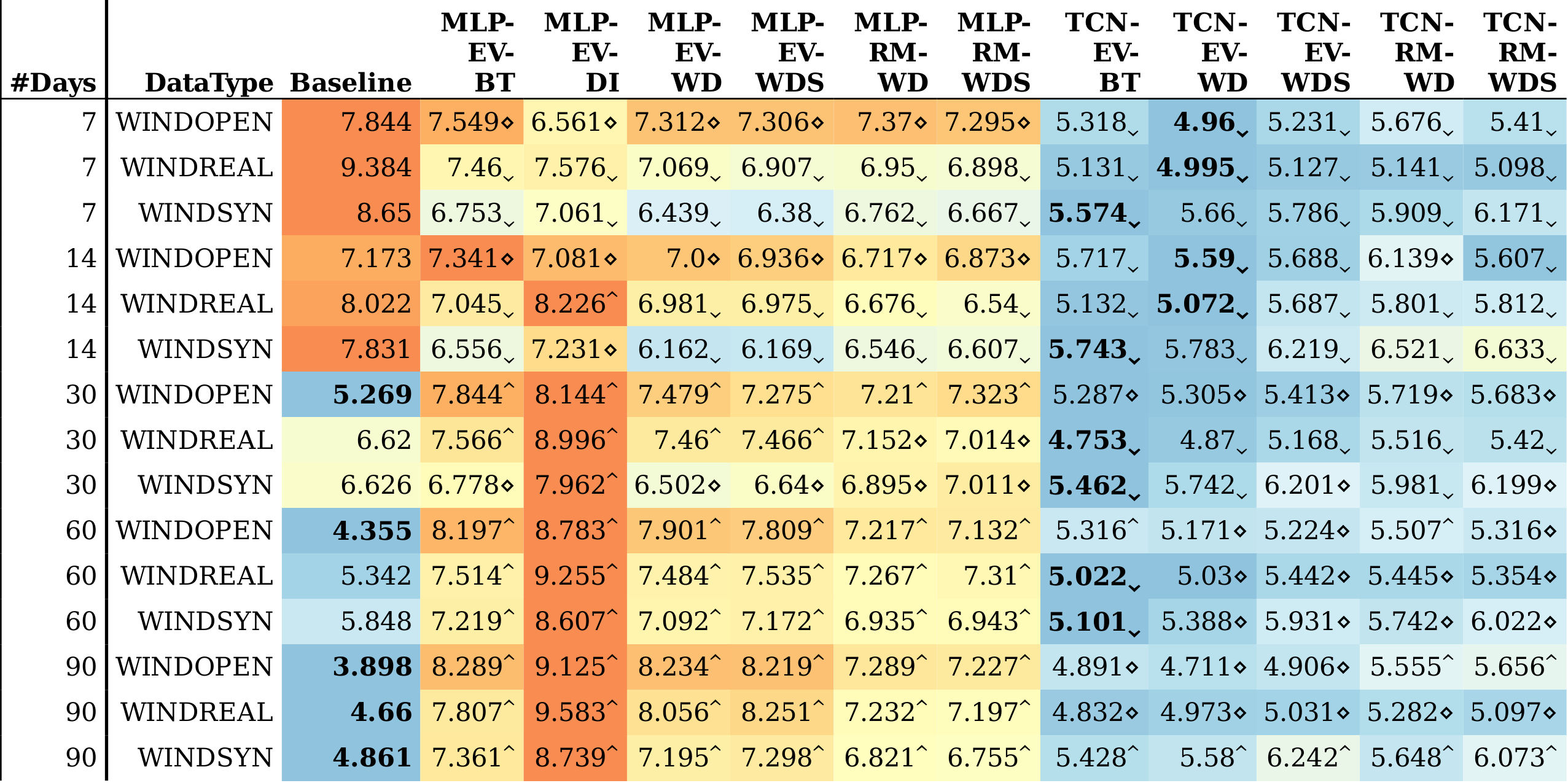}
\end{table}

\clearpage
\subsection{Probabilistic Evaluation of Models}
In the following, we list results for probabilistic forecasts that are utilizable through~\ac{blr}.
The evaluation include \textit{MLP-EV-DILI} and the \textit{TCN-EV-DILI} models.
We also include the combination of those models through the~\ac{bma}.
Finally, we consider the~\ac{belm} where the source model is either selected through the~\ac{nrmse} and or the evidence.
The latter is the \baseline.
All models are evaluated through the~\ac{crps} given by:
\begin{equation}
    CRPS(F,y)=\int_\mathbb{R} ( F(x) - \mathbb{H}(x \ge y) ) ^2 dx,
\end{equation}
where $y \in \mathcal{R}$ are observations, $F$ is the predicted cumulative function, and $\mathbb{H}$ is the Heaviside step function.
Similar to the previous experiments, the mean ranks of the~\ac{crps} for all models is given in~\TBL{tbl_model_selection_rank_pv_crps} and~\ref{tbl_model_selection_rank_wind_crps}.
The mean~\ac{crps} values are given in~\TBL{tbl_model_selection_mean_pv_crps} and~\ref{tbl_model_selection_mean_wind_crps}.
For the \PS and \PR datasets, the TCN-EV-DILI has the best results with more than $14$ days of training data improving the \baseline significantly.
In case of the \PO dataset, the BMABELM has the best rank in most cases.

With less than $60$ days of limited data, the ensemble technique BMATCN has the best results for all wind datasets. 
Through additional data, again, the TCN-EV-DILI has the best results.
This result holds for all datasets and the different number of training data for the wind dataset, except for one case.

The mean CRPS for the \WO dataset are worse than those presented~\cite{GS17} by $8.2$ percent.
Even though results from~\cite{GS17} use about six times more data compared to our experiment with $90$ days of training data, this observation suggestion that additional evaluations need to be conducted in future work.
\begin{table}[ht]
    \centering
    \caption{Rank summary of the CRPS for different ensemble types on the PV datasets. BELM-EV is the baseline. DI and DILI are the direct and direct linear adaptation as ensemble members. BMA models are always updated on the target by the~\ac{blr}. Compare also~\TBL{tbl_model_selection_rank_pv}.}\label{tbl_model_selection_rank_pv_crps}
    \includegraphics[width=1\textwidth,trim=0.1cm 0cm 0.0cm 0cm, clip]{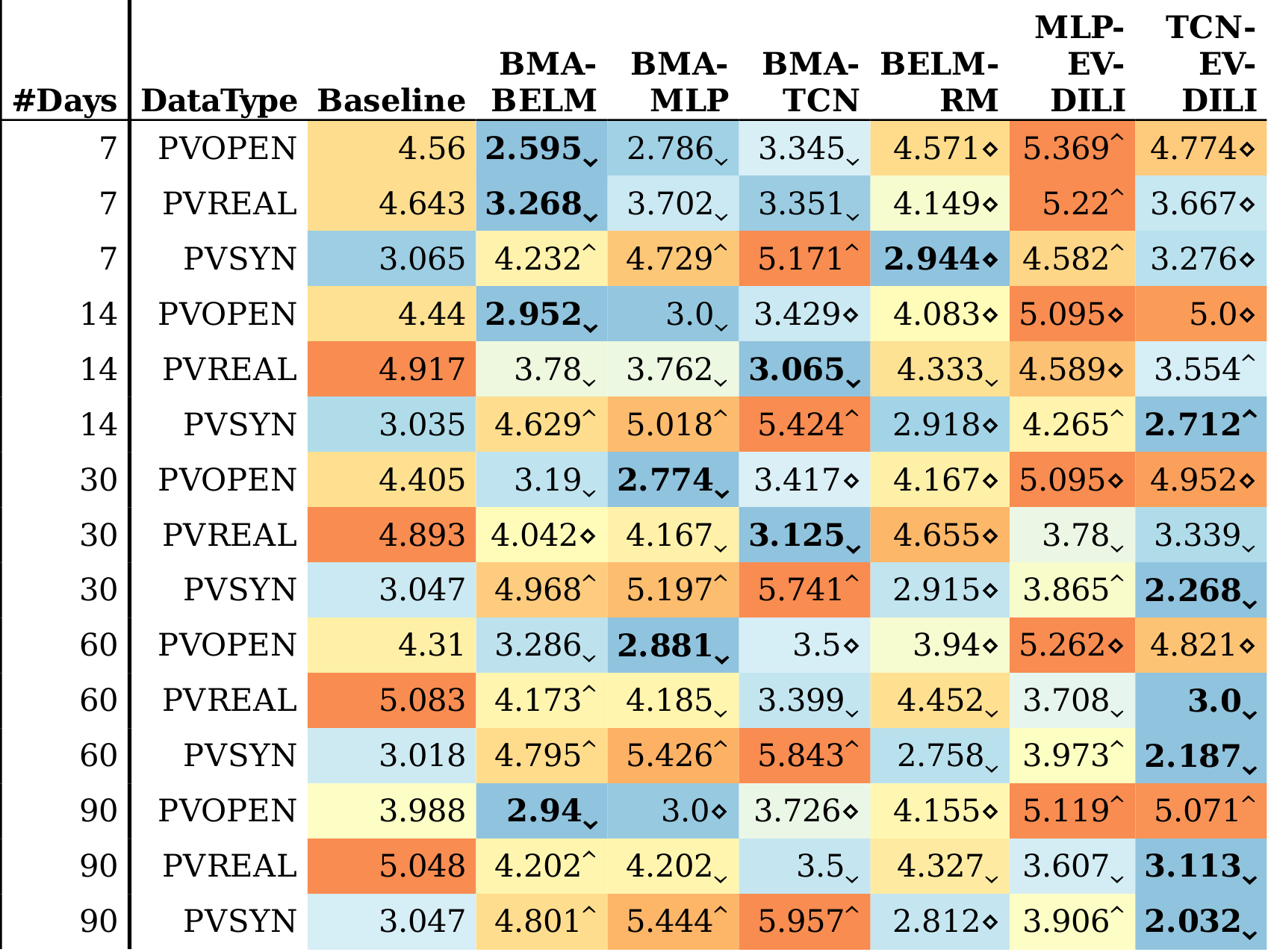}
\end{table}
\begin{table}[ht]
    \centering
    \caption{Rank summary of the CRPS for different ensemble types on the WIND datasets. BELM-EV is the baseline. DI and DILI are the direct and direct linear adaptation as ensemble members. BMA models are always updated on the target by the~\ac{blr}. Compare also~\TBL{tbl_model_selection_rank_pv}.}\label{tbl_model_selection_rank_wind_crps}
    \includegraphics[width=1\textwidth,trim=0.1cm 0cm 0.0cm 0cm, clip]{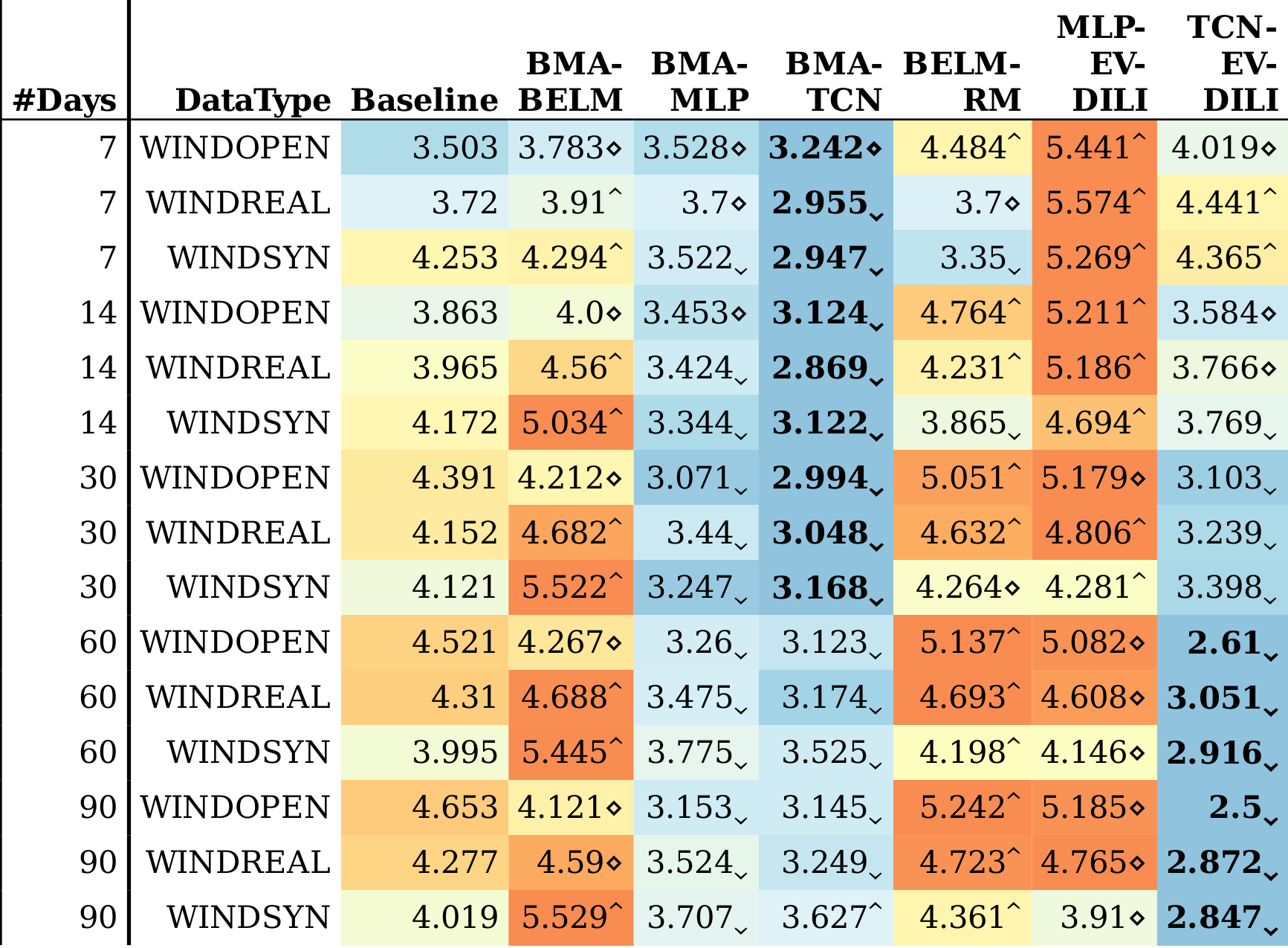}
\end{table}
\begin{table}[ht]
    \centering
    \caption{Mean CRPS values for different ensemble types on the PV datasets. BELM-EV is the baseline. DI and DILI are the direct and direct linear adaptation as ensemble members. BMA models are always updated on the target by the~\ac{blr}. Compare also~\TBL{tbl_model_selection_rank_pv}.}\label{tbl_model_selection_mean_pv_crps}
    \includegraphics[width=1\textwidth,trim=0.1cm 0cm 0.0cm 0.0cm, clip]{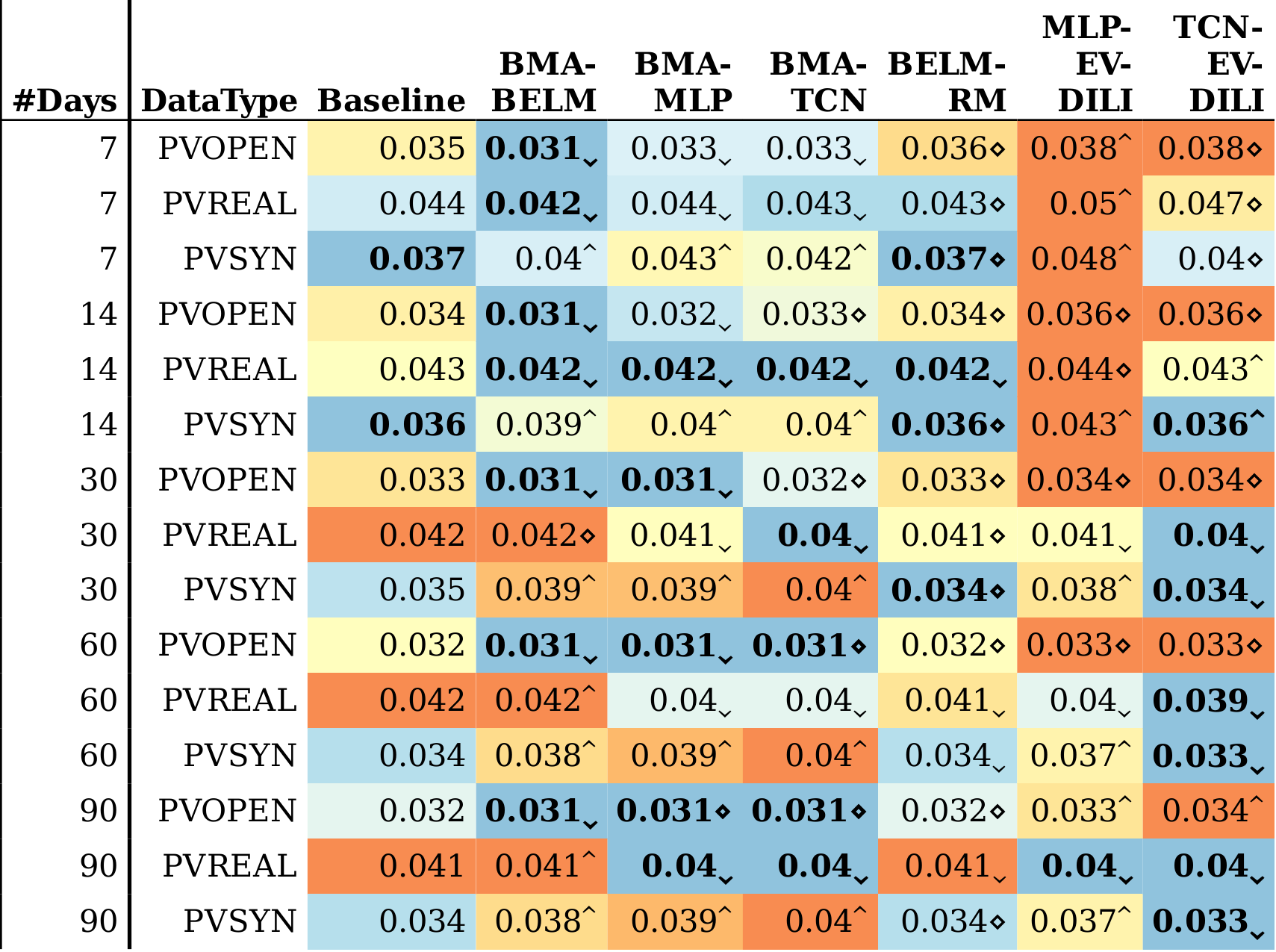}
    \end{table}
\begin{table}[ht]
    \centering
    \caption{Mean CRPS values for different ensemble types on the PV datasets. BELM-EV is the baseline. DI and DILI are the direct and direct linear adaptation as ensemble members. BMA models are always updated on the target by the~\ac{blr}. Compare also~\TBL{tbl_model_selection_rank_pv}.}\label{tbl_model_selection_mean_wind_crps}
    \includegraphics[width=1\textwidth,trim=0.1cm 0cm 0.0cm 0cm, clip]{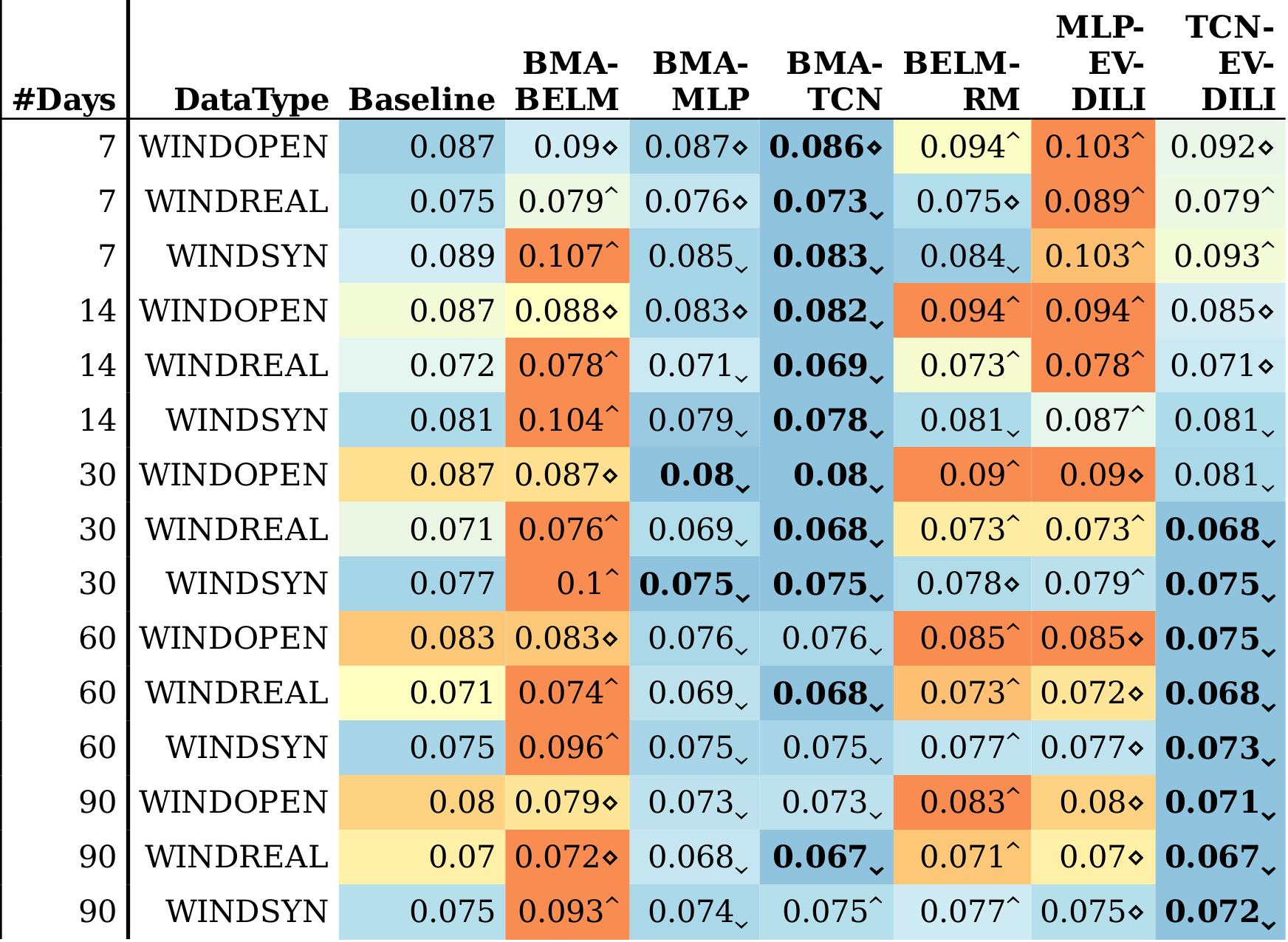}
\end{table}
\clearpage
\subsection{Deterministic and Probabilistic Forecasts}
The following figures give an impression of the different forecasting techniques.
All plots include power generation and forecasts from $14$ days from the real-world datasets.
We include the standard deviation as a confidence interval for the probabilistic forecasts.
Often these probabilistic forecasts have a large uncertainty when the power generation is close to zero.
The PV dataset shows the sun's diurnal cycle in the form of a Gaussian curve.
Such a periodic pattern is not present in the wind dataset.
\begin{figure}
    \centering
    \includegraphics[width=0.95\textwidth]{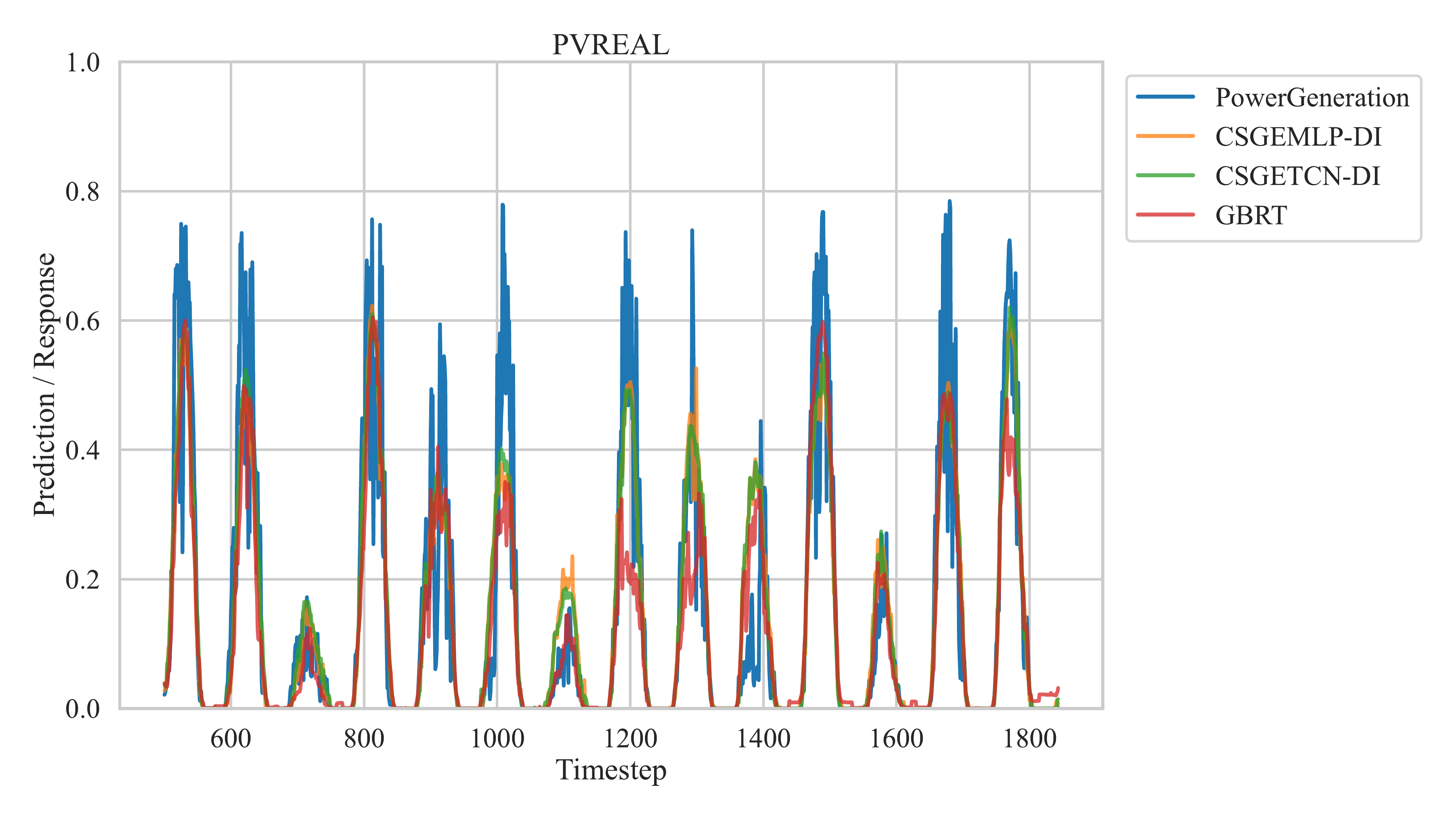}
    \caption{Example of a deterministic forecast for a PV Park.}
\end{figure}
\begin{figure}
    \centering
    \includegraphics[width=0.95\textwidth]{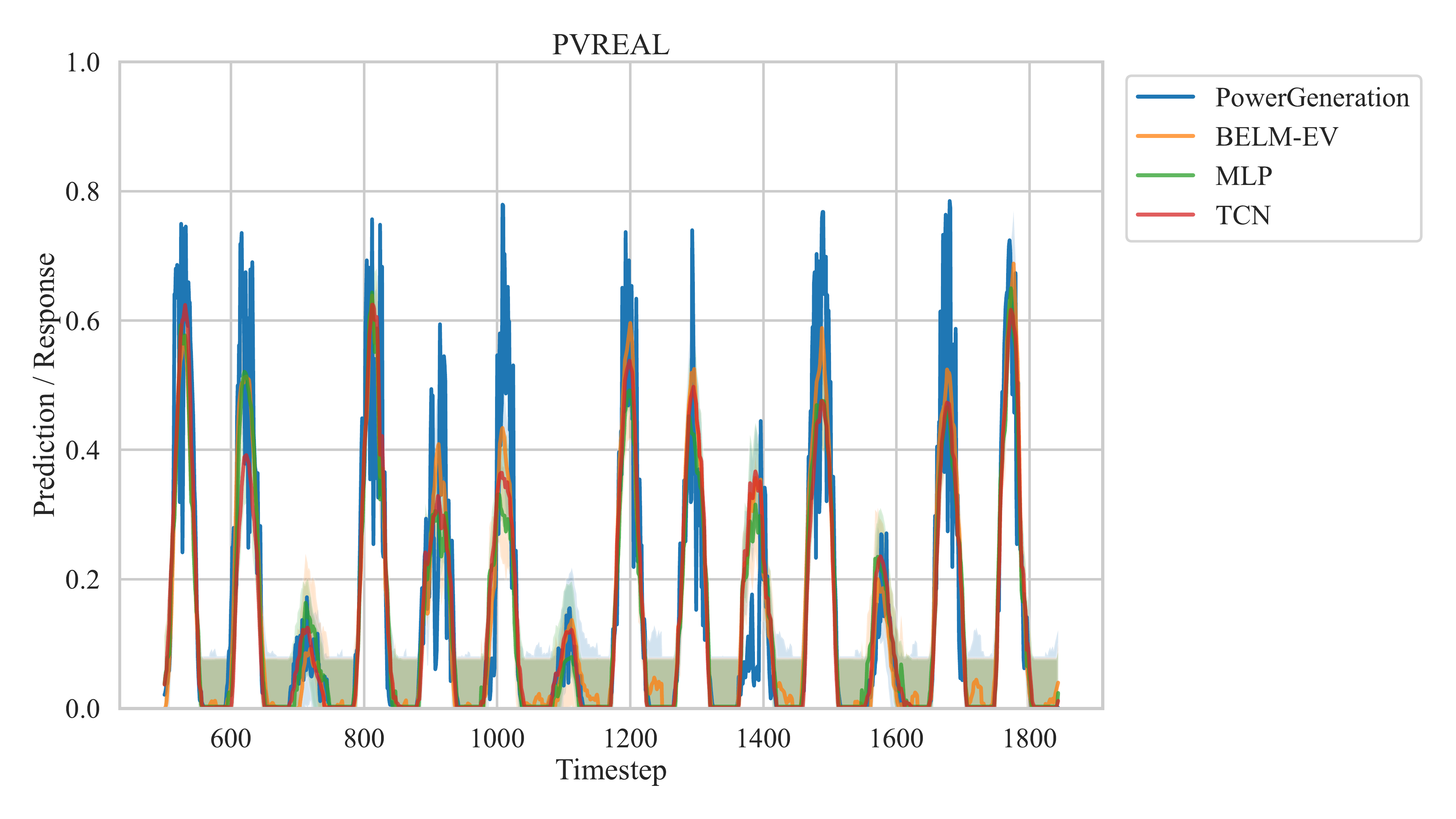}
    \caption{Example of a probabilistic forecast from a single model for a PV Park.}
\end{figure}
\begin{figure}
    \centering
    \includegraphics[width=0.95\textwidth]{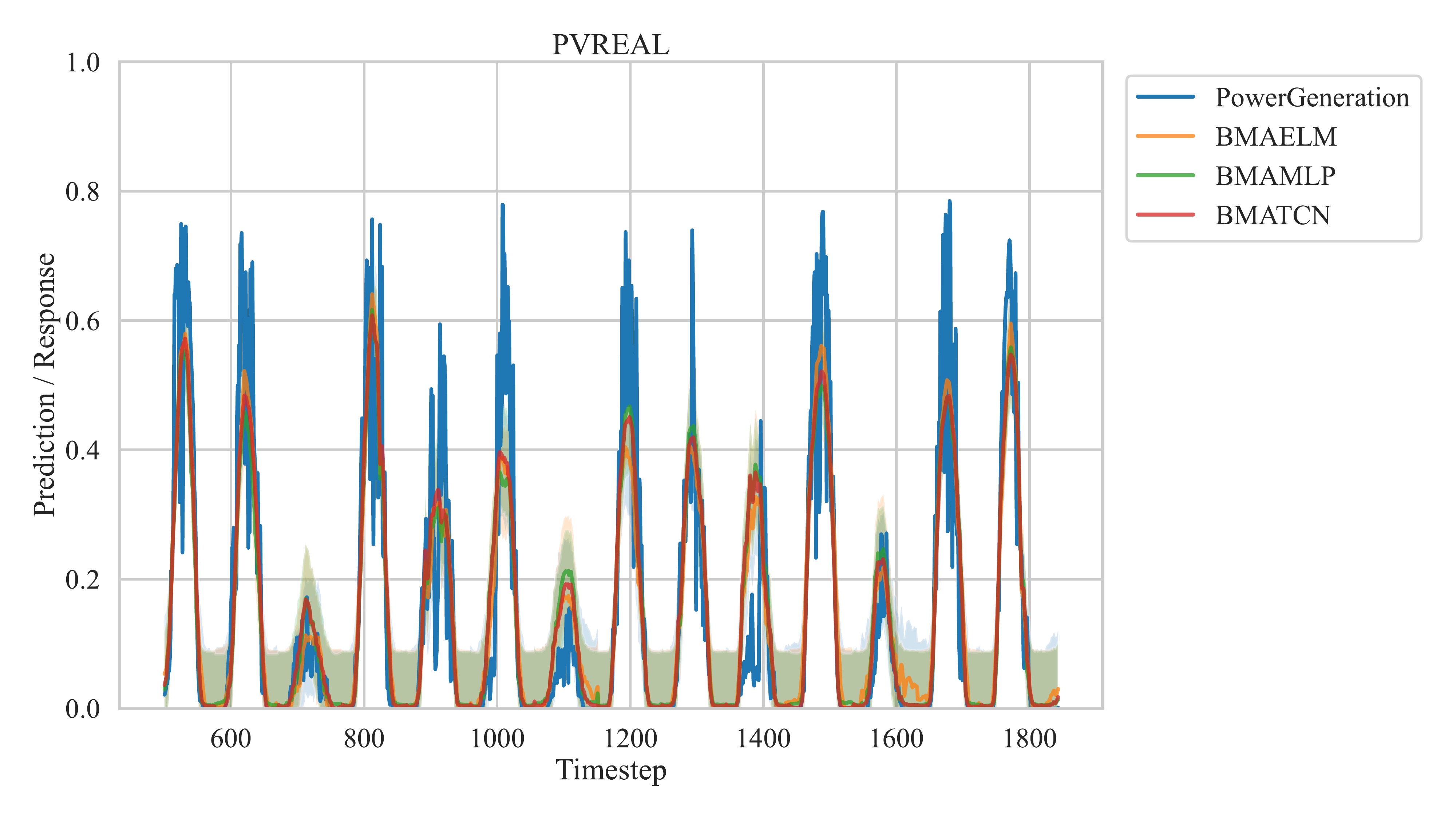}
    \caption{Example of a probabilistic forecast based on an ensemble for a PV Park.}
\end{figure}
\begin{figure}
    \centering
    \includegraphics[width=0.95\textwidth]{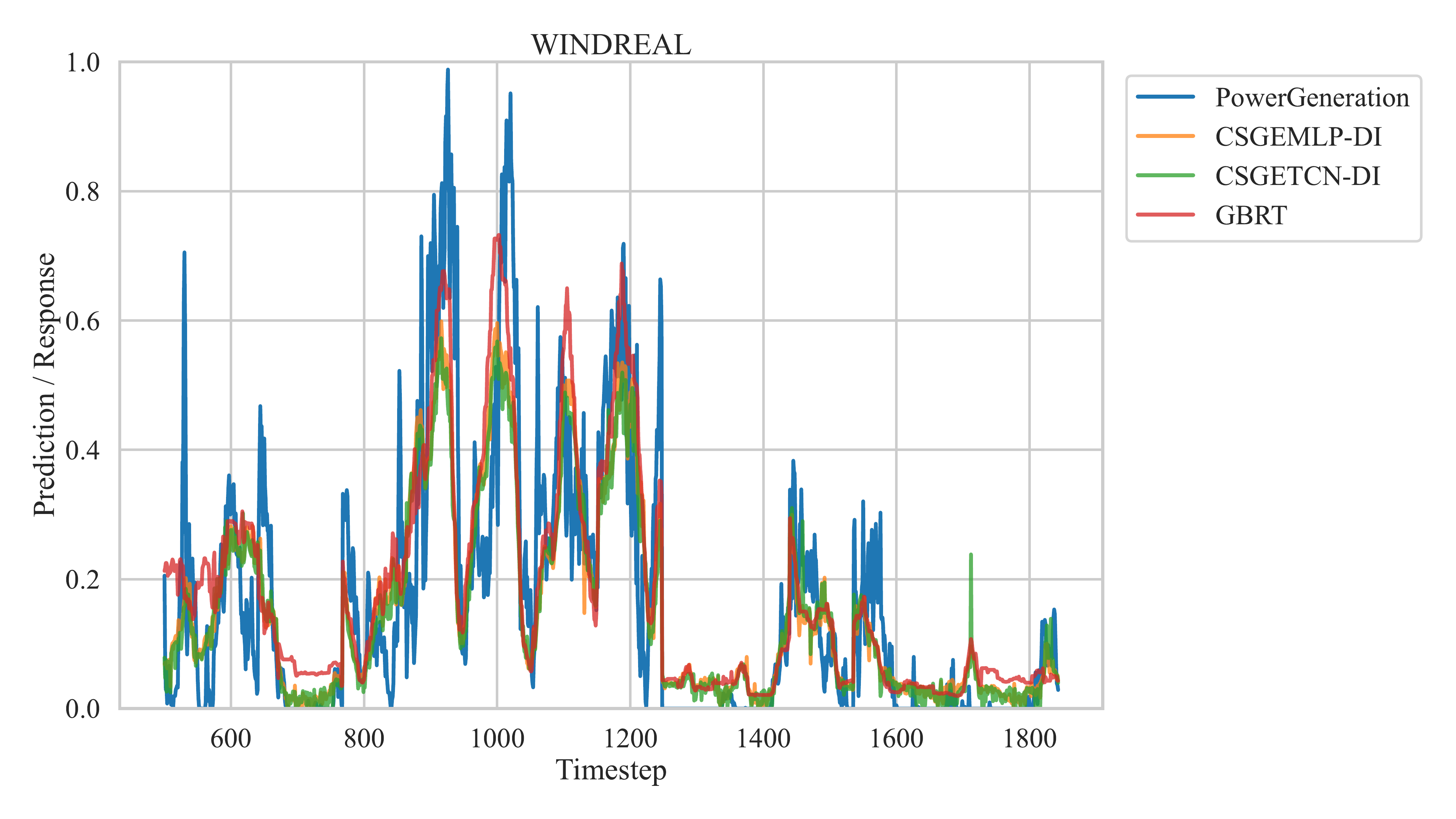}
    \caption{Example of a deterministic forecast for a wind Park.}
\end{figure}
\begin{figure}
    \centering
    \includegraphics[width=0.95\textwidth]{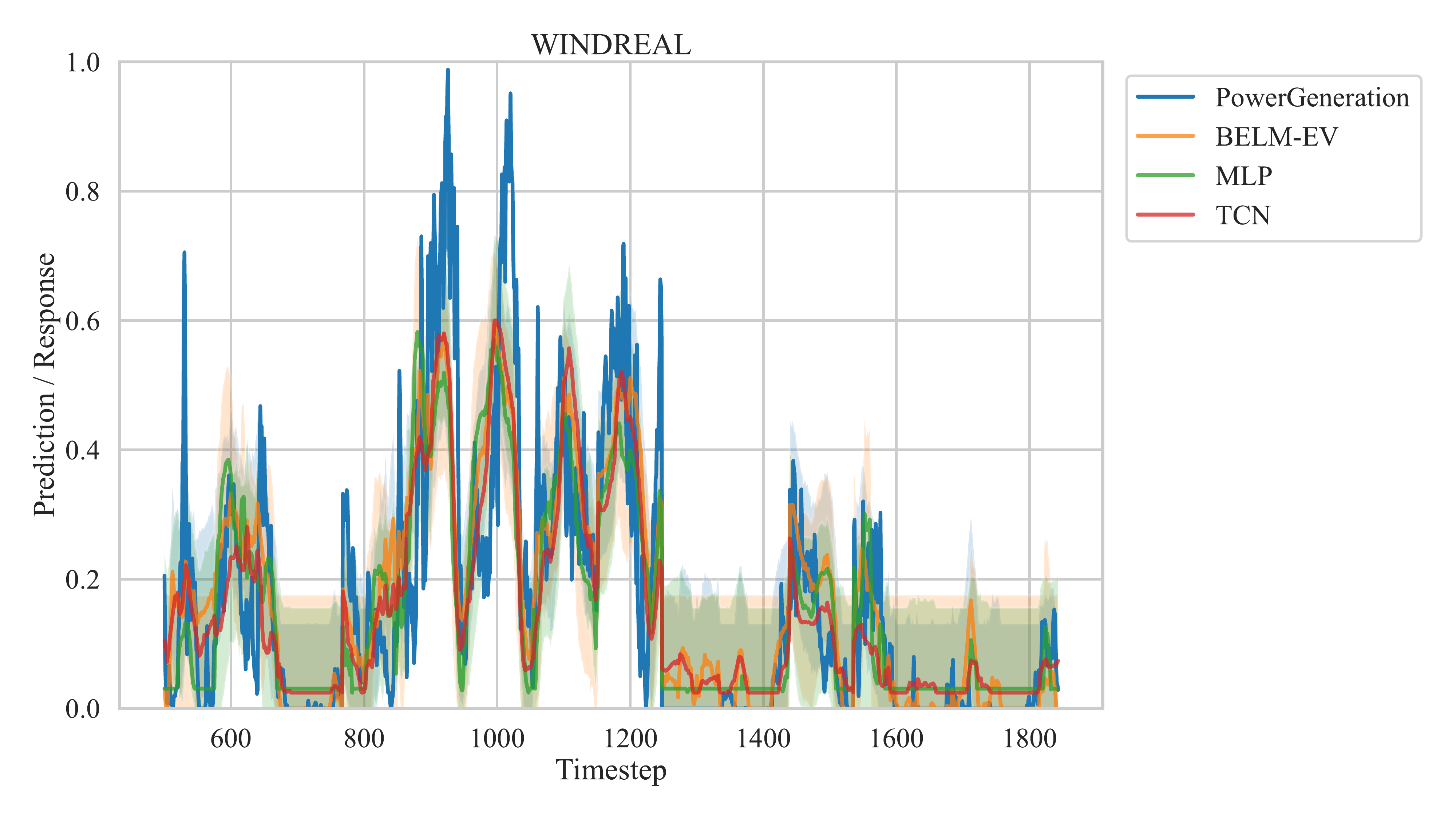}
    \caption{Example of a probabilistic forecast from a single model for a wind Park.}
\end{figure}
\begin{figure}
    \centering
    \includegraphics[width=0.95\textwidth]{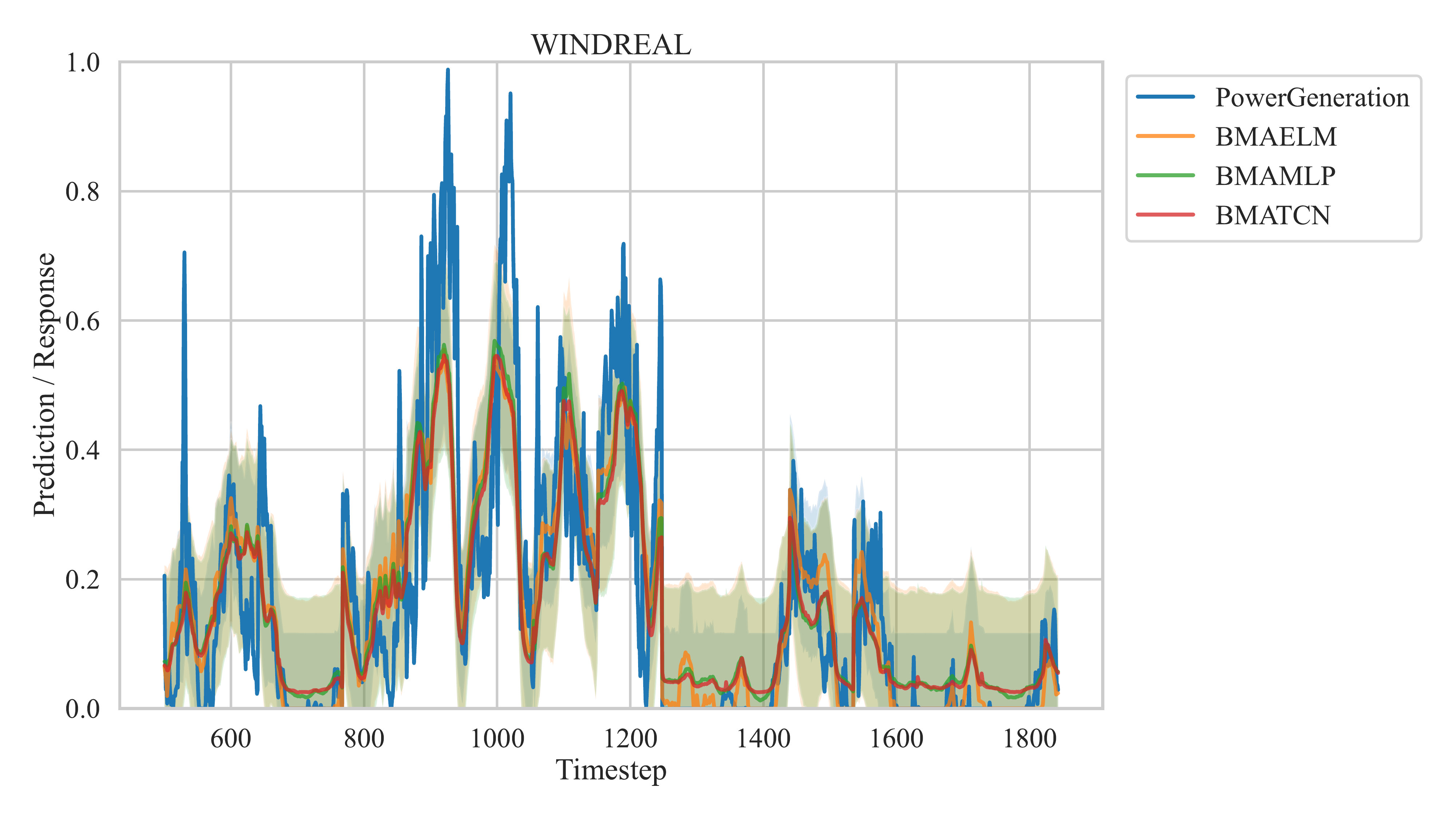}
    \caption{Example of a probabilistic forecast based on an ensemble for a wind Park.}
\end{figure}

\end{document}